\documentclass[twoside]{article}

\usepackage[accepted]{aistats2024}

\usepackage{graphicx} 
\usepackage{amsfonts,amsmath,amssymb}       

\usepackage[utf8]{inputenc} 
\usepackage[T1]{fontenc}    
\usepackage{hyperref}       
\usepackage{url}            
\usepackage{booktabs}       
\usepackage{amsfonts}       
\usepackage{nicefrac}       
\usepackage[dvipsnames,svgnames]{xcolor}         

\usepackage{amsthm}
\usepackage[ruled]{algorithm2e}
\usepackage{comment}
\usepackage{subcaption}

\newcommand{\R}{\mathbb{R}}
\newcommand{\E}{\mathbb{E}}
\newsavebox\CBox 
\newcommand*{\textBF}[1]{\rlap{\raisebox{0pt}[0pt][0pt]{\textbf{#1}}}\phantom{#1}}

\makeatletter
\newcommand\notsotiny{\@setfontsize\notsotiny{8}{8}}
\newcommand\supertiny{\@setfontsize\supertiny{6}{6}}
\makeatother

\newtheorem*{rep@theorem}{\rep@title}
\newcommand{\newreptheorem}[2]{%
\newenvironment{rep#1}[1]{%
 \def\rep@title{#2 \ref{##1}}%
 \begin{rep@theorem}}%
 {\end{rep@theorem}}}
\makeatother

\newtheorem{theorem}{Theorem}

\usepackage[round]{natbib}

\begin{document}

%

%

\twocolumn[

\aistatstitle{Generating and Imputing Tabular Data via Diffusion and Flow-based Gradient-Boosted Trees}

\aistatsauthor{ Alexia Jolicoeur-Martineau \And Kilian Fatras$^\dagger$ \And Tal Kachman }

\aistatsaddress{ Samsung - SAIT AI Lab \And  Mila \& McGill University \And Radboud University \& Donders Institute } ]

\begin{abstract}
    {Tabular data is hard to acquire and is subject to missing values. This paper introduces a novel approach for generating and imputing mixed-type (continuous and categorical) tabular data utilizing score-based diffusion and conditional flow matching. In contrast to prior methods that rely on neural networks to learn the score function or the vector field, we adopt XGBoost, a widely used Gradient-Boosted Tree (GBT) technique. To test our method, we build one of the most extensive benchmarks for tabular data generation and imputation, containing 27 diverse datasets and 9 metrics. Through empirical evaluation across the benchmark, we demonstrate that our approach outperforms deep-learning generation methods in data generation tasks and remains competitive in data imputation. Notably, it can be trained in parallel using CPUs without requiring a GPU. Our Python and R code is available at \url{https://github.com/SamsungSAILMontreal/ForestDiffusion}}.
\end{abstract}

\begin{figure*}[ht]
    \vspace{-0.5cm}
    \centering
    \includegraphics[width=0.92\textwidth]{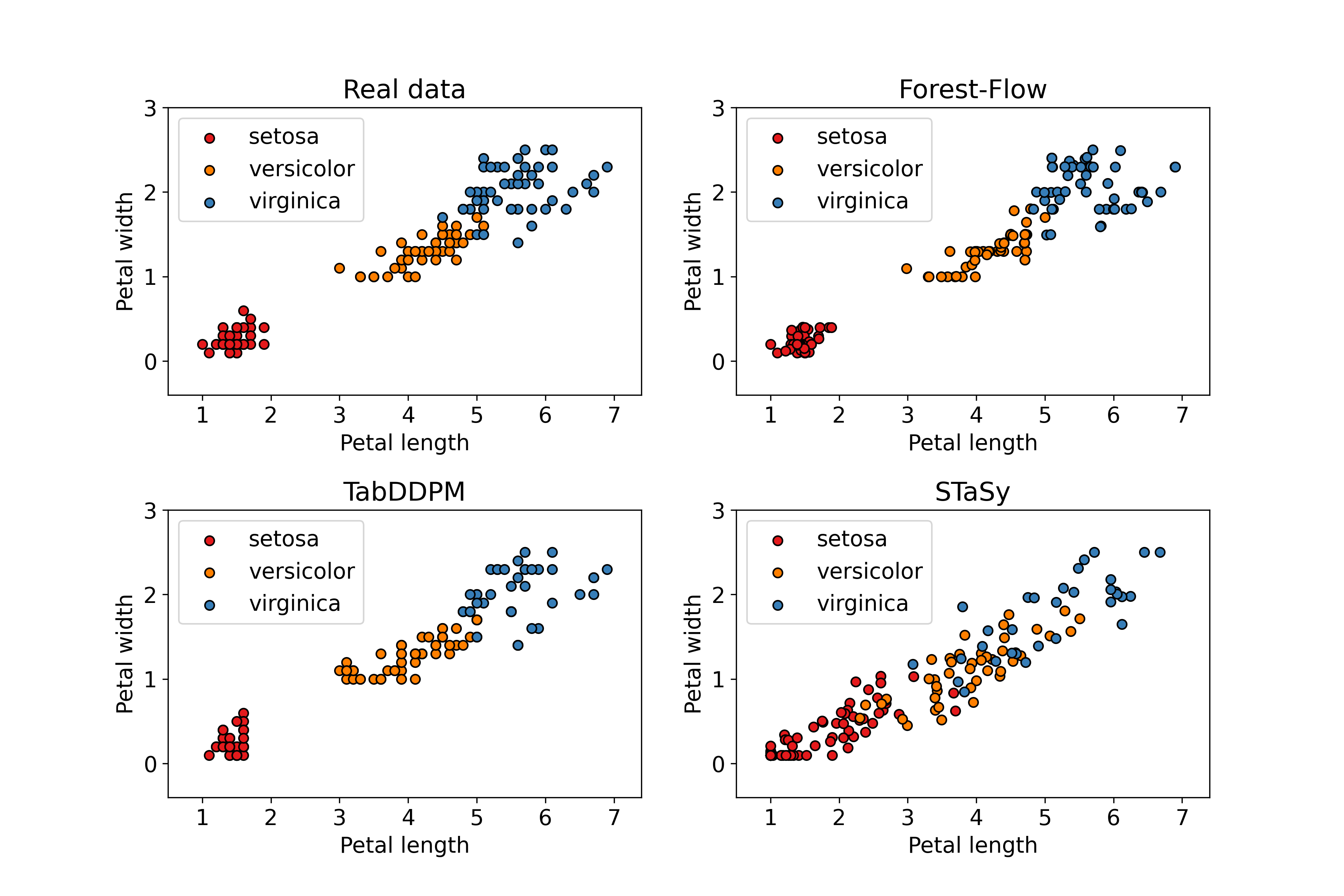}
    \vspace{-0.6cm}
    \caption{Iris dataset: Three-way interaction between Petal length, width, and species using real or fake samples (using Forest-Flow, our XGBoost method, or TabDDPM and STaSy, deep-learning diffusion methods)} 
    \label{fig:iris_1}
\end{figure*}

\section{INTRODUCTION}

{Tabular datasets are omnipresent across various fields like economics, medicine, and social sciences.  Two interconnected fundamental challenges with this type of data are small training datasets and missing values\citep{little1989analysis, bennett2001can}. To make the problems worse, different participants may have missing data for different variables. Consequently, training a model solely on complete cases would significantly shrink the participant pool, potentially exacerbating the already limited dataset size \citep{muzellec2020missing}, thereby reducing statistical power and leading to bias \citep{donner1982relative}.} 

{The most common method to deal with missing data is to use single or multiple imputations. Single imputation replaces the missing values in the dataset with predictions and trains the model on this imputed dataset. This can work for prediction or classification, but relying on a single guess for each missing data point can lead to incorrect inferences because of the bias in the imputed dataset. Multiple imputations generate several imputed datasets, each with its own set of imputed values. Then, a separate model is trained per imputed dataset, and the results are combined to account for the uncertainty inherent in the imputation process \citep{little1987statistical}}.

{Remarkably, there has been significant advancements in the field of generative models such as images \citep{brock2018large, karras2019style, rombach2022high}, audio \citep{chen2020wavegrad,kong2020diffwave}, videos \citep{voleti2022mcvd, harvey2022flexible}, graphs \citep{niu2020permutation}, and tabular data \citep{kim2022stasy, kotelnikov2023tabddpm, borisov2022language}. These methods offer the possibility to artificially expand datasets, addressing the challenge of limited data akin to data augmentation techniques \cite{MUMUNI2022100258}. Furthermore, the latest generative models have demonstrated remarkable effectiveness at inpainting \citep{meng2021sdedit,lugmayr2022repaint, zheng2022diffusion, ouyang2023missdiff}, which involves restoring missing image portions, much like imputing missing data in the case of tabular data \citep{yun2023imputation}. This demonstrates the potential of generative models to alleviate the challenges of limited data and missing values encountered in tabular datasets.}

One of the most successful recent generative models is Diffusion Models (DMs) \citep{sohl2015deep, song_generative_2019, song_improved_2020, song_score-based_2021, ho_denoising_2020}. DMs estimate the score-function (gradient log density) and rely on Stochastic Differential Equations (SDEs) to generate samples. A more recent approach called Conditional Flow Matching (CFM) instead estimates a vector field and relies on Ordinary Differential Equations (ODEs) to generate data \citep{liu2022flow, albergo2023building, lipman_flow_2022, tong2023improving, tong2023simulationfree}. {Both methods have been successful in data generation tasks.}

Traditionally, DMs and CFMs have relied on deep neural networks to estimate the score-function or the vector field because Neural Networks (NNs) are considered Universal Function Approximators (UFAs) \citep{hornik1989multilayer}. However, many other UFAs exist. For example, Decision Trees and more complex tree-based methods such as Random Forests \citep{breiman2001random} or Gradient-Boosted Trees (GBTs) \citep{friedman2000additive, friedman2001greedy} are also UFAs \citep{royden1968real, watt2020machine}. Furthermore, for tabular data prediction and classification, GBTs tend to perform better than neural networks \citep{shwartz2022tabular, borisov2022deep, grinsztajn2022tree}, and most of them can natively handle missing data through careful splitting \citep{chen2015xgboost,prokhorenkova2018catboost}.

Because Gradient-Boosted Trees (GBTs) perform particularly well on tabular data, we sought to use GBTs as function estimators in flow and diffusion models. In doing so, our \textbf{main contributions} are:
\begin{itemize}\itemsep-0.4em 
    \item We create the first diffusion and flow models for tabular data generation and imputation using XGBoost {(see Figure \ref{fig:iris_1})}, a popular GBT method, instead of neural networks.
    \item Contrary to most generative models for tabular data, our method can be trained directly on incomplete data thanks to XGBoost, which learns the best split for missing values.
    \item We provide an extensive benchmark for generation and imputation methods on 27 real-world datasets with a wide range of evaluation metrics tackling four quadrants: closeness in distribution, diversity, prediction, and statistical inference.
    \item Our method generates highly realistic synthetic data when the training dataset is either clean or tainted by missing data.
\end{itemize}

\section{BACKGROUND}

\subsection{Gradient-boosted trees and XGBoost}

Decision Trees are predictive models that partition input data into distinct subsets via decision splits, thereby culminating in terminal nodes, each furnishing a definitive prediction. They recursively partition the feature space to maximize the homogeneity of predictions within each partition. By strategically selecting decision splits based on certain criteria, they efficiently maximize predictive performance.

GBTs \citep{friedman2000additive, friedman2001greedy} take decision trees a step further by building decision trees sequentially, where each tree corrects the errors made by the previous one. It starts with a simple tree, often called a weak learner, and then iteratively adds more trees while emphasizing the examples that the previous trees predicted incorrectly. This iterative process continues until a specified number of trees are built or until a certain level of accuracy is achieved. GBTs have shown great success for tabular data prediction and classification \citep{zhang2017up,touzani2018gradient,machado2019lightgbm,ma2020diagnostic}.

XGBoost (eXtreme Gradient Boosting) is a popular open-source GBT. It uses a second-order Taylor expansion, fast parallelized/distributed system, fast quantile splitting, and sparsity-aware split finding (allowing it to naturally handle missing values) for maximum performance \citep{shwartz2022tabular, florek2023benchmarking}. We rely on XGBoost as our ablation (see Section \ref{sec:ablation}) showed it performed the best.

\subsection{Generative diffusion and conditional flow matching models}
Previous generations of generative models, such as GANs \cite{goodfellow2014generative} or VAEs \cite{kingma2013auto}, requires to differentiate through two models. This prevents the use of non-differentiable models such as GBTs. However, the new diffusion and flow-based models only need one model, which paves the way for using GBTs in generative modeling.

\paragraph{SDEs and score-based models} \label{sec:sdes}
 The purpose of generative models is to generate realistic data from Gaussian noise. As highlighted in \cite{song_score-based_2021}, a possible manner to generate data can be done through stochastic differential equations (SDE) \citep{Feller1949OnTT}. The forward diffusion process consists of transforming data into Gaussian noise through an SDE of the form:
\begin{equation}\label{eq:sde}
dx = u_t(x)\,dt + g(t) dw,
\end{equation}
where $t \in [0,1]$, $u : [0, 1] \times \R^d \to \R^d$ is a smooth time-varying vector field, $g : \R \to \R$ is a scalar function and $w$ is a Brownian motion. Different choices of forward diffusion process exists to ensure that $x(t=0)$ is a real data, while $x(t=1)$ is pure Gaussian noise. Some of the popular choices are the Variance Preserving (VP) \citep{ho_denoising_2020} or the Variance Exploding (VE) \citep{song_generative_2019} SDEs. Importantly, it is possible to reverse the forward diffusion process \eqref{eq:sde} through a reverse SDE of the form:
\begin{equation}\label{eq:reverse_sde}
dx = [u_t(x) - g(t)^2 \nabla_x \log p_t(x)]\,dt + g(t) d\bar{w},
\end{equation}
where $\bar{w}$ is the reverse Brownian motion. The main idea of score-based models is to i) learn the score function $\nabla_x \log p_t(x)$ by perturbing real data with Gaussian noise \cite{song_generative_2019} with the score-matching loss, which reads:
\begin{equation}\label{eq:score_matching}\hspace{-0.25cm}
    l_{\rm sm}(\theta)=\E_{t} \lambda_t^2 \E_{x_0, x_t|x_0} \|s_\theta(t,x_t)-\nabla\log p_t(x_t|x_0)\|^2,
\end{equation}
where $t\sim U(0,1)$ and $\lambda_t$ is a positive weight. 
Then, ii) generate fake data by solving the reverse SDE \eqref{eq:reverse_sde} with the score function approximation while starting at random Gaussian noise (effectively going backward from noise to data). In practice, to solve the reverse SDE, we generally discretize $t$ over a fixed set of noise levels $n_t$ going from $t=1$ to $t=0$.

\begin{figure*}[ht]
    \centering
    \includegraphics[width=\textwidth]{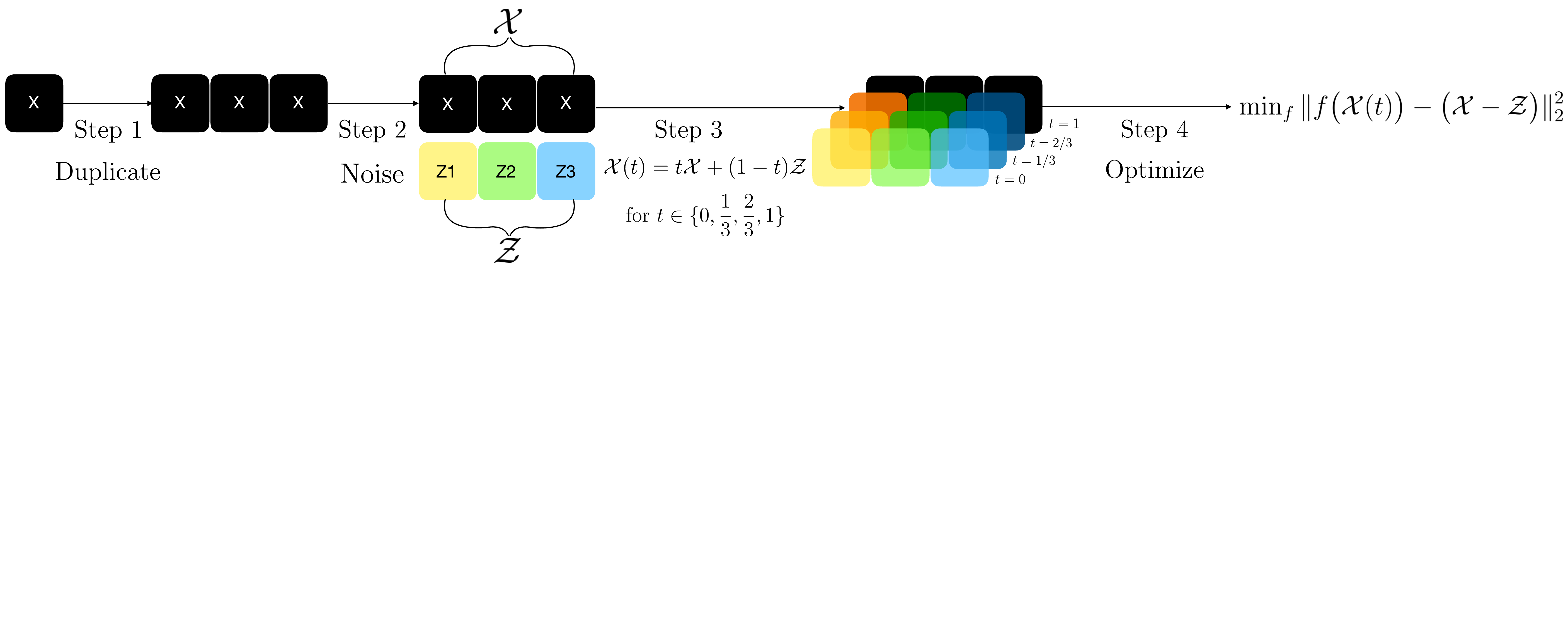}
    \vspace{-4.25cm}
    \caption{{Illustration of our Forest-Flow method based on I-CFM \cite{tong2023improving} (see \S\ref{app:fm_euc} for more details). The first step duplicates the original dataset. The second step adds a different noise to each duplicated dataset. The third step computes the linear interpolation between the duplicated dataset and their corresponding noise for different time $t$ (\emph{i.e.,} $\mathcal{X}_i(t) = t\mathcal{X} + (1-t) \mathcal{Z}_i, \forall i \in [1, \ldots, n_{noise}]$ and $\forall t \in t_{\text{levels}}$). The final step is to regress a GBT model at each noise level against the vector field; the training of the $n_t$ models is parallelized over CPUs.}}
    \label{fig:expectation_approx}
\end{figure*}

\paragraph{ODEs and conditional flow matching} \label{sec:odes}

When $g(t)=0$ in \eqref{eq:sde}, we recover the setting of an ordinary differential equation (ODE), making the forward and reverse flow deterministic and simplifying the problem to be solved. We denote by $\phi_t(x)$ the solution of this ODE with initial conditions $\phi_0(x)=x$; that is, $\phi_t(x)$ is the point $x$ transported along the vector field $u$ from $t=1$ up to $t=0$. Given a density $p_0$ over $\R^d$, we can define the time-varying density $p_t=[\phi_t]_\#(p_0)$, where $[\phi_t]_\#$ is the pushforward operator of $\phi$. By construction, the density $p_t$ verifies the well-known \emph{continuity equation} with initial conditions $p_0$.

Approximating ODEs with neural networks is a rich research direction \citep{kidger2022neural}. \cite{chen_neural_2018} approximated the vector field $u_t(x)$ with a neural network $v_\theta(t,x):[0,1]\times \R^d\to\R^d$. However, this direction requires the simulation of the ODE, which does not scale well to high dimensions. Another direction is to regress $v_\theta(t,x)$ against the vector field $u_t(x)$. Unfortunately, this loss is intractable as it requires knowing the probability path $p_t(x)$ and the vector field $u_t(x)$. 

A workaround was proposed in \cite{lipman_flow_2022}. In their framework, they assume that the probability path is a mixture of conditional probability paths. Formally, $p_t(x) = \int p_t(x|z) q(z)dz$, where $q$ is a latent distribution and the conditional probability paths $p_t(x | z)$ are supposed to be generated from some conditional vector fields $\mu_t(x | z)$. Then, they defined the following vector field: $\mu_t(x) = \E_{q(z)} \frac{\mu_t(x | z) p_t(x | z)}{p_t(x)}$ and showed that it generates the probability path $p_t(x)$ (\emph{i.e.,} $\mu_t$ and $p_t$ verify the continuity equation). They also proved that regressing $v_\theta(t,x)$ against the conditional vector field $\mu_t(x|z)$ leads to the same gradients as regressing $v_\theta(t,x)$ against the vector field $\mu_t(x)$. Therefore, they minimize: 
\begin{equation}\label{eq:CFM}
l_{\rm cfm}(\theta) = \E_{t, q(z), p_t(x | z)} \|v_\theta(t, x) - \mu_t(x | z)\|^2.
\end{equation}
 After training $v_\theta$, they generate fake samples by solving the following ODE ($t=1$ to $t=0$)\footnote{Note that we have considered ODE from $t=1$ to $t=0$ for consistencies with respect to diffusion models.}: 
\begin{equation}\label{eq:ode}
dx = v_\theta(t, x)\,dt.
\end{equation} This direction has made ODE-based generative models competitive compared to their stochastic variant. 

In practice, we use the  I-CFM method \cite{tong2023improving}. This method assumes that $z$ is a tuple of noise and real data (\emph{i.e.,} $z=(x_0, x_1)$) and the distribution $q$ is the independent coupling $q(x_0, x_1) = q(x_0) q(x_1)$. We choose the conditional probability path to be $p_t(x | (x_0, x_1)) = \mathcal{N}(t x_1 + (1-t)x_0, \sigma)$ for $\sigma>0$. This results in the following conditional vector field: $\mu_t(x | (x_0, x_1)) = x_1 - x_0$. We refer to \S\ref{app:fm_euc} for a longer discussion on Flow Matching.

\section{TRAINING DIFFUSION \& FLOW MODELS WITH XGBOOST }

Traditionally (and exclusively as far as we know), flow and diffusion models have relied on deep neural networks. Instead, our method relies on Gradient-boosted Trees (GBTs) to estimate the vector field or score-function. We explain the details of our method below (see also Figure \ref{fig:expectation_approx} for an illustration of the method).

Let us assume that we have a training dataset $X$ of size $[n,d]$ and that we have discrete set of $n_t$ noise levels $t \in t_{levels} = \{ \frac{1}{n_{t}}, \ldots, \frac{n_{t}-1}{n_{t}}, 1 \}$, where $t=0$ corresponds to real data, while $t=1$ corresponds to Gaussian noise.

\subsection{Duplicating the dataset to estimate the expectation in diffusion and flow losses}

{A key component of NNs training is Stochastic Gradient Descent (SGD). To use SGD for training a diffusion model, a mini-batch of data of size $b$ is selected from the training dataset along with random Gaussian noise of the same size (i.e., $[b,d]$). For each of the $b$ samples, a random noise level $t$ and a Gaussian noise $z$ (of size $d$) are sampled. These are then used to compute the noisy sample $x(t)$ using the forward diffusion/flow step. By using SGD with random sampling, we can minimize the expectations \eqref{eq:score_matching} and \eqref{eq:CFM} over mini-batch of data, Gaussian noise, and noise level.

Unfortunately, GBTs do not rely on SGD and are instead trained on the entire dataset (although each tree can be trained on different subsamples of the data as a regularization). Therefore, we need to prepare a single input and output training dataset precomputed in advance to approximate the expectations \eqref{eq:score_matching} and \eqref{eq:CFM}. Since the expectations are over all possible noise-data pairs, we need to sample multiple Gaussian noise $z$ per data sample $x$, \emph{i.e.,} we build a set $\{(z_i, x)_{i \in [1, \ldots, n_{noise}]}\}$. To achieve this, we duplicate the observations (\emph{i.e.,} the rows) of the original dataset $n_{noise}$ times (going from size $[n, d]$ to size $[n_{noise} n, d]$) as illustrated in Step 1 of Figure \ref{fig:expectation_approx}. Then, we sample random Gaussian noise of the same dimension ($[n_{noise} n, d]$), as illustrated in Step 2 of Figure \ref{fig:expectation_approx}, so that each sample get $n_{noise}$ different random noise. These two datasets (containing the duplicated data samples and the random Gaussian noise) are used to produce one fixed dataset $X(t)$ containing all the pre-calculated noisy samples $x(t)$ and one fixed dataset $Y(t)$ containing their corresponding outputs $y(t)$ ($\nabla\log p_t(x_t|x_0)$ for diffusion model and $\mu_t(x | z)$ for flow model) for all $t$\footnote{Since the release of XGBoost 2.0, data duplication can be avoided using XGBoost's novel data iterator which mimics minibatches. Other GBTs still require this procedure.}. It is illustrated in the third step of  Figure \ref{fig:expectation_approx}. The entire process for our Flow-based method is illustrated in Figure \ref{fig:expectation_approx}.

In practice, $n_{noise}$ is one of the most critical hyperparameters for achieving good performance because it controls the approximation of the expectation over data-noise pairs. Therefore, $n_{noise}$ should be as high as possible, given memory constraints. We set $n_{noise}=100$ for all datasets, unless memory is an issue (when $nd$ is large), in which case we reduce it to $n_{noise}=50$.}

\subsection{Training different models per noise level}

As seen from \eqref{eq:score_matching} and \eqref{eq:CFM}, the model depends on both $x_t$ and $t$. Since training a neural network to approximate the vector field or score-function can be very slow, it is usual to take the noise level $t$ as an input (which is processed through a complex Sinusoidal \citep{vaswani2017attention} or Fourier feature \citep{tancik2020fourier} embedding). The same idea can be done with GBTs by taking $t$ as an additional model feature. However, this approach is problematic when the number of variables $d$ is large. For simplicity, assume that the variables used for the tree splits are randomly chosen, then the probability of splitting a node by $t$ is $\frac{1}{d+1}$. This highlights the fact that with many variables, the chance of splitting by the noise level is very small, which greatly minimizes the influence of $t$ on the output.

\begin{figure}[t!]
    \hspace{0.35cm}
    \includegraphics[width=0.7\textwidth]{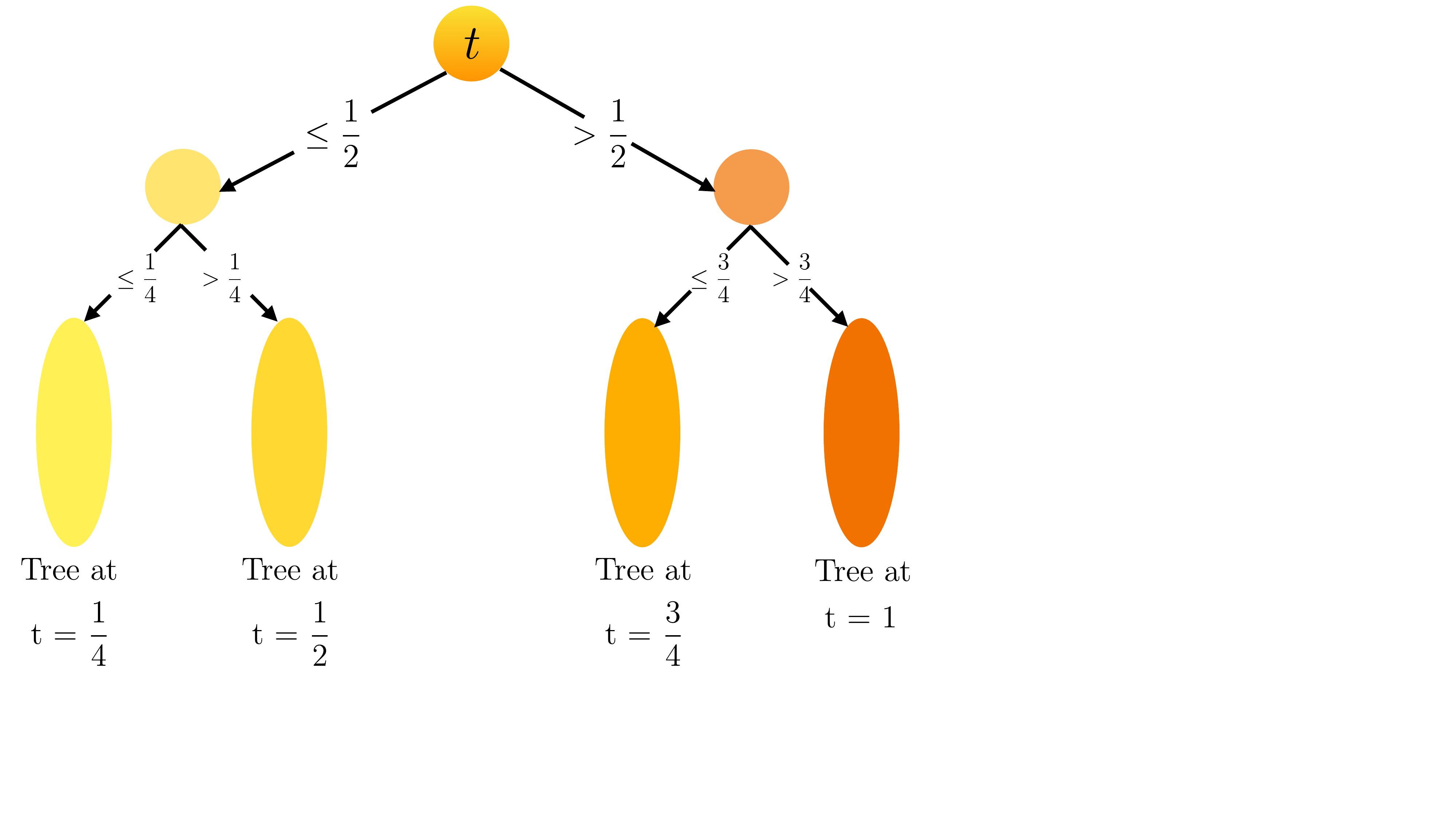}
    \vspace{-1.6cm}
    \caption{Our method learns a different GBT model (with 100 trees) for each noise-level (here $n_t =4$). We can re-interpret this as a single model with giants trees where the time variable splits are hard-coded.} 
    \label{fig:fig_tree}
\end{figure}
To better differentiate the vector field or score at different noise levels, \textit{we train a different model per noise level $t$}, leading to $n_t$ models whose training can be parallelized over multiple CPUs. cost. It can be seen as a model where the time variable splits are hard-coded (see Figure \ref{fig:fig_tree}). In practice, $n_t=50$ is enough to reach state-of-the-art or competitive performance. Increasing $n_t$ could lead to greater performance but at the price of a higher computational cost.

\subsection{Choice of Gradient-boosted Trees}

In theory, we could use any Gradient-boosted Trees. We experimented our method with Random Forests \citep{ho1995random, breiman2001random}, XGBoost \citep{chen2015xgboost}, LightGBM \citep{ke2017lightgbm}, and CatBoost \citep{prokhorenkova2018catboost}. We have found that XGBoost achieved superior performance than all other GBT methods (see  Ablation \S\ref{sec:ablation}). Furthermore, Random Forests cannot be trained with missing data, motivating our choice to exclusively use XGBoost as GBT for our approach (see \S\ref{sec:imputation}).

\subsection{Forward Diffusion/Flow process}

There are many possible choices for the forward process of diffusion and flow models. For diffusion, we use the Variance Preserving (VP) process \citep{song_score-based_2021}. For CFM, we use the deterministic independent-coupling conditional flow matching (I-CFM) \citep{tong2023improving}.
See Appendix \ref{app:forward_reverse} for the forward and reverse processes of VP and I-CFM.

\begin{algorithm}[t!]
    \caption{Forest-Diffusion Training}
    \label{alg:train}
     \textbf{Input:} Dataset $X$ of size $[n,d]$, $n_{noise}=100$ noise sample per data sample, $n_{t}=50$ noise levels.

     $X' \gets$ Duplicate the rows of $X$ $n_{noise}$ times 
     
     $Z \gets $ Dataset of  $z \sim \mathcal{N}(0,1)$ with the size of $X'$

     $t_{levels} = \{ \frac{1}{n_{t}}, \ldots, \frac{n_{t}-1}{n_{t}}, 1 \}$

    \tcp{Parallelized on CPUs}
     \For {$t \in t_{levels}$}
     {
        $X(t), Y(t) \gets \text{Forward}(Z, X', 0, t \mid n_t)$

        $f(t) \gets$ Regression XGBoost model to predict $Y(t)$ given $X(t)$
     }
     
     \Return $\{f(t) \}_{t \in t_{levels}}$
\end{algorithm}

\begin{algorithm}[t]
    \caption{Forest-Diffusion Sampling}
    \label{alg:samp}
     \textbf{Input:} XGBoost models $\{f(t)\}_{t \in t_{levels}}$, $n_{obs}$ samples, $d$ variables, $n_{t}=50$ noise levels.

     $X(1) \gets $ Dataset of $z \sim \mathcal{N}(0,1)$ with size $[n_{obs},d]$

     $ t \gets 1$

     \While {$t > 0$}
     {
        $X(t - \frac{1}{n_t}) \gets \text{Reverse}(X(t) \mid f(t), t, n_t)$
        $ t \gets t - \frac{1}{n_t}$
     }

     \Return $X(0)$
\end{algorithm}

\subsection{Imputation via diffusion and XGBoost}\label{sec:imputation}

As previously mentioned, inpainting (restoring missing parts of an image) is effectively the same problem as imputing missing data when dealing with tabular data. For this reason, as a method of imputation, we use REPAINT \citep{lugmayr2022repaint}, a powerful inpainting diffusion models method.

A critical distinction between the training of diffusion models for image inpainting and the training of GBTs diffusion/flow models is that inpainting models are trained on complete images. In contrast, we must train on incomplete data. For this reason alone, XGBoost is particularly interesting for this task because it can handle missing data by learning the best splitting direction for missing values. Thus, we leverage this property to train a diffusion model on incomplete data and use it to impute missing data with REPAINT. 

Importantly, REPAINT is an algorithm made for diffusion models. By their deterministic nature, flow models are not suited for inpainting and we are not aware of any CFM method which tackles inpainting. We provide some additional intuition as to why it may not be possible to impute nor inpaint in Appendix \ref{app:inpaint_flow}.

\subsection{Data processing} 
\label{sec:proc}

{\bfseries Pre-processing} Tabular data can contain both numerical and categorical variables. Our method takes continuous variables in $\mathbb{R}$. We make the categorical variables continuous by dummy encoding them \citep{suits1957use}. Then, we min-max normalize every variable (including dummies) to the range $[-1,1]$.

{\bfseries Post-processing} After data generation (or imputation), we clip the generated values between $[-1,1]$ and reverse the min-max normalization. For integer variables, we also round these variables to remove any decimals. Finally, we round the dummy variables to the nearest class to obtain the categorical variables.

\begin{algorithm}[t!]
    \caption{Forest-Diffusion Imputation}
    \label{alg:imp}
     \textbf{Input:} XGBoost models $\{f(t)\}_{t \in t_{levels}}$, dataset X of size $[n_{obs},d]$, $n_{t}=50$ noise levels.

     $M \gets $ Mask indicating which values are non-missing in $X$ (1: non-missing, 0: missing)

     $X(1) \gets $ Dataset of $z \sim \mathcal{N}(0,1)$ with size $[n_{obs},d]$

     $ t \gets 1$

     \While {$t > 0$}
     {

        $X(t-\frac{1}{n_t}) \gets $ Empty Dataset with size $[n_{obs},d]$

        \tcp{Reverse process for missing values}
     
        $X(t-\frac{1}{n_t})[1-M] \gets \text{Reverse}(X(t)[1-M] \mid f(t), t, n_t)$

        \tcp{Set non-missing values to truth}

        $X(t-\frac{1}{n_t})[M] \gets \text{Forward}(Z[M], X[M], 0, t \mid n_t)$

        $ t \gets t - \frac{1}{n_t}$
        
     }
     \Return $X(0)$
\end{algorithm}

\subsection{Gradient-Boosted Tree hyperparameters}  \label{sec:reg}

Since our goal is to estimate the flow or score-function of the distribution, overfitting is of very little concern, while underfitting is a major concern. For this reason, we do not use any regularization, such as $L_{1}$ or $L_{2}$ penalizations. Outside of the $L_{2}$ hyperparameter, which we set to 0, all other hyperparameters of XGBoost are left at their default values (note that the default number of trees is 100).

\subsection{Training one model per category}  \label{sec:catreg}

When the dataset has a categorical outcome (i.e., a dataset for classification), \citet{kotelnikov2023tabddpm} train their score function conditional on the label. Then, they generate data by i) sampling a random label with the same probabilities as the training data and ii) generating fake data conditional on that label. They do so to improve performance and reduce the degrees of freedom by one. We use this idea, but instead of conditioning on the label, we train a different XGBoost model per label. It is equivalent to forcing each tree to split by the labels before building deep trees. We find that it improves generation performance.

\subsection{Algorithm details}

The training algorithm is described in Algorithm \ref{alg:train}. The sampling and imputation methods are described in Algorithms \ref{alg:samp} and \ref{alg:imp}. Note that we removed REPAINT from the imputation algorithm for simplicity, but the algorithm using REPAINT can be found in the Appendix (see Algorithm \ref{alg:imp_repaint}). Again, for simplicity, We also leave out the details on the Forward and Reverse process in the Appendix (see Algorithms \ref{alg:forward}, \ref{alg:forward_next}, and \ref{alg:reverse}). 

\begin{table}[t!]
\centering
\setlength{\tabcolsep}{4pt}
\notsotiny
\caption{Evaluation metrics used in our experiments}
\label{tab:metrics}
\begin{tabular}{p{2.3cm}p{1.3cm}p{3.6cm}}
\toprule
Metric & Abbreviation & \qquad Purpose \\ \hline
\multicolumn{3}{c}{ \emph{Distance in distribution} (or to ground-truth)} \\  \hline
Wasserstein distance & $W_{train}$, $W_{test}$ & Distance to train/test data distributions \\ \hline
Minimum Mean Absolute Error & MinMAE & Distance between closest imputation and non-missing data \\ \hline
Average Mean Absolute Error & AvgMAE & Average distance between imputation and non-missing data \\ \hline
\multicolumn{3}{c}{ \emph{Diversity}} \\  \hline
Coverage & $cov_{train}$, $cov_{test}$ &  Diversity of fake samples relative to train/test data \\ \hline
Mean Absolute Deviation & MAD & Diversity of imputations \\ \hline
\multicolumn{3}{c}{ \emph{Prediction}} \\  \hline
R-squared and F1-score & $R^{2}_{fake}$, $F1_{fake}$, 
$R^2_{imp}$, $F1_{imp}$ & Usefulness of fake or imputed data for Machine Learning prediction/classification \\ \hline
Discriminator F1-score & $F1_{disc}$ & Ability to distinguish real from fake data \\ \hline
\multicolumn{3}{c}{ \emph{Statistical inference}} \\  \hline
Percent Bias & $P_{bias}$ & Regression parameter precision \\ \hline
Coverage Rate & $cov_{rate}$ & Regression parameter confidence intervals coverage \\ 
\bottomrule 
\end{tabular} 
\end{table}

\begin{table*}[ht]
\caption{Tabular data generation with complete data (27 datasets, 3 experiments per dataset); \textit{averaged rank} over all datasets and experiments (standard-error). Best highlighted in \textbf{bold}.}
\label{tab:gen}
\notsotiny
\centering
\begin{tabular}{r|ll|ll|lll|ll}
  \toprule
 & $W_{train} \downarrow$ & $W_{test} \downarrow$ & $cov_{train}$ $\downarrow$ & $cov_{test}$ $\downarrow$ & $R^2_{fake} \downarrow$ & $F1_{fake} \downarrow$ & $F1_{disc} \downarrow$ & $P_{bias} \downarrow$ & $cov_{rate} \downarrow$ \\ \hline
GaussianCopula & 6.1 (0.3) & 6.2 (0.3) & 6.3 (0.3) & 6.4 (0.3) & 5.2 (0.2) & 5.6 (0.3) & 6.1 (0.4) & 5.7 (1.0) & 6.7 (0.6) \\ 
  TVAE & 4.3 (0.2) & 4.1 (0.2) & 4.7 (0.2) & 4.7 (0.2) & 5.5 (0.7) & 5.2 (0.5) & 4.9 (0.2) & 6.5 (0.5) & 6.0 (0.4) \\ 
  CTGAN & 7.4 (0.1) & 7.4 (0.2) & 7.3 (0.2) & 7.1 (0.2) & 7.5 (0.2) & 7.3 (0.2) & 6.1 (0.3) & 4.7 (1.0) & 6.3 (0.5) \\ 
  CTAB-GAN+ & 5.8 (0.3) & 5.7 (0.3) & 6.3 (0.3) & 6.1 (0.3) & 6.0 (0.3) & 6.0 (0.3) & 6.5 (0.2) & 6.8 (0.7) & 6.0 (0.8) \\ 
  STaSy & 5.1 (0.2) & 5.3 (0.2) & 4.3 (0.2) & 4.4 (0.2) & 5.5 (1.0) & 4.3 (0.3) & 5.1 (0.3) & 4.2 (0.7) & 3.9 (0.9) \\ 
  TabDDPM & 2.7 (0.6) & 3.4 (0.5) & 2.6 (0.4) & 2.9 (0.4) & \textBF{1.2} (0.2) & 3.4 (0.5) & \textBF{2.0} (0.4) & 2.7 (0.7) & \textBF{1.4} (0.2) \\ 
  Forest-VP & 2.7 (0.1) & 2.5 (0.1) & 2.9 (0.2) & 2.7 (0.3) & 2.8 (0.3) & \textBF{2.0} (0.2) & 2.6 (0.3) & 3.0 (0.8) & 3.2 (0.6) \\ 
  Forest-Flow & \textBF{1.8} (0.1) & \textBF{1.4} (0.1) & \textBF{1.6} (0.2) & \textBF{1.6} (0.2) & 2.3 (0.4) & 2.1 (0.3) & 2.7 (0.3) & \textBF{2.5} (0.3) & 2.5 (0.3) \\ 
   \hline
\bottomrule 
\end{tabular}
\end{table*}

\begin{table*}[ht]
\caption{Tabular data imputation (27 datasets, 3 experiments per dataset, 10 imputations per experiment) with 20\% missing values; \textit{averaged rank} over all datasets and experiments (standard-error). Best highlighted in \textbf{bold}.}
\label{tab:imp}
\notsotiny
\centering
\begin{tabular}{r|llll|l|ll|ll}
  \toprule
 & MinMAE $\downarrow$ & AvgMAE $\downarrow$ & $W_{train} \downarrow$ & $W_{test} \downarrow$ & MAD $\downarrow$ & $R^2_{imp} \downarrow$ & $F1_{imp} \downarrow$ & $P_{bias} \downarrow$ & $Cov_{rate} \downarrow$   \\ \hline
KNN & 4.8 (0.4) & 5.5 (0.4) & 4.2 (0.4) & 4.2 (0.3) & 7.4 (0.0) & 5.7 (0.9) & 5.0 (1.0) & 5.5 (0.8) & 4.8 (0.5) \\ 
  ICE & 6.0 (0.4) & 3.9 (0.4) & 6.2 (0.5) & 6.4 (0.4) & \textBF{1.4} (0.2) & 5.2 (1.0) & 6.0 (0.6) & 4.8 (0.8) & 4.7 (0.6) \\ 
  MICE-Forest & 3.4 (0.4) & \textBF{2.0} (0.3) & 2.5 (0.2) & 2.5 (0.3) & 3.2 (0.2) & \textBF{3.3} (1.2) & 2.6 (0.9) & 4.8 (1.0) & 3.9 (0.6) \\ 
  MissForest & \textBF{2.3} (0.4) & 3.5 (0.4) & \textBF{1.5} (0.2) & \textBF{1.7} (0.3) & 4.6 (0.1) & \textBF{3.3} (1.2) & \textBF{1.9} (0.4) & 4.8 (1.3) & \textBF{3.0} (0.4) \\ 
  Softimpute & 5.9 (0.4) & 6.7 (0.3) & 6.2 (0.4) & 6.5 (0.4) & 7.4 (0.0) & 5.2 (0.8) & 6.8 (0.4) & 5.3 (0.9) & 5.9 (0.4) \\ 
  OT & 5.2 (0.4) & 5.3 (0.3) & 5.2 (0.4) & 5.2 (0.4) & 3.2 (0.2) & 5.2 (0.5) & 5.8 (0.6) & 4.5 (0.8) & 4.3 (0.5) \\ 
  GAIN & 3.9 (0.4) & 5.6 (0.3) & 5.2 (0.3) & 5.2 (0.2) & 5.9 (0.1) & 4.7 (0.8) & 4.4 (0.8) & 3.7 (1.0) & 4.5 (0.5) \\ 
  Forest-VP & 4.5 (0.4) & 3.5 (0.4) & 5.0 (0.3) & 4.5 (0.4) & 2.9 (0.3) & 3.5 (0.9) & 3.6 (0.8) & \textBF{2.5} (0.7) & 4.8 (0.6) \\ 
\bottomrule 
\end{tabular}
\end{table*}

\section{EXPERIMENTS}

We evaluate our method on 27 real-world classification/regression datasets (with input $X$ and outcome $y$) on the following tasks: 1) generation with complete data, 2) imputation, and 3) generation with incomplete data. For all tasks, we split the datasets in training (80\%) and testing (20\%) splits. We consider similar datasets as \citet{muzellec2020missing}, and they are from the UCI Machine Learning Repository~\citep{UCI} or scikit-learn \citep{scikit-learn}. The list of all datasets can be found in \S\ref{app:datasets}. To get a broad and careful assessment of the data quality, we consider a wide range of evaluation metrics accross four quadrants: closeness in distribution (or ground truth for imputations), diversity, prediction, and statistical inference (comparing regression parameters estimated with the true data versus those estimated with fake/imputed data). We summarize the evaluation metrics in Table \ref{tab:metrics}. For more details on the choice and definitions of metrics, we refer to $\S$\ref{app:metrics}.

To condense the information across the 27 datasets, we report the \textbf{average rank (with standard-error)} of each method relative to other methods. Looking at the average rank allows for easier interpretation and reduces outliers by preventing a single low or high performance from providing an unfair (dis)advantage. We provide the tables with the \textbf{averaged raw scores} of each evaluation metric as well as the \textbf{bar plots} for each dataset, metric and method in $\S$\ref{app:tables} and \S\ref{app:bar_plots}. We denote our method \textbf{Forest-VP} with VP-diffusion and \textbf{Forest-Flow} with conditional flow matching.

\begin{table*}[ht]
\caption{Tabular data generation with incomplete data (27 datasets, 3 experiments per dataset, 20\% missing values), MissForest is used to impute missing data except in Forest-VP and Forest-Flow; \textit{average rank} (standard-error) over all datasets and experiments. Best highlighted in \textbf{bold}.}  
\label{tab:gen_miss}
\notsotiny
\centering
\begin{tabular}{r|ll|ll|lll|ll}
  \toprule
 & $W_{train} \downarrow$ & $W_{test} \downarrow$ & $cov_{train}$ $\downarrow$ & $cov_{test}$ $\downarrow$ & $R^2_{fake} \downarrow$ & $F1_{fake} \downarrow$ & $F1_{disc} \downarrow$ & $P_{bias} \downarrow$ & $cov_{rate} \downarrow$ \\ \hline
GaussianCopula & 6.0 (0.3) & 6.2 (0.2) & 6.3 (0.3) & 6.1 (0.3) & 5.3 (0.4) & 5.8 (0.2) & 6.4 (0.5) & 4.7 (0.8) & 6.7 (0.6) \\ 
  TVAE & 4.2 (0.3) & 3.9 (0.2) & 4.8 (0.3) & 4.8 (0.2) & 5.0 (1.0) & 4.9 (0.5) & 5.0 (0.3) & 7.0 (0.4) & 5.5 (0.8) \\ 
  CTGAN & 7.3 (0.2) & 7.4 (0.2) & 7.4 (0.2) & 7.3 (0.2) & 7.3 (0.3) & 7.4 (0.2) & 6.0 (0.2) & 4.2 (1.1) & 6.1 (0.7) \\ 
  CTABGAN & 5.7 (0.4) & 5.6 (0.3) & 6.1 (0.3) & 5.8 (0.3) & 6.5 (0.4) & 6.3 (0.3) & 6.1 (0.3) & 6.7 (0.8) & 5.2 (0.5) \\ 
  Stasy & 4.9 (0.2) & 5.1 (0.3) & 4.3 (0.2) & 4.1 (0.3) & 5.2 (0.7) & 3.7 (0.3) & 4.3 (0.3) & 3.3 (0.4) & 4.2 (0.9) \\ 
  TabDDPM & 2.7 (0.6) & 3.0 (0.6) & \textBF{2.1} (0.4) & 2.7 (0.5) & \textBF{1.3} (0.2) & 2.9 (0.5) & \textBF{2.0} (0.4) & 3.3 (1.1) & \textBF{1.7} (0.3) \\ 
  Forest-VP & 2.8 (0.2) & 2.7 (0.2) & 3.0 (0.2) & 3.0 (0.3) & 2.7 (0.2) & \textBF{1.9} (0.2) & 2.6 (0.3) & 3.7 (0.7) & 3.8 (1.0) \\ 
  Forest-Flow & \textBF{2.3} (0.3) & \textBF{2.2} (0.3) & \textBF{2.1} (0.2) & \textBF{2.2} (0.2) & 2.7 (0.6) & 3.2 (0.3) & 3.7 (0.3) & \textBF{3.2} (0.7) & 2.8 (0.7) \\ 
   \hline
\bottomrule 
\end{tabular}
\end{table*}

\subsection{Generation with complete tabular data} \label{sec:gen_complete}
For the generation task, we generate both input $X$ and outcome $y$. We compare our method to a wide range of tabular data generative models. We consider a statistical method: Gaussian Copula \citep{joe2014dependence, SDV}. We also consider deep-learning VAE and GAN based methods: TVAE \citep{xu2019modeling}, CTGAN \citep{xu2019modeling}, CTAB-GAN+ \citep{zhao2021ctab}. Finally, we consider deep-learning diffusion methods: STaSy \citep{kim2022stasy}, and TabDDPM \citep{kotelnikov2023tabddpm}. Details on methods can be found in \S\ref{app:hyperparams}.

Averaged score of the different methods are presented in Table \ref{tab:gen}, their raw scores in Table \ref{tab:gen2} and the bar plots in \ref{app:bar_plot_gen}. We find that Forest-Flow is the highest performing method (across nearly all metrics), followed closely by both Forest-VP and TabDDPM. Forest-Flow even nearly match the Wasserstein distance to the test set of the training data (denoted oracle in the table). This shows that our method does performs extremely well at generating realistic synthetic data.

\begin{table}[t!]
\caption{Forest-Flow ablation on the Iris dataset (baseline: $n_t=50$, $n_{noise}=100$, $y_{cond}=False$)}
\label{tab:ablation_iris}
\notsotiny
\centering
\begin{tabular}{r|ccc}
  \toprule
 & $W_{test} \downarrow$ & $cov_{test}$ $\uparrow$ & $F1_{fake} \uparrow$ \\ \hline
    $n_t=10$  & 0.34   & 0.93 & 0.92 \\
  $n_t=25$ & 0.32  & 0.93 & 0.95 \\ 
  (Base) $n_t=50$  & 0.32  & 0.92 & 0.95 \\
  $n_t=100$  & 0.33  & 0.92 & 0.93 \\ 
  $n_t=200$ & 0.33  & 0.95 & 0.94 \\  \hline
  $n_{noise}=1$ & 0.52  & 0.86 & 0.81 \\ 
  $n_{noise}=5$ & 0.38  & 0.91 & 0.90 \\ 
  $n_{noise}=10$  & 0.37  & 0.91 & 0.93 \\ 
  $n_{noise}=25$ & 0.36  & 0.93 & 0.92 \\ 
  $n_{noise}=50$  & 0.35 & 0.93 & 0.91 \\ 
  (Base) $n_{noise}=100$ & 0.32 & 0.92 & 0.95 \\ \hline
  (Base) XGBoost & 0.32 & 0.92 & 0.95 \\
  LightGBM & 0.32 & 0.94 & 0.93 \\  
  CatBoost & 0.48 & 0.78 & 0.85 \\
  Random Forests & 0.55  & 0.78 & 0.91 \\ 
  \hline
  (Base) $y_{cond}=False$  & 0.34 & 0.92 & 0.95 \\
  $y_{cond}=True$  & 0.35 & 0.92 & 0.97 \\ 
\bottomrule 
\end{tabular}
\end{table}

\begin{table}[ht]
\caption{Forest-VP ablation on the Iris dataset (baseline: $n_t=50$, $n_{noise}=100$, $y_{cond}=False$)}
\label{tab:ablation_iris2}
\notsotiny
\centering
\begin{tabular}{r|ccc}
  \toprule
 & $W_{test} \downarrow$ & $cov_{test}$ $\uparrow$ & $F1_{fake} \uparrow$ \\ \hline
  $n_t=10$ & 0.57 & 0.56 & 0.84 \\ 
  $n_t=25$  & 0.37 & 0.88 & 0.93 \\ 
  (Base) $n_t=50$  & 0.33 & 0.93 & 0.96 \\
  $n_t=100$  & 0.34  & 0.93 & 0.95 \\ 
  $n_t=200$  & 0.35  & 0.91 & 0.95 \\  
  \hline
  $n_{noise}=1$  & 0.65  & 0.71 & 0.69 \\ 
  $n_{noise}=5$  & 0.44  & 0.89 & 0.90 \\ 
  $n_{noise}=10$  & 0.40 & 0.90 & 0.93  \\ 
  $n_{noise}=25$  & 0.35  & 0.92 & 0.94 \\ 
  $n_{noise}=50$  & 0.35  & 0.90 & 0.95 \\  
  (Base) $n_{noise}=100$  & 0.33 & 0.93 & 0.96 \\
  \hline
  (Base) XGBoost  & 0.33 & 0.93 & 0.96 \\
  LightGBM & 0.33 & 0.92 & 0.97 \\  
  CatBoost & 0.64 & 0.56 & 0.81 \\  
  Random Forests  & 0.94  & 0.34 & 0.78 \\ 
  \hline
 (Base) $y_{cond}=False$  & 0.33 & 0.93 & 0.96 \\
  $y_{cond}=True$  & 0.33 & 0.93 & 0.97 \\ 
\bottomrule 
\end{tabular}
\end{table}

\subsection{Imputation on missing tabular data}

For imputation, we randomly remove 20\% of the values in the input $X$ (Missing Completely at Random; MCAR). The outcome $y$ is left as non-missing. We compare our method to a wide range of tabular data generative models. We consider non-deep methods: $k$NN-Imputation \citep{troyanskaya2001missing}, ICE \citep{van2011mice, buck1960method}, MissForest \citep{stekhoven2012missforest}, MICE-Forest \citep{van1999multiple, wilson_miceforest_2023}, softimpute \citep{hastie2015matrix} and minibatch Sinkhorn divergence \citep{muzellec2020missing}. We also consider deep-learning methods: MIDAS \citep{lall2022midas}, and GAIN \citep{yoon2018gain}. Details on each method can be found in Appendix \ref{app:hyperparams}. 

Results are presented in Table \ref{tab:imp}. We find that the best overall methods are MICE-Forest and MissForest; both methods are nearly equally good, with the difference being that MissForest has low diversity and is thus best for single imputation. These results highlight the power of tree-based methods on tabular data. 

For the remaining methods, it is hard to rank them globally since they vary in performance across the different metrics. With raw scores, our method generally outperforms them (See Table \ref{tab:imp2} and bar plots in \S\ref{app:bar_plot_imp}). However, with rank scores, our method can rank above or below GAIN, OT, and KNN, depending on the metrics. However, we note that our method is much more diverse than other methods, making it appealing for multiple imputation strategies.

\subsection{Generation with incomplete tabular data}

For generation of incomplete data, we follow the same setup as the generation with complete data (see Section \ref{sec:gen_complete}) but randomly remove 20\% of the values in the input $X$ (MCAR). Our method can work on incomplete data directly thanks to XGBoost, so we do not impute the data. For other methods, we first impute the input data $X$ with MissForest prior to training (we could also follow this strategy for our Forest Flow and Diffusion methods). 

The averaged rank of each method is presented in Table \ref{tab:gen_miss}, their raw scores in Table \ref{tab:gen_miss2} and the bar plots in \ref{app:bar_plot_gen_miss}. Forest-Flow and Forest-VP perform similarly and are generally the highest-performing methods except on training coverage, R-squared, and coverage rate metrics, where TabDDPM is slightly better. These results show that our method performs similarly or better 
than their deep-learning counterpart without needing GPUs and by training directly on missing data instead of relying on missing data imputation.

\subsection{Ablation of Forest-VP and Forest-Flow} \label{sec:ablation}

We run an ablation on the Iris dataset when varying the hyperparameters of the algorithm and the choice of tree-based approximator (with default hyperparameters and L1/L2 penalizations disabled). The results can be found for Forest-Flow in Table \ref{tab:ablation_iris} and for Forest-Diffusion in Table \ref{tab:ablation_iris2}. For both flow and diffusion, the best performance is obtained at $n_t=50$, and the larger $n_{noise}$ is. XGBoost and LightGBM are superior to other tree-based methods. However, LightGBM freezes when running multiple models in parallel, and training not-in-parallel is much slower than training XGBoost in parallel; thus, given its significant speed advantage, we use XGBoost as our main GBT method.  We observe that conditioning on the outcome label improves performance for some metrics, but the difference is not significative.

\section{CONCLUSION}

We presented the first approach to train diffusion and flow-based models using XGBoost (and other Gradient-Boosted Tree methods) instead of neural networks. Our method generates highly realistic synthetic tabular data even when the training data contains missing values and they also generate diverse imputations. 
Our method performs better or on-par to deep-learning methods without requiring GPUs. Our work shows that tree-based methods are currently the best algorithms to deal with tabular data on data generation tasks. Our Python \citep{python3} and R \citep{Rprogramming} code is available at \url{https://github.com/SamsungSAILMontreal/ForestDiffusion}.

While powerful, our method has limitations. Although we obtain state-of-the-art results on the generation task, MICE-Forest and MissForest remain better for imputation. Furthermore, because of not using mini-batch training, the memory demand of GBTs can potentially be higher than deep-learning methods for large datasets. This limitation is further exacerbated by the fact that we duplicate the rows of the dataset $n_{noise}$ times, thus making the dataset much bigger.

As future work, we could potentially use multinomial diffusion \citep{hoogeboom2021argmax} to improve performance. We could also consider using new diffusion processes \citep{karras2022elucidating} or classifier-free guidance \citep{ho2022classifier}.  It would be helpful to find a way to train Tree models with mini-batches to remove the need to train on a duplicated dataset (similarly to the novel data iterator from XGBoost 2.0). As future applications, it would be interesting to apply this technique to data-augmentation, class imbalance, and domain translation tasks. 

\begin{figure}[t!]
    \centering
    \includegraphics[width=0.5\textwidth]{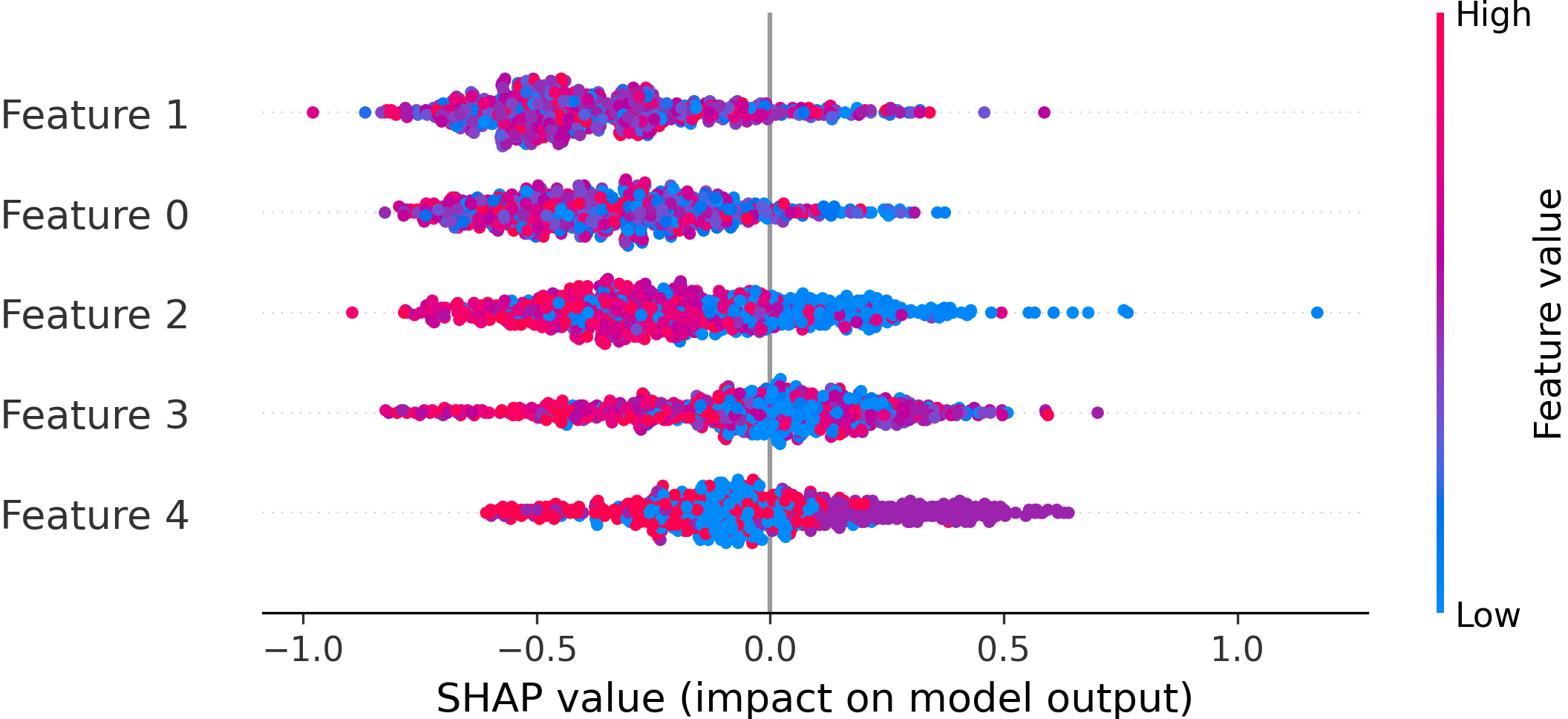}
    \caption{Average feature importance (SHAP value) across all XGBoost models trained with Forest-Flow on Iris dataset. The features (from 0 to 4) are, respectively, sepal length, sepal width, petal length, petal width, and species. Instead of average, one could also get a different plot per noise-level or variable predicted.} 
    \label{fig:fi}
\end{figure}

To better understand the data generation process, one could extract feature importance across all XGBoost models (an example is shown in Figure \ref{fig:fi} for Forest-Flow). This could also be used for variable selection. 

\section{BROADER IMPACTS}

Our method generates new observations that closely match the data's true observations. Although the synthetic samples are highly realistic, they are still generated. One should be careful about drawing inferences on fake participants, considering their potential real-world impact. 
We encourage practitioners to always compare fake data inference to real data inference.

Missing data imputation is ultimately an \emph{elaborate guess}. Imputing data is sensible when one has few missing values per participant 
or when non-missing values are highly correlated to missing values. 
Otherwise, imputation methods effectively fabricate most of the information about an observation (or person), and the practitioner may draw conclusions based on this information. We encourage practitioners to 1) use multiple imputations to average over different plausible imputations when making inferences and 2) not use imputations when they have too much missing data (get more participants or find ways to retrieve the missing data).

Perhaps one of the biggest advantages of our method is its conservative nature with respect to resources. Typical diffusion models and generative AI algorithms are extremely resource intensive, taking special dedicated hardware as well as distributed training. 
Meanwhile, our method can be run on a regular laptop (with 4-12 CPUs). This is in extreme contrast to most deep neural networks, which require expensive GPUs.


\subsubsection*{Acknowledgments}

This research was enabled in part by compute resources provided by Mila (\url{mila.quebec}). KF is supported by NSERC Discovery grant (RGPIN-2019-06512), a CIFAR AI Chair and a Samsung grant. TK is supported by the Ethereum foundation and Lineage logistics foundation grant. We thank Simon Lacoste-Julien, Quentin Bertrand and Emy Gervais for their valuable input.

\bibliographystyle{plainnat}
\bibliography{bib}

\newpage
\section*{Checklist}

 \begin{enumerate}

 \item For all models and algorithms presented, check if you include:
 \begin{enumerate}
   \item A clear description of the mathematical setting, assumptions, algorithm, and/or model. [Yes]
   \item An analysis of the properties and complexity (time, space, sample size) of any algorithm. [Yes]
   \item (Optional) Anonymized source code, with specification of all dependencies, including external libraries. [Yes]
 \end{enumerate}

 \item For any theoretical claim, check if you include:
 \begin{enumerate}
   \item Statements of the full set of assumptions of all theoretical results. [Not Applicable]
   \item Complete proofs of all theoretical results. [Not Applicable]
   \item Clear explanations of any assumptions. [Not Applicable]     
 \end{enumerate}

 \item For all figures and tables that present empirical results, check if you include:
 \begin{enumerate}
   \item The code, data, and instructions needed to reproduce the main experimental results (either in the supplemental material or as a URL). [Yes]
   \item All the training details (e.g., data splits, hyperparameters, how they were chosen). [Yes]
         \item A clear definition of the specific measure or statistics and error bars (e.g., with respect to the random seed after running experiments multiple times). [Yes]
         \item A description of the computing infrastructure used. (e.g., type of GPUs, internal cluster, or cloud provider). [Yes]
 \end{enumerate}

 \item If you are using existing assets (e.g., code, data, models) or curating/releasing new assets, check if you include:
 \begin{enumerate}
   \item Citations of the creator If your work uses existing assets. [Yes]
   \item The license information of the assets, if applicable. [Yes]
   \item New assets either in the supplemental material or as a URL, if applicable. [Not Applicable]
   \item Information about consent from data providers/curators. [Yes]
   \item Discussion of sensible content if applicable, e.g., personally identifiable information or offensive content. [Not Applicable]
 \end{enumerate}

 \item If you used crowdsourcing or conducted research with human subjects, check if you include:
 \begin{enumerate}
   \item The full text of instructions given to participants and screenshots. [Not Applicable]
   \item Descriptions of potential participant risks, with links to Institutional Review Board (IRB) approvals if applicable. [Not Applicable]
   \item The estimated hourly wage paid to participants and the total amount spent on participant compensation. [Not Applicable]
 \end{enumerate}

 \end{enumerate}


\clearpage
\appendix

\thispagestyle{empty}

\onecolumn

\aistatstitle{Supplementary material:\\Generating and Imputing Tabular Data via Diffusion and Flow-based Gradient-Boosted Trees}

The supplementary material is structured as follows:
\begin{itemize}
    \item Appendix \ref{app:sec_alg} describes the full algorithms of our methods Forest-VP and Forest-Flow.
    \item Appendix \ref{app:sec_datasets} describes the datasets, the metrics and the competitors that we used in our experiments.
    \item Appendix \ref{app:avg_res_bar_plots} reports all averaged raw scores over the 27 datasets as well as the bar plots for each dataset, metric and method.
\end{itemize}

\section{ALGORITHMS}\label{app:sec_alg}

\subsection{REPAINT Imputation Algorithm}

\begin{algorithm}[h]
    \caption{Forest-Diffusion Imputation with REPAINT}
    \label{alg:imp_repaint}
     \textbf{Input:} XGBoost models $\{f(t)\}_{t \in t_{levels}}$, dataset X of size $[n_{obs},d]$, $d$ variables, $n_{noise}=100$, $n_{t}=50$ noise levels, $r=10$ repaints, jump size $j=5$.

     $M \gets $ Mask indicating which values are non-missing in $X$ (1: non-missing, 0: missing)

     $X(1) \gets $ Dataset of $z \sim \mathcal{N}(0,1)$ with size $[n_{obs},d]$

     $n_{repaint}  \gets 1$

     $t \gets 1$
    
     \While {$t>0$}
     {

        $X(t-\frac{1}{n_t}) \gets $ Empty Dataset with size $[n_{obs},d]$

        \tcp{Reverse process for missing values}
     
        $X(t-\frac{1}{n_t})[1-M] \gets \text{Reverse}(X(t)[1-M] \mid f(t), t, n_t)$

        \tcp{Set non-missing values to truth}

        $X(t-\frac{1}{n_t})[M] \gets \text{Forward}(Z[M], X[M], 0, t \mid n_t)$

        \If {$n_{repaint} < r$ and $t+1 \hspace{-2pt} \mod j = 0$}{ 

        \tcp{go back a few steps to repaint}
        
           $n_{repaint} \gets n_{repaint} + 1$

         $Z \gets $ Dataset of $z \sim \mathcal{N}(0,1)$ with size $[n_{obs},d]$

           $X(t+\frac{j}{n_t})[M] \gets \text{Forward}(Z[M], X(t)[M], t, t+\frac{j}{n_t}, \mid n_t)$

           $t \gets t + \frac{j}{n_t}$

        }

        \If {$n_{repaint} = r$ and $n_tt+1 \hspace{-2pt} \mod j = 0$}{
           $n_{repaint} \gets 0$ \tcp{finished repainting}
        }

        $t \gets t - \frac{1}{n_t}$
     }
     \Return $X(0)$
\end{algorithm}


\subsection{Forward and Reverse step Algorithms}\label{app:forward_reverse}

\begin{algorithm}[H]
    \caption{Forward $(0 \to t)$}
    \label{alg:forward}
     \textbf{Input:} Noise data $Z$, dataset $X$, time $t$, number of noise levels $n_t$, boolean $flow$ (if True, use conditional flow matching; if False, use VP diffusion), $\beta=[0.1, 20]$.

     \If {$flow$}{
     
        $X(t) \gets (1-t)Z + t X$
        
        $Y(t) \gets X - Z$
        
    }
    \Else{
    
        $C \gets -\frac{1}{4} t^{2} (\beta[1] - \beta[0]) - 0.5 t \beta[0]$
        
        $X(t) \gets \exp{C} X + \sqrt{1-\exp{2C}} Z$
        
        $Y(t) \gets Z$
        
    }

    \Return $X(t), Y(t)$
    
\end{algorithm}

\begin{algorithm}[H]
    \caption{Forward $(t \to t_{next})$}
    \label{alg:forward_next}
     \textbf{Input:} Noise data $Z$, dataset $X$, next time $t_{next}$ (where $t_{next}$ > $t$) , current time $t$, number of noise levels $n_t$, boolean $flow$ (if True, use conditional flow matching; if False, use VP diffusion), $\beta=[0.1, 20]$.

     $h \gets t_{next} - t$

     \If {$flow$}{
     
        $X(t_{next}) \gets Z + h(X - Z)$
        
    }
    \Else{
    
        $\beta_t = \beta[0] + t (\beta[1] - \beta[0])$
        
        $X(t_{next}) \gets X(t) - h \frac{1}{2} \beta_t X(t) + \sqrt{\beta_t} Z$
        
    }

    \Return $X(t_{next})$
    
\end{algorithm}

\begin{algorithm}[H]
    \caption{Reverse}
    \label{alg:reverse}
     \textbf{Input:} dataset $X(t)$, time $t$, XGBoost model $f(t)$, number of noise levels $n_t$, boolean $flow$ (if True, use conditional flow matching; if False, use VP diffusion), $\beta=[0.1, 20]$.

    $h \gets \frac{1}{n_t}$

     \If {$flow$}{
     
        $X(t-\frac{1}{n_t}) \gets X(t) + h f(t)$

    }
    \Else{

        $\beta_t = \beta[0] + t (\beta[1] - \beta[0])$
    
        $C \gets -\frac{1}{4} t^{2} (\beta[1] - \beta[0]) - 0.5 t \beta[0]$

        $score \gets -\frac{f(t)}{\sqrt{1-\exp{2C}}}$

        $\mu = -\frac{1}{2}\beta_t X(t) - \beta_t score$
        
        $X(t-\frac{1}{n_t}) \gets X(t) - h \mu + \beta_t \sqrt{h} Z$
        
    }

    \Return $X(t-\frac{1}{n_t})$
    
\end{algorithm}

\subsection{Flow matching} \label{app:fm_euc}
\label{app:flow_matching_rd}

Flow matching has been introduced by several works under different names \cite{lipman_flow_2022, albergo2023building, liu2022flow}. As explained in the main text, Flow Matching's objective is to regress a conditional vector field built from conditional probability paths. In this section, we provide more details about the conditional probability paths and conditional vector fields that were used respectively in \cite{lipman_flow_2022} and \cite{tong2023improving}.

The natural choice of the conditional probability path $p_t(x|z)$ is a Gaussian conditioned on a latent variable $z \sim q(z)$ with variance $\sigma_t$ leading to $p_t(x) = \int \mathcal{N}(\nu_t(z), \sigma_t) q(z) dz$. In particular, we want that $p_1$ approximates the distribution $q$. When we use a Gaussian for the conditional probability path, it is possible to compute the conditional vector field in closed-form thanks to the next Theorem:

\begin{theorem}[Theorem 3 of \cite{lipman_flow_2022}]\label{thm:gaussian_flow}
The unique vector field whose integration map satisfies $\rho_t(x) = \nu_t + \sigma_t x$ has the form
\begin{equation}\label{eq:gaussian_ut}
\mu_t(s) = \frac{\sigma_t'}{\sigma_t} (s - \nu_t) + \nu_t',
\end{equation}
\end{theorem}

Therefore to develop a conditional flow matching variant, we have to choose four different quantities: the conditioning $z$, the Gaussian mean $\nu_t$, the Gaussian standard deviation $\sigma_t$ and the latent distribution $q$.

{\bfseries Flow Matching.} We first consider the condition $z$ to be a single training sample $z = x_1$, the Gaussian mean is $\nu_t(x_1) = tx_1$ and its standard deviation $\sigma_t = (t \sigma - t + 1)^2$ where $\sigma>0$. Regarding $q$, we set it to $q(z)=q(x_1)$ as the uniform distribution over the training dataset. Now we define the conditional probability path and the conditional vector field as:
\begin{align}\label{eq: fm_density_vectorfield}
p_t(x | z) &= \mathcal{N}(t x_1, (t \sigma - t + 1)^2), \\
\mu_t(x | z) &= \frac{x_1 - (1 - \sigma) x}{1 - (1 - \sigma) t},
\end{align}
which is a probability path from the standard normal distribution ($p_0(x|z)=\mathcal{N}(x;0,1)$) to a Gaussian distribution centered at $x_1$ with standard deviation $\sigma$ ($p_1(x|z)=\mathcal{N}(x;x_1,\sigma^2)$) in order to ensure that $p_1 \approx q$. 

{\bfseries Independent-Conditional Flow Matching.} Another recent Flow Matching variant is \emph{Independent-Conditional Flow Matching}. In this variant, the condition $z$ is a tuple of a source and a target sample $z=(x_0, x_1)$, the mean of Gaussian conditional probability path is set to $\nu_t(x_0, x_1) = t x_1 + (1-t) x_0$, the standard deviation $\sigma_t$ to a constant independent of $t$ and the latent distribution to the independent coupling $q(x_0, x_1) = q(x_0) q(x_1)$.

\begin{align}
p_t(x | x_0, x_1) &= \mathcal{N}(x\mid t x_1 + (1 - t) x_0, \sigma^2),\label{eq:cfm:ptc}\\
p_t(x) &= \int \mathcal{N}(x\mid t x_1 + (1 - t) x_0, \sigma^2) \pi(x_0, x_1) dx_0 dx_1,\label{eq:cfm:pt}\\
\mu_t(x | x_0, x_1) &= x_1 - x_0. \label{eq:cfm:ut}
\end{align}

We note that we also recover $p_1(x|z)=\mathcal{N}(x;x_1,\sigma^2)$ as the above Flow Matching variant. The I-CFM loss is equal to $\mathcal{L}_{\text{Forest-Flow}} = \|v_\theta(t, x) - \mu_t(x|z)\| = \|v_\theta(t, x) - (x_1 - x_0)\|$. The advantage of this variant is that it is simple to implement and available in open-source in the TorchCFM package\footnote{\url{https://github.com/atong01/conditional-flow-matching}} \cite{tong2023improving}. Note that in our experiments, we have set $\sigma$ to 0.

\subsection{Why Flow methods cannot be used for imputation or inpainting}\label{app:inpaint_flow}

We considered doing imputation for flow models through regular inpainting or REPAINT. However, we found that it could not work due to the deterministic nature of the method, as explained below.

A flow is such that a single Gaussian sample $x(1)=z$ (of dimension $d$) follows a deterministic trajectory leading to a single data sample $x(0)$. For imputation, we need the final $x(0)$ to have the same non-missing values as the true sample containing missing data. However, since the trajectory from $x(1)$ to $x(0)$ is deterministic, we have no control over which data sample we will end up with at $t=0$. Trying to force $x(t)$ into the direction that we want (toward the true non-missing values) does not work because we diverge from the actual deterministic trajectory. Therefore, for imputation, we had to rely exclusively on diffusion methods.

\section{DATASETS, METRICS AND METHODS}\label{app:sec_datasets}
\subsection{Datasets}\label{app:datasets}

We list all the datasets in Table \ref{tab:datasets}. We mostly use the same datasets as \citet{muzellec2020missing} (with the exception of \emph{bean}), and the datasets come either from the UCI Machine Learning Repository \citep{UCI} or scikit-learn \citep{scikit-learn}. All UCI datasets are licensed under the Creative Commons Attribution 4.0 International license (CC BY 4.0). For scikit-learn, the Iris dataset is licensed with the BSD 3-Clause License, and the California housing is freely provided by the authors with no license. Irrespective of the source, all datasets are openly shared with no restriction on usage.

\begin{table}[htp]
\caption{Datasets}
\label{tab:datasets}
\centering
\begin{tabular}{ccccc}
\toprule
 Dataset & Citation & $n$ &   $d$ & Outcome \\
\midrule
        airfoil self noise & \citep{misc_airfoil_selfnoise_291} &   1503 &   5 & continuous \\
        bean & \citep{koklu2020multiclass} & 13611 & 16 & categorical \\
          blood transfusion & \citep{misc_blood_transfusion_service_center_176} &    748 &   4 & binary\\
  breast cancer diagnostic & \citep{misc_breast_cancer_wisconsin_diagnostic_17} &    569 &  30 & binary \\
                  california housing & \citep{pace1997sparse} &  20640 &   8 & continuous \\
     climate model crashes & \citep{misc_climate_model_simulation_crashes_252} &    540 &  18 & binary \\
       concrete compression & \citep{misc_concrete_compressive_strength_165} &   1030 &   7 &continuous \\
             concrete slump & \citep{misc_concrete_slump_test_182} &    103 &   7 & continuous \\
 connectionist bench sonar & \citep{misc_connectionist_bench_sonar_mines_vs_rocks_151} &    208 &  60 &  binary \\
 connectionist bench vowel & \citep{misc_connectionist_bench_vowel_recognition__deterding_data_152} &    990 &  10 &  binary \\
                       ecoli & \citep{misc_ecoli_39} &    336 &   7  & categorical \\
                       glass & \citep{misc_glass_identification_42} &    214 &   9 & categorical \\
                  ionosphere & \citep{misc_ionosphere_52} &    351 &  34 & binary \\
                        iris & \citep{misc_iris_53} &    150 &   4 & categorical \\
                      libras & \citep{misc_libras_movement_181} &    360 &  90  & categorical \\
                  parkinsons & \citep{misc_parkinsons_174} &    195 &  23  & binary \\
             planning relax & \citep{misc_planning_relax_230} &    182 &  12  & binary \\
        qsar biodegradation & \citep{misc_qsar_biodegradation_254} &   1055 &  41  & binary \\
                       seeds & \citep{misc_seeds_236} &    210 &   7    & categorical \\
                        wine & \citep{misc_wine_109} &    178 &  13    & categorical \\
          wine quality red & \citep{misc_wine_quality_186} &   1599 &  10  & integer \\
        wine quality white & \citep{misc_wine_quality_186} &   4898 &  11 & integer \\
        yacht hydrodynamics & \citep{misc_yacht_hydrodynamics_243} &    308 &   6 & continuous \\
                       yeast & \citep{misc_yeast_110} &   1484 &   8 & categorical \\
   tic-tac-toe & \citep{misc_tic-tac-toe_endgame_101} &   958 &   9 & binary \\
   congressional voting & \citep{misc_congressional_voting_records_105} &   435 &   16 & binary \\
   car evaluation & \citep{misc_car_evaluation_19} &   1728 &   6 & categorical \\
\bottomrule
\end{tabular}
\end{table}

\subsection{Metrics}\label{app:metrics}

We describe below the choices of metrics used in this paper.

\subsubsection{Gower distance}

Since we have mixed-type data (continuous and categorical variables) at varying scales (one variable can be thousands and one in decimals), the Euclidean distance is not an adequate distance measure. We take inspiration from the k-Nearest Neighbors (KNN) literature by instead relying on the Gower distance \citep{gower1971general}, a popular metric to uniformize distances between continuous and categorical variables. The Gower distance is the sum of the per-variable costs, and these costs are defined to always be between 0 and 1 for both continuous and categorical variables. We use this distance for every evaluation metric that relies on a distance metric (i.e., Wasserstein distance, coverage, and MAE). The Gower distance can be implemented by taking the L1 distance over min-max normalized continuous variables or one-hot categorical variables divided by two. 

\subsubsection{Imputation}

\paragraph{Wasserstein Distance}

For imputation, to assess how close the real and imputed data distributions are, we use the Wasserstein distance. We report the Wasserstein distance to either the training data or testing data. Since calculating the Wasserstein distance scales quadratically with sample size, to the reduce time taken, we only calculate the distance for datasets with less than 5000 training samples. This approach is also chosen by \citet{muzellec2020missing}.

\paragraph{Mean Absolute Deviation around the median/mode (MAD)}

To assess the diversity of the imputations, we look at the Mean Absolute Deviation (MAD) around the median (for continuous variables) or mode (for categorical variables) of imputed values across $k=5$ imputations. MAD measures the variability using the L1 distance to the median/mode, which is more sensible than using the variance given our reliance on the Gower Distance, which is L1-based. This measure says nothing about quality, but given equal levels of quality, one should prefer a method that is more diverse in its imputations to avoid creating false precision and ensure valid uncertainty on parameter estimates \citep{li2015multiple}.

\paragraph{Minimum and Average MAE}

A common approach to assess the distance between the imputed values and the ground-truth values is the root mean squared error (RMSE) or mean absolute error (MAE) \citep{stekhoven2012missforest}. However, the usefulness of these metrics has been severely debated \citep{van2018flexible, wang2022deep}. An imputation method that accounts for uncertainty by producing multiple different imputations will inherently have a higher RMSE/MAE than one centered across the mean or median of possible imputations. We see this in \citet{tashiro2021csdi}, where instead of directly taking the multiple imputations they generate, the authors only report the median imputation of their 100 different imputations to be as close as possible to the ground truth and thus obtain a great RMSE/MAE; however, in doing so, they destroy all the sampling diversity produced by their method. Furthermore, this approach is only usable when the number of variables is 1, given that no exact concept of multivariate median exists.

Since taking the median approach of \citet{tashiro2021csdi} cannot work in the multivariate setting, we adjust the metric through the standard approach used in the video prediction literature \citep{denton2018stochastic, castrejon2019improved,franceschi2020stochastic, voleti2022mcvd}, which is to produce different samples ($k=5$ different imputations for a given missing observation in our case) and take the distance between the ground truth and the sample closest to this ground truth. The idea is that different imputations may still be plausible and valid even though they differ from the ground truth. Hence, this approach only tries to ensure that the ground truth is at least close to one of the possible imputations (rather than to all the possible imputations).

While the average MAE is unfair against stochastic methods, taking the minimum may be considered unjust against deterministic methods (because it gives more chance for the stochastic method to be closer by increasing the number of imputations $k$). Thus, we report both the minimum and the average MAE. We use the MAE instead of the RMSE because it matches the Gower Distance, which relies on the L1 distance.

\paragraph{Efficiency}

To assess the practical use for machine learning purposes, we use the efficiency/utility \citep{xu2019modeling}; this measure is defined as the average F1-score for classification problems and $R^2$ for regression problems for models trained on imputed data and evaluated on the test set. 

We average the efficiency over four useful non-deep models: linear/logistic regression, AdaBoost, Random Forests, and XGBoost. The choice of models used is a matter of taste. \citet{kotelnikov2023tabddpm} include CatBoost instead of XGBoost.
\citet{xu2019modeling} do not include any Gradient-Boosted Trees (GBTs), but they include both Decision Trees (DTs) and Multilayer perceptrons (MLPs). We prefer not to use DTs because a single tree has limited capacity, and almost no one uses them in practice (over GBTs or Random Forests). We also prefer not to use MLPs since data scientists and statisticians rarely use them for tabular data given their massive complexity, computational expense (GPUs), and generally inferior performance \citep{shwartz2022tabular}. 

\paragraph{Statistical measures}

To assess the ability to obtain statistically valid inferences from incomplete data, we rely on classic statistical measures which train a linear regression model and compare the parameter estimates obtained by the imputed data to the ones obtained through the ground-truth data with no missing values \citep{van2018flexible}. The metrics we use are: the percent bias ($| \mathbb{E}\left[ \frac{\hat{\beta}-\beta}{\beta} \right] |$), and the coverage rate (i.e., the proportion of confidence intervals containing the true value).

We report the average of those measures across all regression parameters. These metrics focus on ensuring that the imputed data do not lead to wrong incorrect assessments about the importance of each variable during inference. For example, if one seeks to know if a certain gene influence a certain outcome, one needs to make sure the imputed data does not incorrectly give a significant effect for this gene when there isn't one (or the converse).

\subsubsection{Generation}

\paragraph{Wasserstein Distance}

For generation, to assess how close the real and fake data distributions are, we use the Wasserstein distance. We report the Wasserstein distance to either the training data or testing data. Since calculating the Wasserstein distance scales quadratically with sample size, to the reduce time taken, we only calculate the distance for datasets with less than 5000 training samples. This approach is also taken by \citet{muzellec2020missing}.

\paragraph{Coverage}

To assess the diversity of generated samples, we use the coverage \citep{naeem2020reliable}, a measure of the ratio of real observations that have at least one fake observation within a sphere of radius $r$, where $r$ is the distance between the sample and its $k$-th nearest neighbor. The number of nearest neighbors $k$ is the smallest value such that we obtain at least 95\% coverage on the true data. We report the coverage for both training and testing data separately.

\paragraph{Efficiency}

We again use the average efficiency when training on synthetic data to assess the practical use for machine learning purposes.

\paragraph{Discriminator}

We follow the same design as the Efficiency F1 score metric but use a classifier to predict if a sample is real or fake. We test the classifier on fake data and report the F1 score. A high F1 score means the classifier recognizes the fake data as fake; thus, lower values correspond to better generation quality. Note that we train with the training dataset and fake data of the same size but test on another random sample of fake data to counter overfitting.

\paragraph{Statistical measures}

We use the same statistical measures while using datasets filled with fake samples instead of imputed datasets.

\subsection{Details of each methods} \label{app:hyperparams}

\subsubsection{Computational resources used}

We trained the tree-based models on a cluster of 10-20 CPUs with 64-256Gb of RAM.

We trained the other models on a cluster with 8 CPUs, 1 GPU, and 48-128Gb of RAM.

\subsubsection{Generation}

\paragraph{Oracle}

This is the actual training data.

\paragraph{ForestFlow and ForestVP}

We use $n_t=50$, $n_{noise}=100$, label conditioning, and dummy encoding of the categorical variables.

\paragraph{GaussianCopula}

GaussianCopula transforms the data into a copula, a distribution on the unit cube \citep{sklar1959fonctions}. Then, it can generate new data by sampling from the unit cube and reversing the copula. We use the Gaussian Copula implementation from the Synthetic data vault \citep{SDV} with the default hyperparameters.

\paragraph{TVAE and CTGAN}

Tabular VAE (TVAE) is a Variational AutoEncoder (VAE) \citep{kingma2013auto} for mixed-type tabular data generation.  Conditional Tabular GAN (CTGAN) is a Generative Adversarial Network (GAN) \citep{goodfellow2014generative} for mixed-type tabular data generation. Both TVAE and CTGAN were introduced by \citet{xu2019modeling}. We use the implementation from the Synthetic data vault \citep{SDV} with the default hyperparameters.

\paragraph{CTAB-GAN+}

CTAB-GAN \citep{zhao2021ctab} is a Generative Adversarial Network (GAN) \citep{goodfellow2014generative} for mixed-type tabular data generation. CTAB-GAN+ is an improvement of the method \citep{zhao2022ctab}. We use the official implementation with the default hyperparameters.

\paragraph{STaSy}

Score-based Tabular data Synthesis (STaSy) \citep{kim2022stasy} is a method using score-based generative models \citep{song_score-based_2021} for mixed-type tabular data generation. We use the official implementation. We initially obtained nonsensical outputs (lots of variables at the min or max value and/or generating only one class) on small data generation (such as with the Iris dataset) with STaSy default hyperparameters . Through correspondence with the authors, we got STaSy to work better through small changes in the hyperparameters (using Naive STaSy, 10000 epochs, and reducing the hidden dimensions to $(64, 128, 256, 128, 64)$). We use the default hyperparameters with those changes.

\paragraph{Tab-DDPM}

Tab-DDPM \citep{kotelnikov2023tabddpm} is a method using denoising diffusion probabilistic models \citep{ho_denoising_2020} for mixed-type tabular data generation. We use the official implementation with the default hyperparameters from the California dataset.

\subsubsection{Imputation}

\paragraph{Oracle}

This is the actual training data before adding missing values.

\paragraph{Forest-Flow and Forest-VP}

We use $n_t=50$, $n_{noise}=100$ (except for the \emph{bean}, where we use $n_{noise}=50$ to reduce memory requirements), and dummy encoding of the categorical variables.

\paragraph{KNN}

We use the $k$-nearest neighbors (KNN) \citep{cover1967nearest} based imputation method by  \citet{troyanskaya2001missing}. We use the scikit-learn \citep{scikit-learn} implementation with $k=1$ and default hyperparameters. There is no L1-based KNN imputer to our knowledge, so we cannot work with the Gower Distance. Thus, we stick to the regular KNN imputer, which depends on a special Euclidean distance made for incomplete data that penalizes distance based on the number of missing values. Since the distance is L2-based, we standardize (z-score) all the variables prior to being imputed and unstandardize them after imputation.

\paragraph{ICE}

Imputation by Chained Equations (ICE) does iterative imputations through conditional expectation. We use the IterativeImputer function from scikit-learn \citep{scikit-learn}, which is based on \citet{van2011mice}. We use 10 iterations and the default hyperparameters. We clip the imputations to be between the minimum and maximum of observed values.

\paragraph{MICE-Forest}

MICE-Forest (also called miceRanger) \citep{wilson_miceforest_2023} is an imputation method using Multiple Imputations by Chained Equations (MICE) \citep{van2011mice} with predictive mean matching \citep{little1988missing} and LightGBM \citep{ke2017lightgbm}, a popular type of Gradient-Boosted Tree (GBT) method. We use the official Python library with the default hyperparameters.

\paragraph{MissForest}

MissForest \citep{stekhoven2012missforest} is an iterative algorithm using Random Forests \citep{breiman2001random} to impute missing data. We use the implementation from the \emph{missingpy} Python library \citep{missingpy} with the default hyperparameters.

\paragraph{Softimpute}

Softimpute \citep{hastie2015matrix} is an iterative soft-threshold Singular Value Decomposition (SVD) method to impute missing data. We use the implementation from \citet{muzellec2020missing} with the default hyperparameters. We clip the imputations to be between the minimum and maximum of observed values.

\paragraph{Sinkhorn}

Minibatch Sinkhorn divergence \citep{muzellec2020missing} is an Optimal Transport (OT) \citep{villani2009optimal} method based on the minibatch Sinkhorn divergence \citep{cuturi2013sinkhorn, genevay18a} to impute missing data. The intuition is that two minibatches from the same distribution should have relatively similar statistics. Therefore, they leverage the minibatch optimal transport \cite{fatras20a, fatras2021minibatch} loss to imput missing data. We use the official implementation with the default hyperparameters. We clip the imputations to be between the minimum and maximum of observed values.

\paragraph{GAIN}

Generative Adversarial Imputation Nets (GAIN) \citep{yoon2018gain} is a GAN \citep{goodfellow2014generative} based method to impute missing data. We use the official implementation with the default hyperparameters. The code does rounding in a problematic way; it assumes that variables with less than 20 unique values are categorical, and those variables are rounded to the nearest integer. We found this to cause problems, so we instead provide the names of the categorical variables to the function so that only those variables are rounded.

\section{AVERAGED RESULSTS AND BAR PLOTS}\label{app:avg_res_bar_plots}

In this section, we  provide the averaged raw score for each method on all datasets as well as the different bar plot for each metric and each method on each dataset.

\subsection{Tables - raw score} \label{app:tables}
In this section, we provide the averaged raw score for each method on all datasets for the generation of tabular data \ref{tab:gen2}, the tabular data imputation \ref{tab:imp2} and the tabular data generation with missing values \ref{tab:gen_miss2}.
\begin{table*}[ht]
\caption{\small{Tabular data generation with complete data (27 datasets, 3 experiments per dataset); raw score - mean (standard-error)}}
\label{tab:gen2}
\supertiny
\centering
\begin{tabular}{r|ll|ll|lll|ll}
  \toprule
 & $W_{train} \downarrow$ & $W_{test} \downarrow$ & $cov_{train}$ $\uparrow$ & $cov_{test}$ $\uparrow$ & $R^2_{fake} \uparrow$ & $F1_{fake} \uparrow$ & $F1_{disc} \downarrow$ & $P_{bias} \downarrow$ & $cov_{rate} \uparrow$ \\ \hline
   GaussianCopula & 2.74 (0.56) & 2.99 (0.61) & 0.18 (0.04) & 0.37 (0.06) & \phantom{-}0.20 (0.14) & 0.46 (0.06) & 0.62 (0.05) & 2.27 (0.77) & 0.23 (0.12) \\ 
  TVAE & 2.12 (0.58) & 2.35 (0.63) & 0.33 (0.04) & 0.63 (0.04) & -0.47 (0.61) & 0.52 (0.08) & 0.44 (0.01) & 4.15 (1.97) & 0.26 (0.09) \\ 
  CTGAN & 3.58 (0.99) & 3.74 (1.01) & 0.12 (0.03) & 0.28 (0.04) & -0.43 (0.08) & 0.35 (0.04) & 0.48 (0.01) & 2.48 (1.30) & 0.20 (0.08) \\ 
  CTAB-GAN+ & 2.71 (0.81) & 2.89 (0.83) & 0.22 (0.04) & 0.44 (0.05) & \phantom{-}0.05 (0.12) & 0.44 (0.05) & 0.52 (0.02) & 2.95 (1.04) & 0.26 (0.07) \\ 
  STaSy & 3.41 (1.39) & 3.66 (1.42) & 0.38 (0.05) & 0.63 (0.05) & -4.21 (4.44) & 0.61 (0.06) & 0.46 (0.02) & 1.23 (0.44) & 0.45 (0.12) \\ 
  TabDDPM & 4.27 (1.89) & 4.79 (1.89) & 0.76 (0.06) & 0.80 (0.06) & \phantom{-}0.60 (0.11) & 0.66 (0.06) & 0.39 (0.03) & 0.76 (0.28) & 0.72 (0.11) \\ 
  Forest-VP & 1.46 (0.40) & 1.94 (0.50) & 0.67 (0.05) & 0.84 (0.03) & \phantom{-}0.55 (0.10) & 0.73 (0.04) & 0.39 (0.01) & 0.94 (0.30) & 0.52 (0.15) \\ 
  Forest-Flow & 1.36 (0.39) & 1.90 (0.5) & 0.83 (0.03) & 0.90 (0.03) & \phantom{-}0.57 (0.11) & 0.73 (0.04) & 0.38 (0.01) & 0.83 (0.23) & 0.63 (0.11)  \\ 
  Oracle & 0.00 (0.00) & 1.81 (0.47) & 0.99 (0.01) & 0.91 (0.04) & \phantom{-}0.64 (0.09) & 0.77 (0.04) & NA & 0.00 (0.00) & 1.00 (0.00) \\ 
   \hline

\bottomrule 
\end{tabular}
\end{table*}

\begin{table*}[ht]
\caption{Tabular data imputation (27 datasets, 3 experiments per dataset, 10 imputations per experiment) with 20\% missing values; raw score - mean (standard-error)}
\label{tab:imp2}
\supertiny
\centering
\begin{tabular}{rllll|l|ll|ll}
  \toprule
 & MinMAE $\downarrow$ & AvgMAE $\downarrow$ & $W_{train} \downarrow$ & $W_{test} \downarrow$ & MAD $\uparrow$ & $R^2_{imp} \uparrow$ & $F1_{imp} \uparrow$ & $P_{bias} \downarrow$ & $Cov_{rate} \uparrow$   \\ \hline
KNN & 0.16 (0.03) & 0.16 (0.03) & 0.42 (0.08) & 1.89 (0.49) & 0.00 (0.00) & 0.59 (0.09) & 0.75 (0.04) & 1.27 (0.25) & 0.40 (0.11) \\ 
  ICE & 0.10 (0.01) & 0.21 (0.03) & 0.52 (0.09) & 1.99 (0.49) & 0.69 (0.10) & 0.59 (0.09) & 0.74 (0.04) & 1.05 (0.29) & 0.39 (0.09) \\ 
  MICE-Forest & 0.08 (0.02) & 0.13 (0.03) & 0.34 (0.07) & 1.86 (0.48) & 0.29 (0.08) & 0.61 (0.10) & 0.76 (0.04) & 0.61 (0.20) & 0.75 (0.11) \\ 
  MissForest & 0.10 (0.03) & 0.12 (0.03) & 0.32 (0.07) & 1.85 (0.48) & 0.10 (0.03) & 0.61 (0.10) & 0.76 (0.04) & 0.62 (0.22) & 0.79 (0.08) \\ 
  Softimpute & 0.22 (0.03) & 0.22 (0.03) & 0.53 (0.07) & 1.99 (0.48) & 0.00 (0.00) & 0.58 (0.09) & 0.74 (0.04) & 1.18 (0.34) & 0.31 (0.09) \\ 
  OT & 0.14 (0.02) & 0.19 (0.03) & 0.56 (0.10) & 1.98 (0.49) & 0.28 (0.05) & 0.59 (0.10) & 0.75 (0.04) & 1.09 (0.27) & 0.39 (0.12) \\ 
  GAIN & 0.16 (0.03) & 0.17 (0.03) & 0.49 (0.11) & 1.95 (0.51) & 0.01 (0.00) & 0.60 (0.10) & 0.75 (0.04) & 1.04 (0.25) & 0.54 (0.12) \\ 
  Forest-VP & 0.14 (0.04) & 0.17 (0.03) & 0.55 (0.13) & 1.96 (0.50) & 0.25 (0.03) & 0.61 (0.10) & 0.74 (0.04) & 0.81 (0.25) & 0.57 (0.14) \\ 
  Oracle & 0.00 (0.00) & 0.00 (0.00) & 0.00 (0.00) & 1.87 (0.49) & 0.00 (0.00) & 0.64 (0.09) & 0.78 (0.04) & 0.00 (0.00) & 1.00 (0.00) \\ 
\bottomrule 
\end{tabular}
\end{table*}

\begin{table*}[ht]
\caption{\small{Tabular data generation with incomplete data (27 datasets, 3 experiments per dataset, 20\% missing values), MissForest is used to impute missing data except in Forest-VP and Forest-Flow; raw score - mean (standard-error)}}
\label{tab:gen_miss2}
\supertiny
\centering
\begin{tabular}{r|ll|ll|lll|ll}
  \toprule
 & $W_{train} \downarrow$ & $W_{test} \downarrow$ & $cov_{train}$ $\uparrow$ & $cov_{test}$ $\uparrow$ & $R^2_{fake} \uparrow$ & $F1_{fake} \uparrow$ & $F1_{disc} \uparrow$ & $P_{bias} \downarrow$ & $cov_{rate} \uparrow$ \\ \hline
GaussianCopula & 2.60 (0.58) & 2.86 (0.63) & 0.20 (0.04) & 0.43 (0.05) & \phantom{-}0.20 (0.18) & 0.48 (0.06) & 0.60 (0.04) & 2.31 (1.00) & 0.21 (0.08) \\ 
  TVAE & 2.17 (0.60) & 2.40 (0.65) & 0.32 (0.04) & 0.63 (0.04) & -0.66 (0.95) & 0.55 (0.08) & 0.45 (0.01) & 4.04 (2.30) & 0.29 (0.09) \\ 
  CTGAN & 3.64 (1.02) & 3.79 (1.04) & 0.12 (0.03) & 0.28 (0.05) & -0.34 (0.14) & 0.36 (0.04) & 0.48 (0.01) & 2.47 (1.24) & 0.20 (0.06) \\ 
  CTABGAN & 2.76 (0.83) & 2.95 (0.86) & 0.23 (0.04) & 0.45 (0.05) & \phantom{-}0.08 (0.12) & 0.45 (0.05) & 0.50 (0.02) & 2.10 (0.58) & 0.25 (0.06) \\ 
  Stasy & 3.40 (1.37) & 3.67 (1.4) & 0.38 (0.05) & 0.63 (0.06) & \phantom{-}0.27 (0.28) & 0.64 (0.06) & 0.46 (0.02) & 1.09 (0.22) & 0.36 (0.10) \\ 
  TabDDPM & 4.36 (1.89) & 4.80 (1.90) & 0.72 (0.06) & 0.78 (0.06) & \phantom{-}0.58 (0.11) & 0.67 (0.06) & 0.42 (0.03) & 1.16 (0.35) & 0.56 (0.10) \\ 
  Forest-VP & 1.84 (0.51) & 2.14 (0.56) & 0.53 (0.04) & 0.78 (0.03) & \phantom{-}0.53 (0.10) & 0.71 (0.04) & 0.42 (0.01) & 1.16 (0.30) & 0.43 (0.12) \\ 
  Forest-Flow & 1.82 (0.51) & 2.12 (0.56) & 0.67 (0.03) & 0.84 (0.03) & \phantom{-}0.55 (0.11) & 0.69 (0.04) & 0.43 (0.01) & 1.16 (0.32) & 0.50 (0.10) \\ 
  Oracle & 0.00 (0.00) & 1.87 (0.49) & 0.99 (0.01) & 0.91 (0.04) & \phantom{-}0.64 (0.09) & 0.78 (0.04) & NA & 0.00 (0.00) & 1.00 (0.00) \\ 
   \hline
\end{tabular}
\end{table*}

\subsection{Bar plots}\label{app:bar_plots}
In this section, we provide the bar plots for each method for all datasets and all metrics for the generation of tabular data \ref{app:bar_plot_gen}, the tabular data imputation \ref{app:bar_plot_imp} and the tabular data generation with missing values \ref{app:bar_plot_gen_miss}.

Please ignore the $F1_{disc}$ results for Oracle; there is not proper way to train and test on different training data, so the values are incorrect.

\subsubsection{Bar plots for generation with complete data}\label{app:bar_plot_gen}

\begin{figure*}[ht]
    \centering
    \includegraphics[width=1\textwidth]{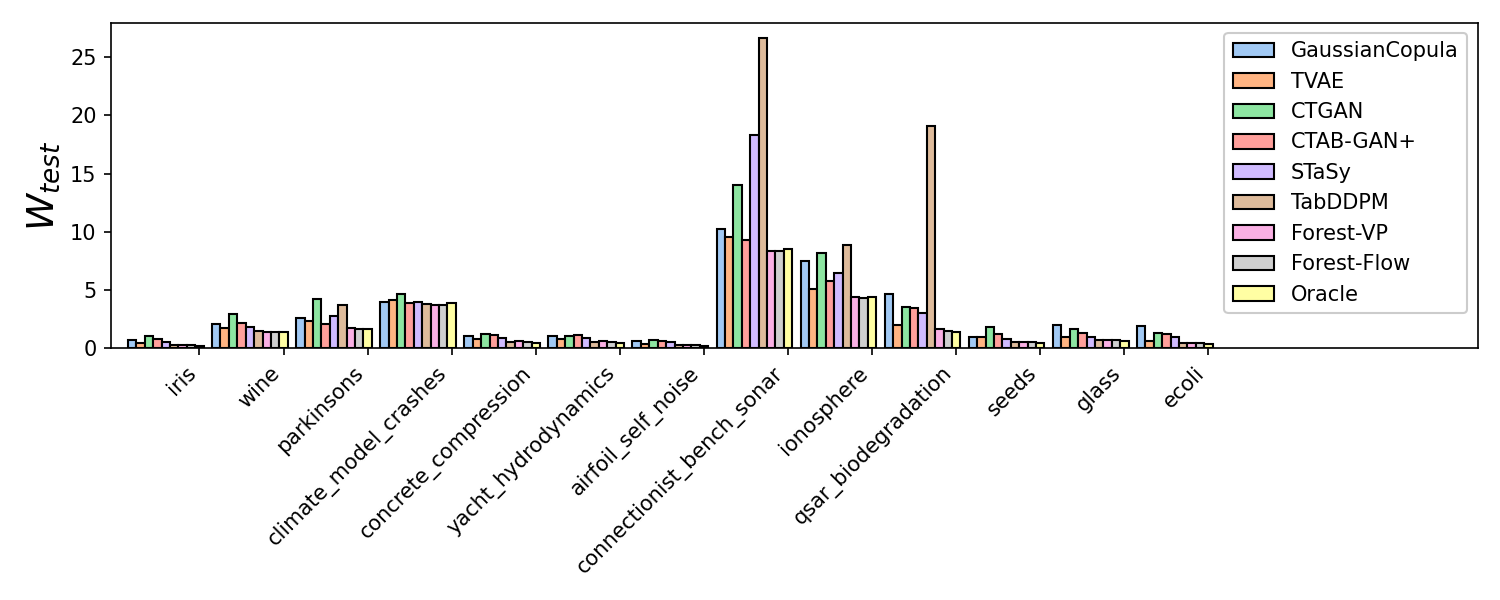}
\end{figure*}
\begin{figure*}[ht]
    \centering
    \includegraphics[width=1\textwidth]{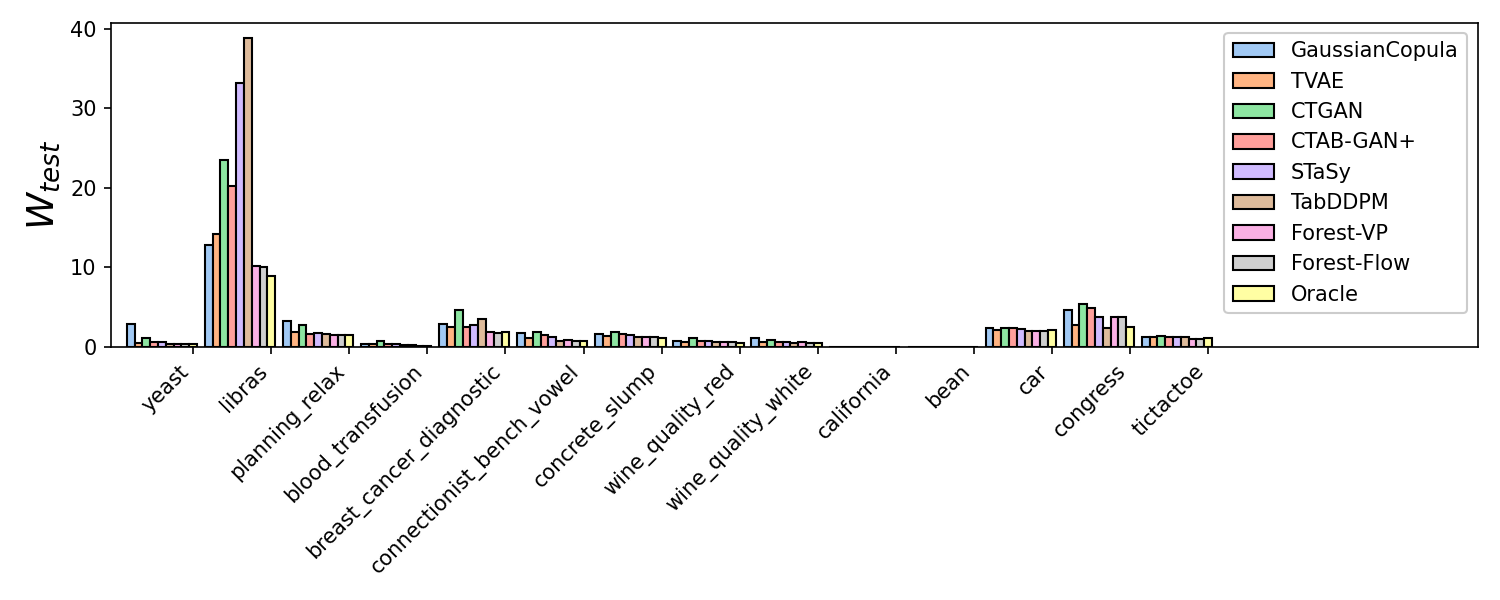}
\end{figure*}
\begin{figure*}[ht]
    \centering
    \includegraphics[width=1\textwidth]{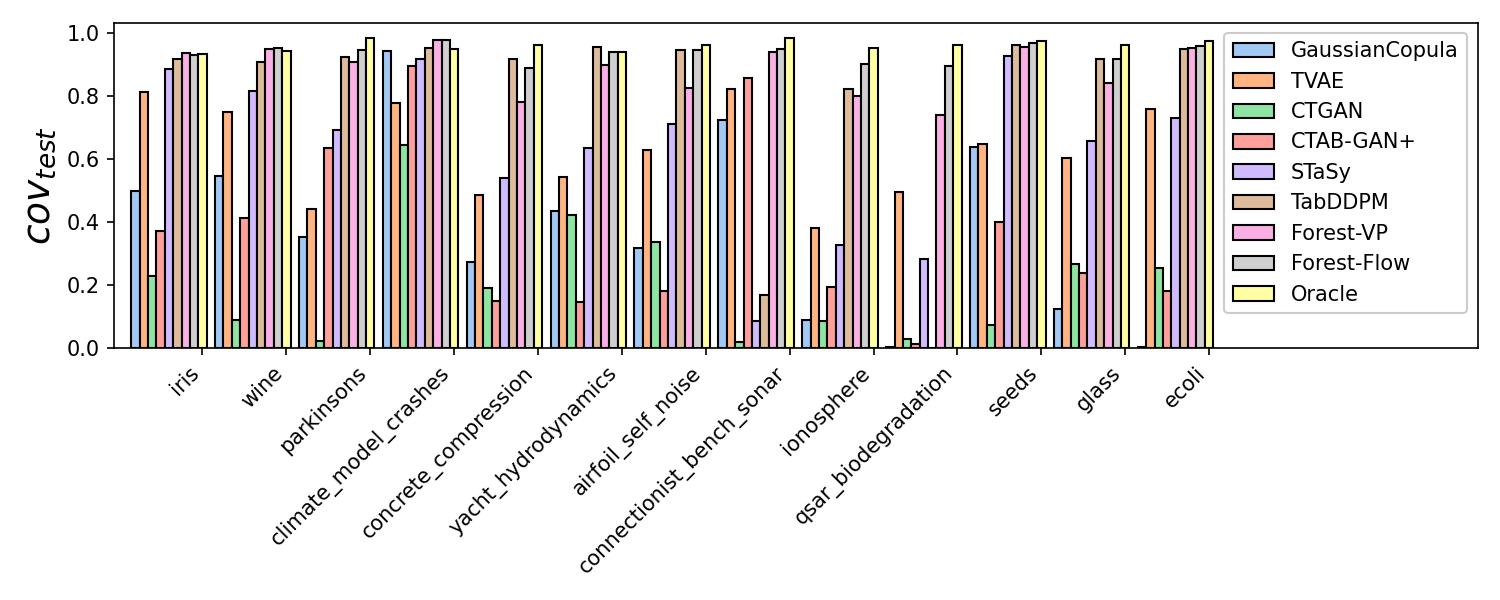}
\end{figure*}
\begin{figure*}[ht]
    \centering
    \includegraphics[width=1\textwidth]{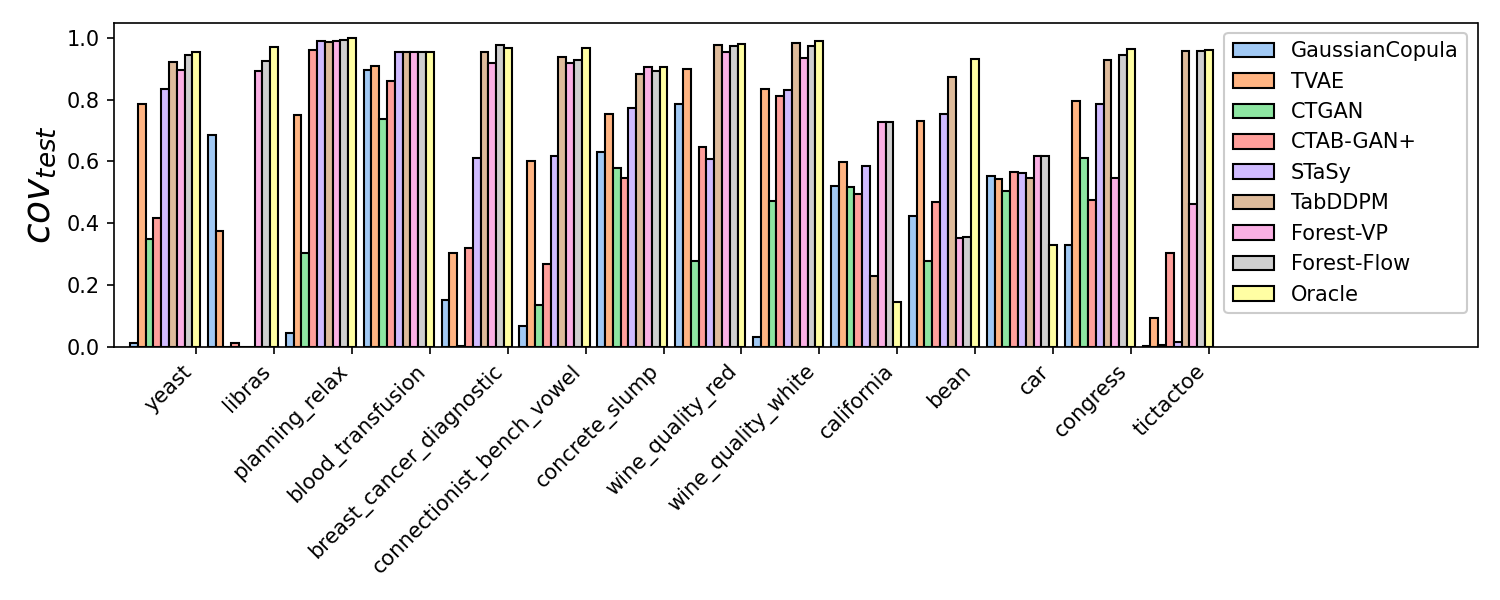}
\end{figure*}
\begin{figure*}[ht]
    \centering
    \includegraphics[width=1\textwidth]{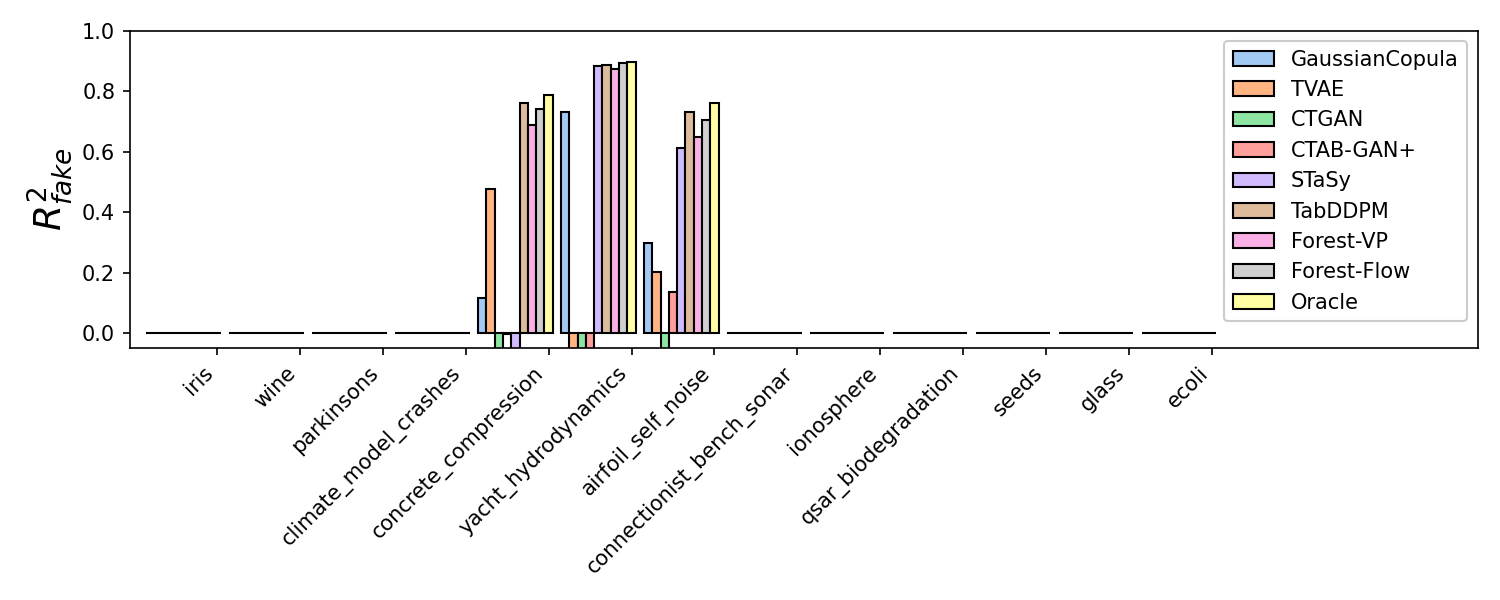}
\end{figure*}
\begin{figure*}[ht]
    \centering
    \includegraphics[width=1\textwidth]{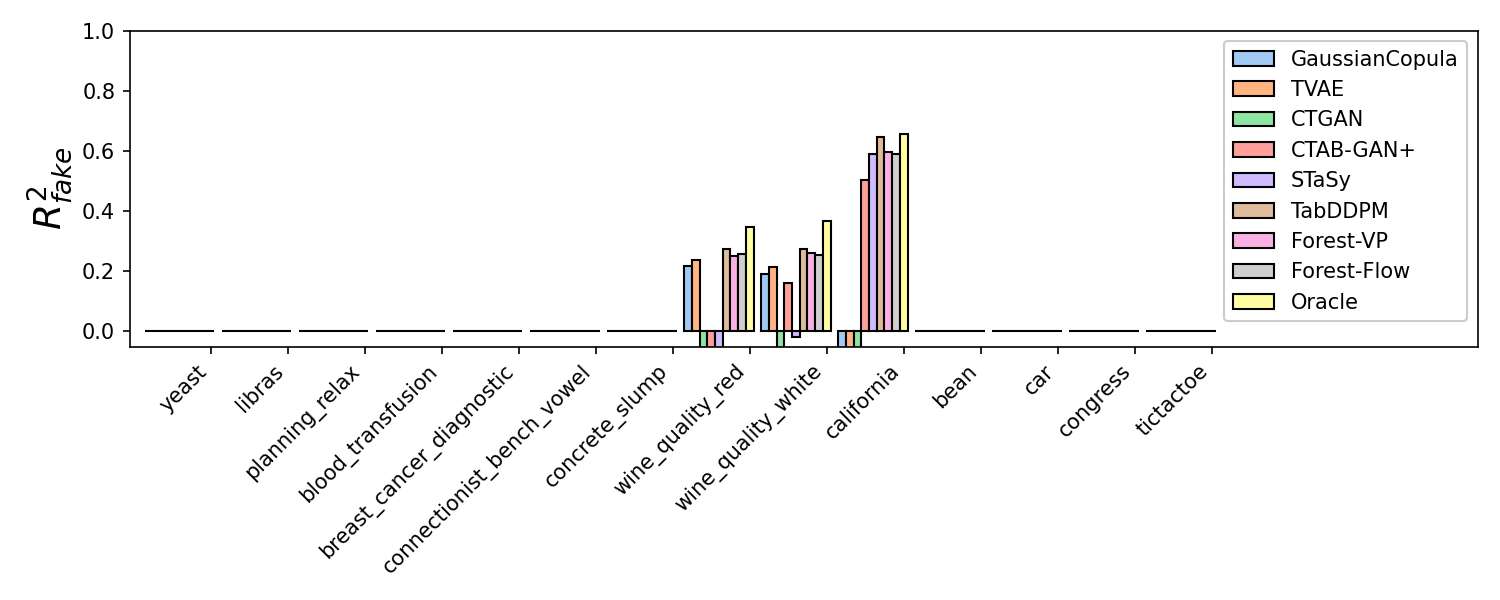}
\end{figure*}
\begin{figure*}[ht]
    \centering
    \includegraphics[width=1\textwidth]{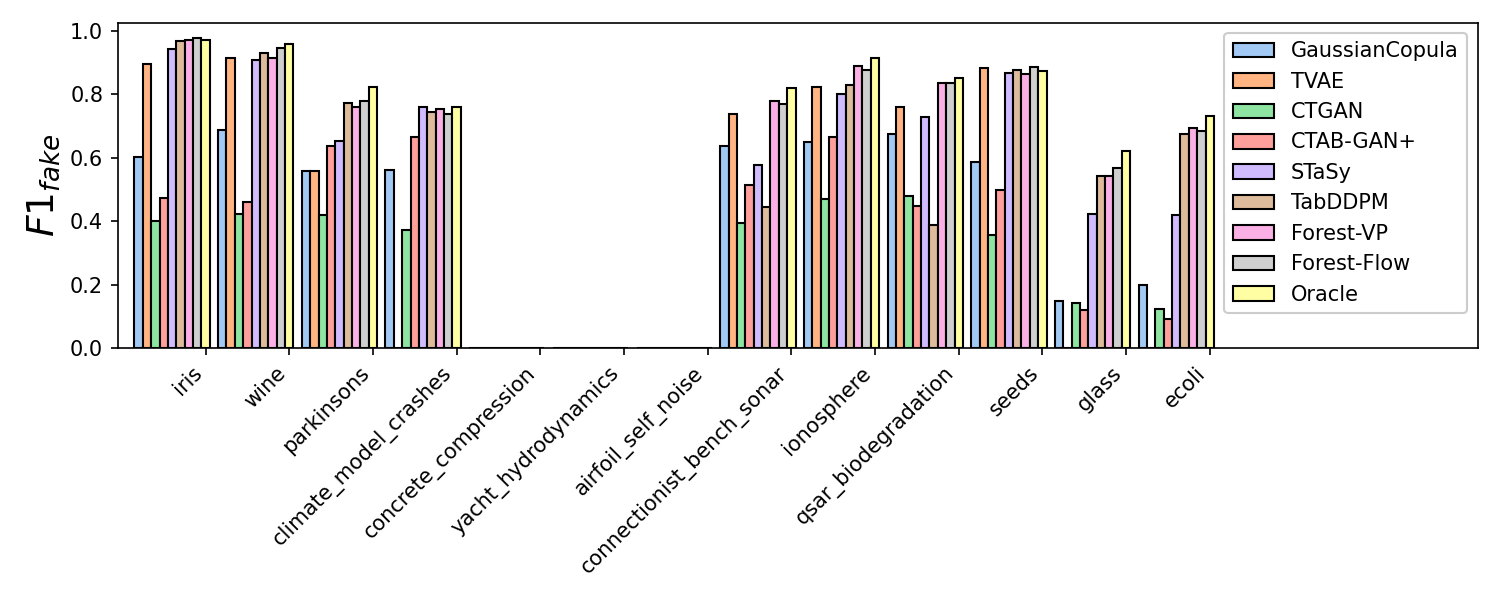}
\end{figure*}
\begin{figure*}[ht]
    \centering
    \includegraphics[width=1\textwidth]{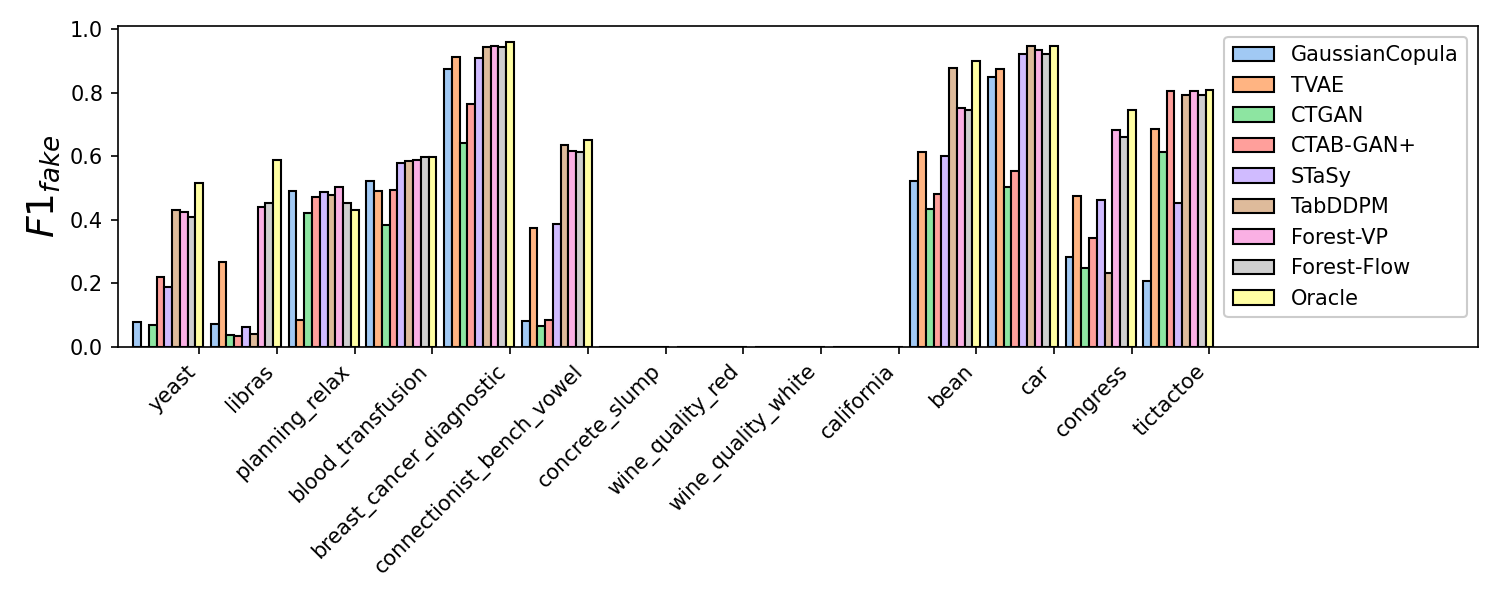}
\end{figure*}
\begin{figure*}[ht]
    \centering
    \includegraphics[width=1\textwidth]{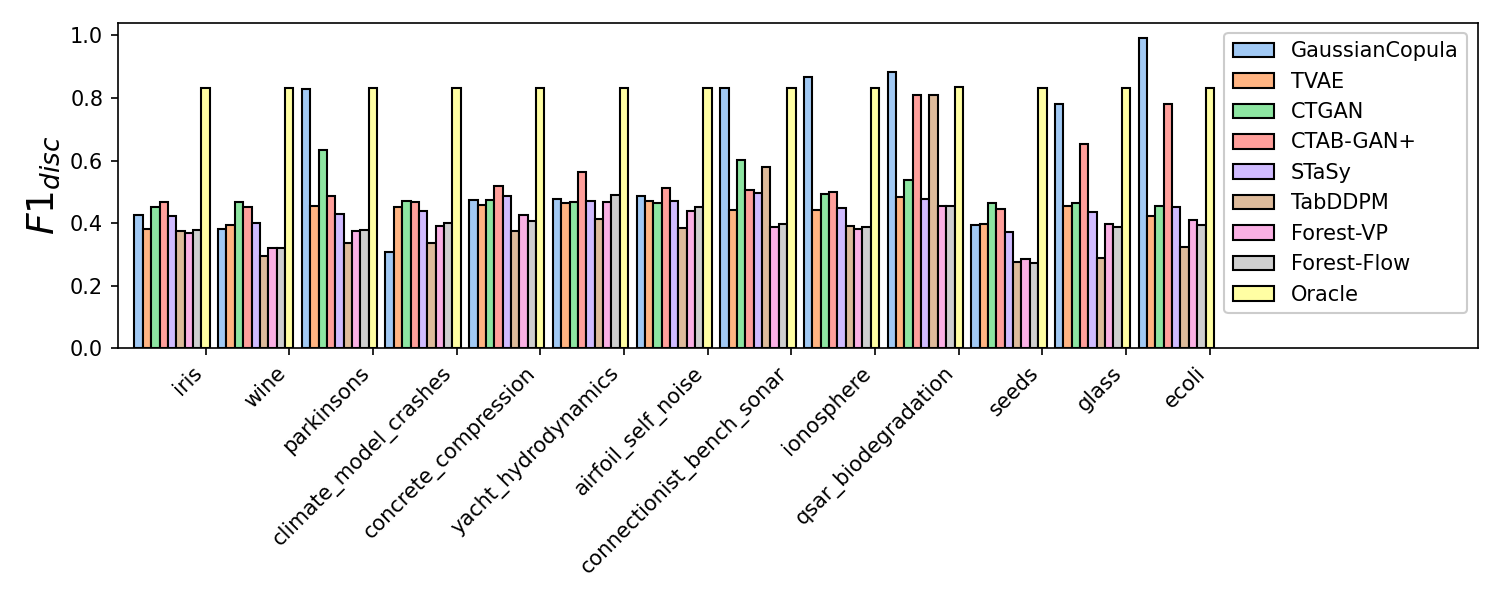}
\end{figure*}
\begin{figure*}[ht]
    \centering
    \includegraphics[width=1\textwidth]{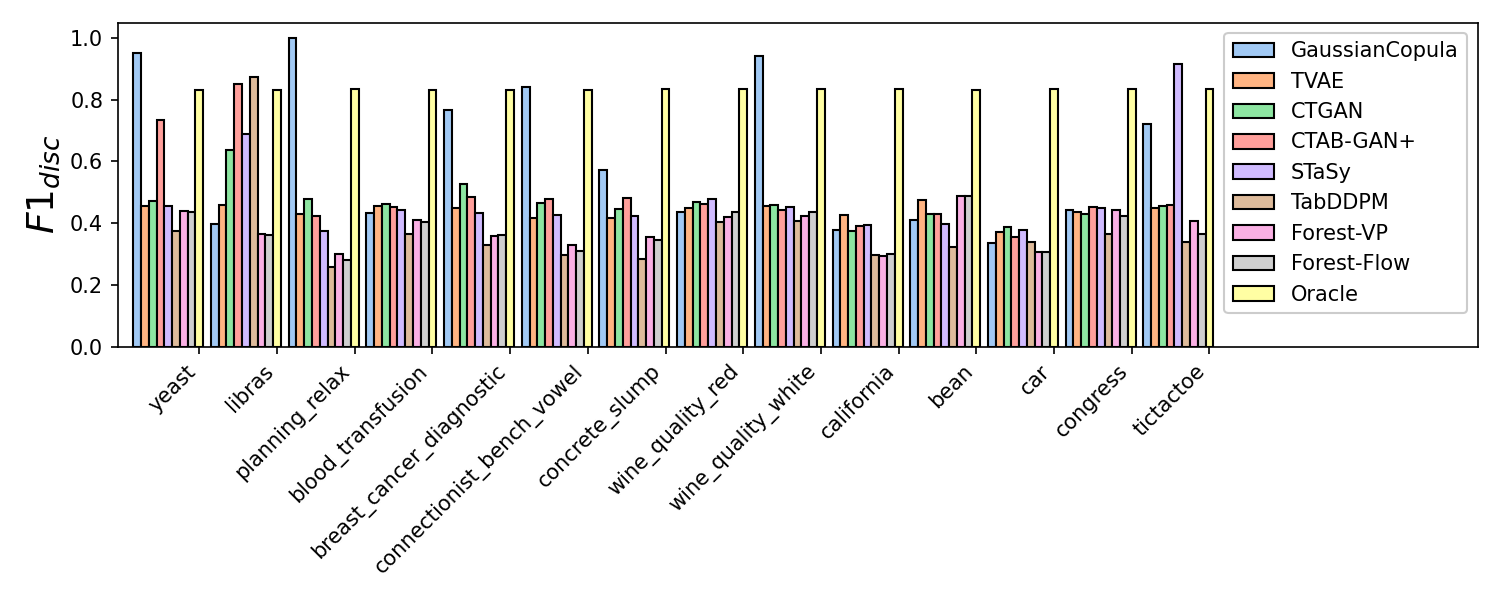}
\end{figure*}
\begin{figure*}[ht]
    \centering
    \includegraphics[width=1\textwidth]{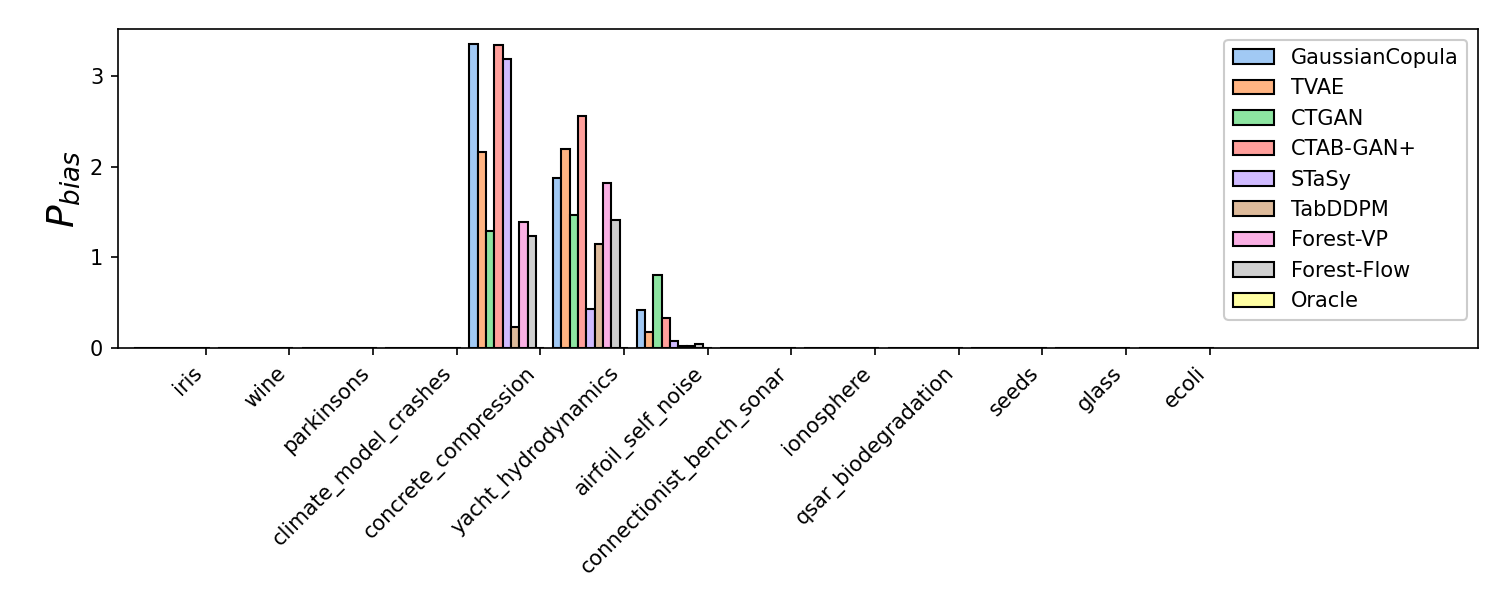}
\end{figure*}
\begin{figure*}[ht]
    \centering
    \includegraphics[width=1\textwidth]{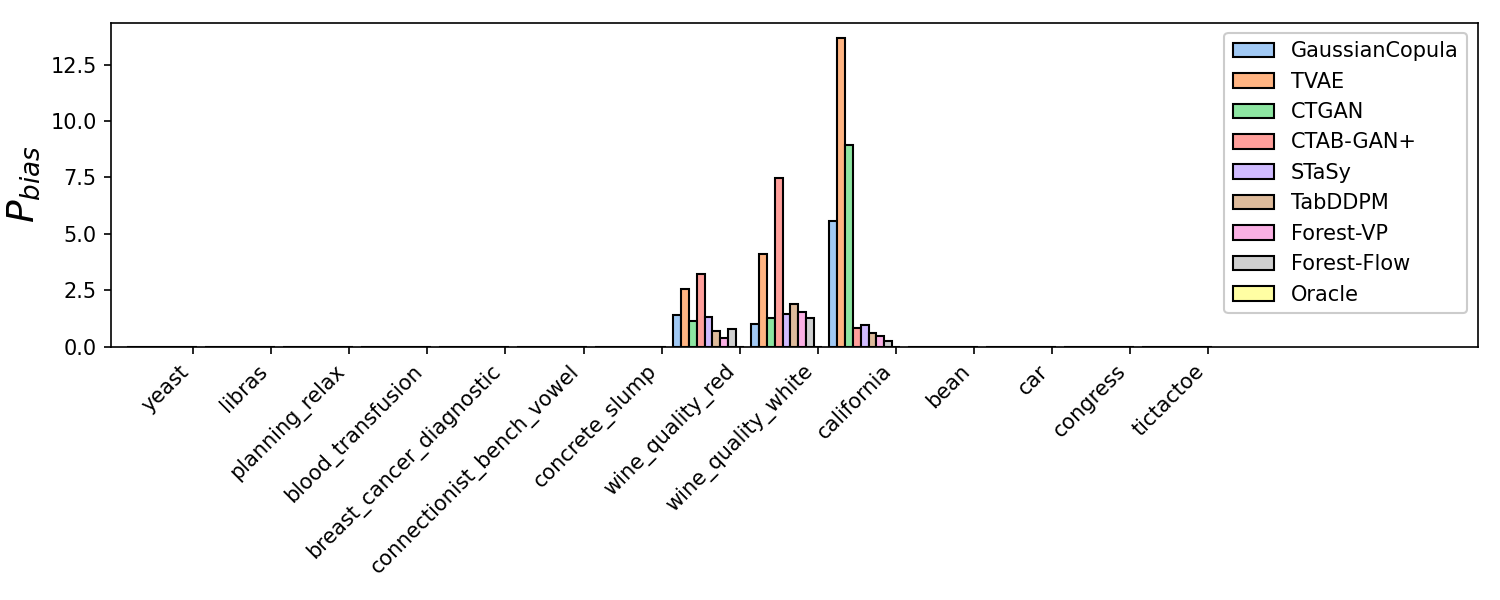}
\end{figure*}
\begin{figure*}[ht]
    \centering
    \includegraphics[width=1\textwidth]{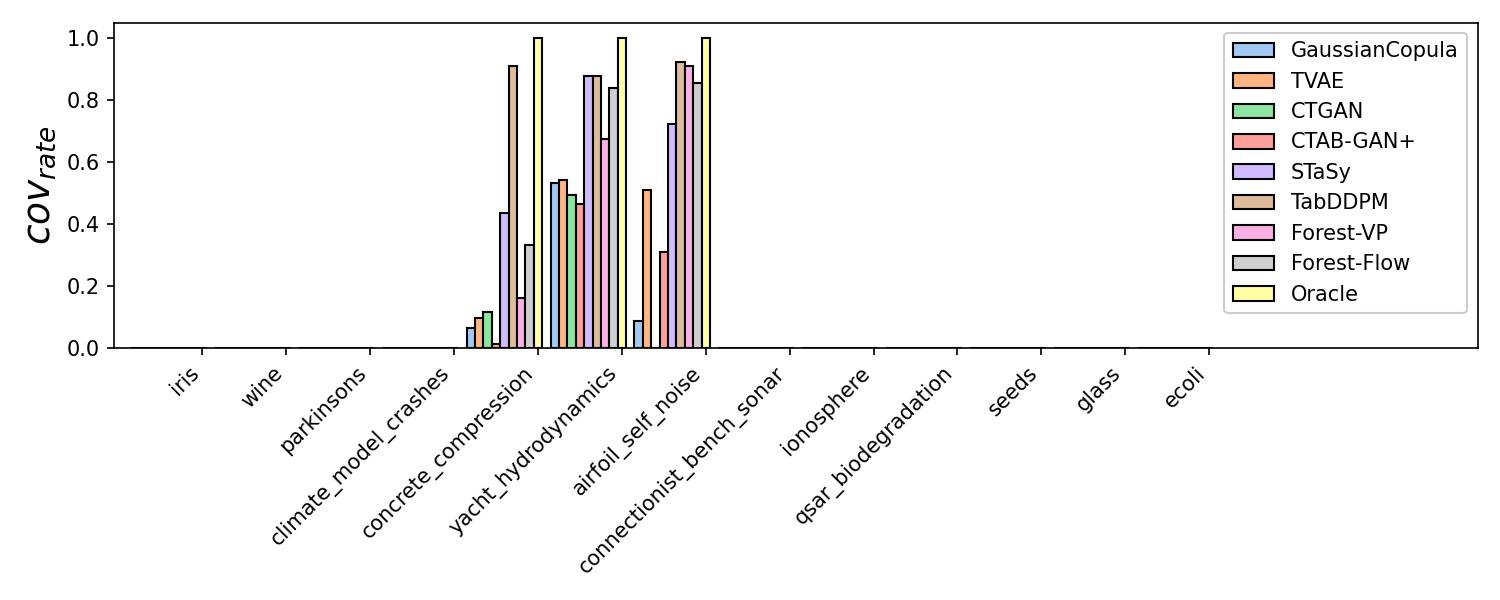}
\end{figure*}
\begin{figure*}[ht]
    \centering
    \includegraphics[width=1\textwidth]{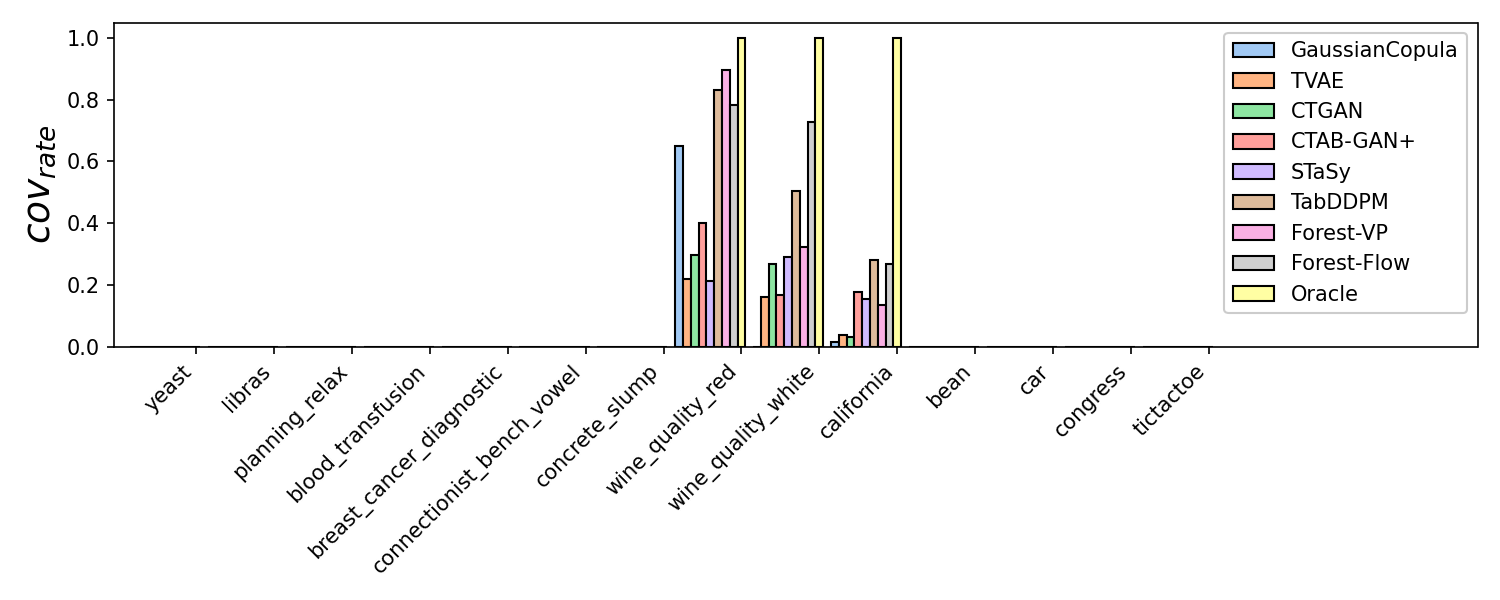}
\end{figure*}

\clearpage

\subsubsection{Bar plots for imputation}\label{app:bar_plot_imp}

\begin{figure*}[ht]
    \centering
    \includegraphics[width=1\textwidth]{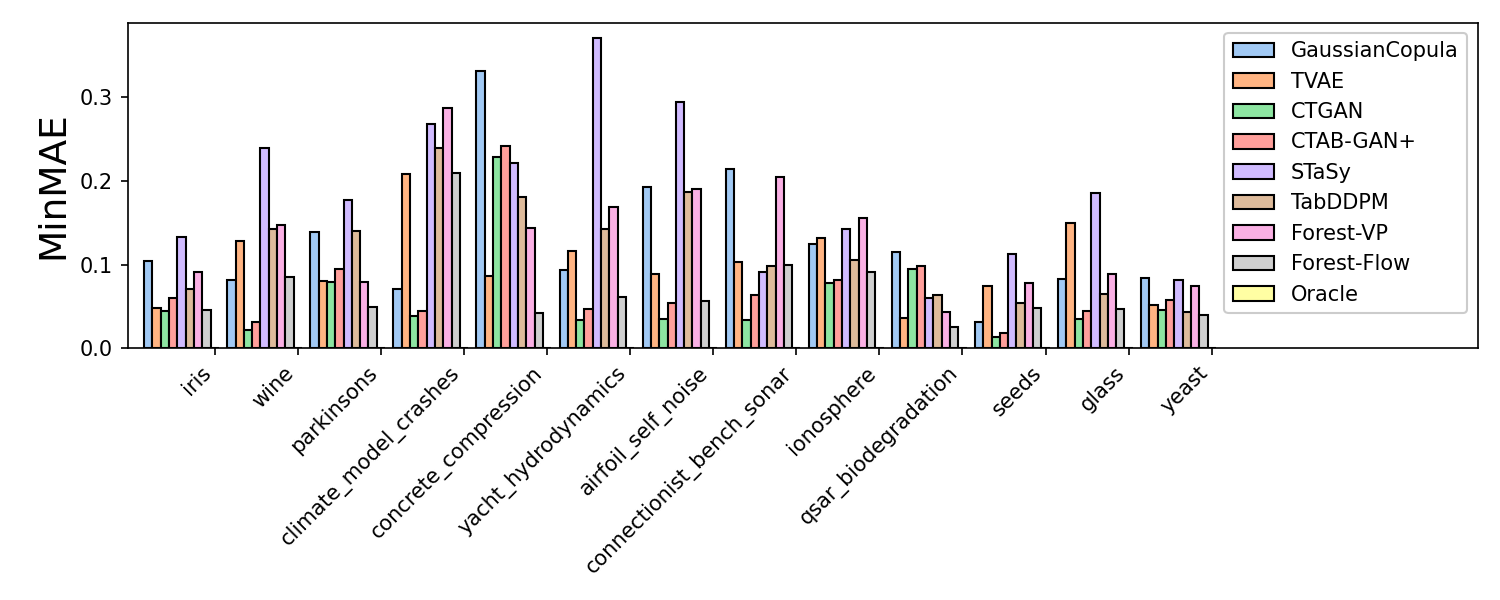}
\end{figure*}
\begin{figure*}[ht]
    \centering
    \includegraphics[width=1\textwidth]{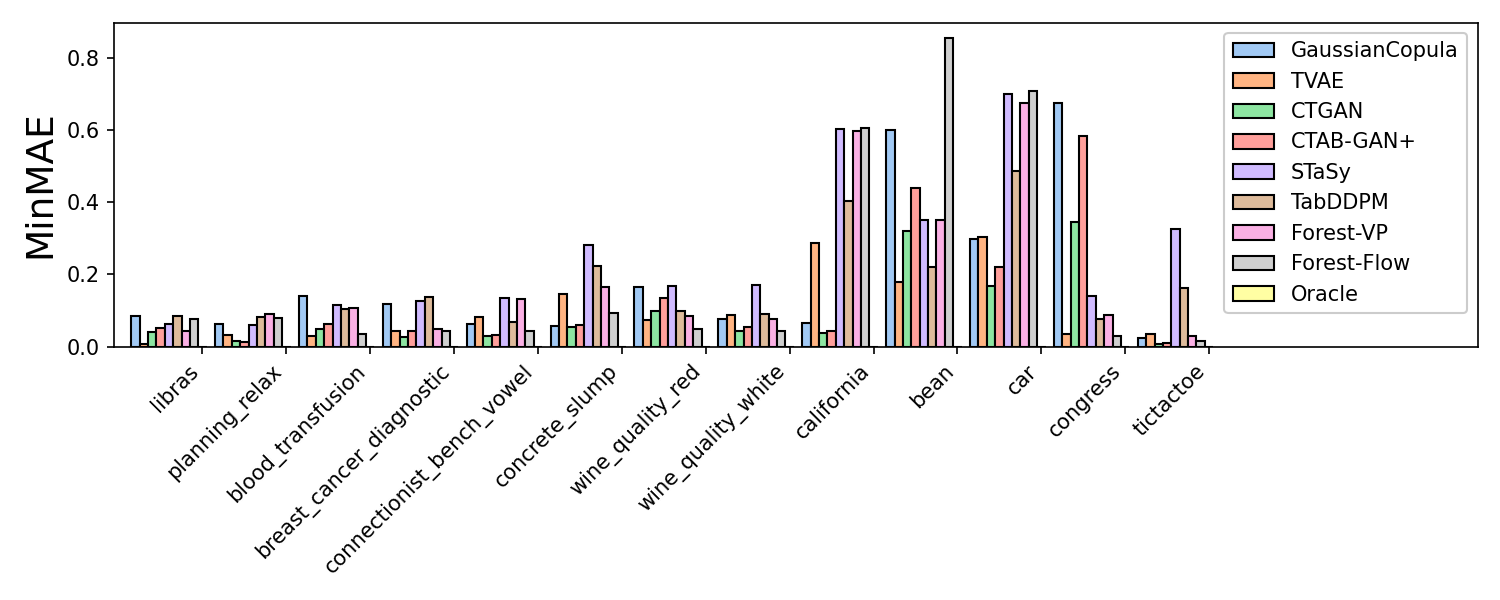}
\end{figure*}
\begin{figure*}[ht]
    \centering
    \includegraphics[width=1\textwidth]{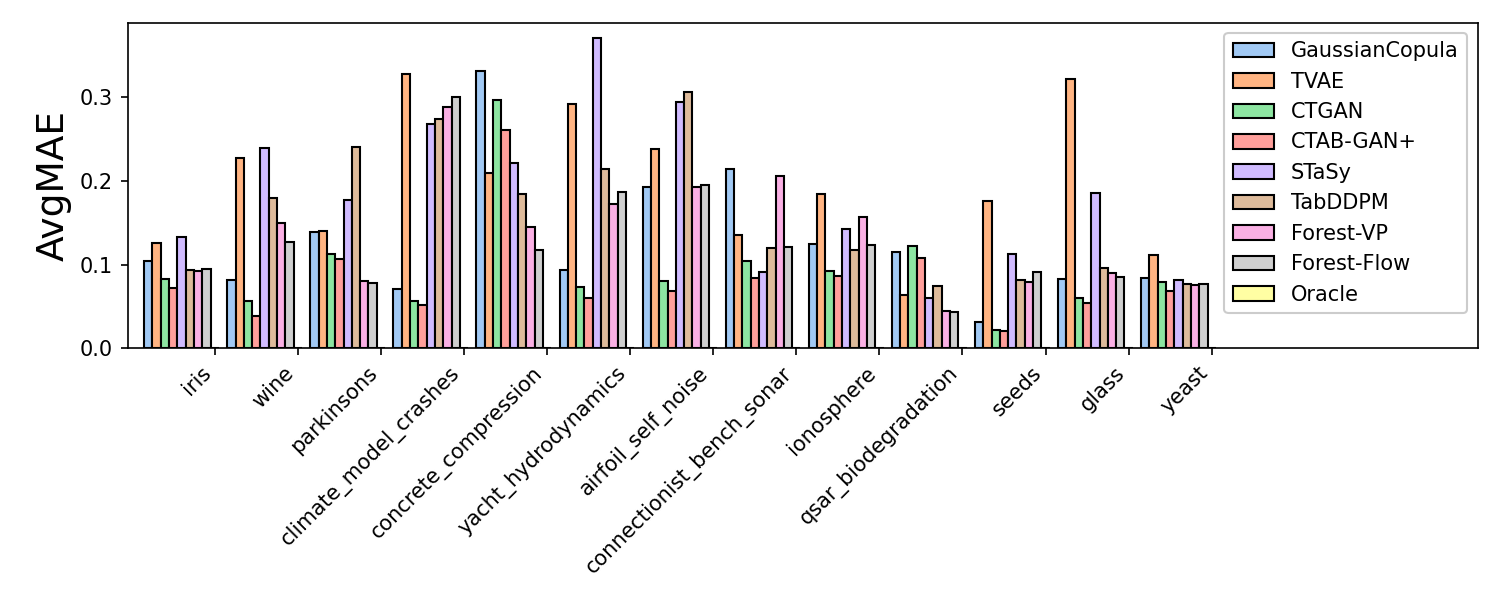}
\end{figure*}
\begin{figure*}[ht]
    \centering
    \includegraphics[width=1\textwidth]{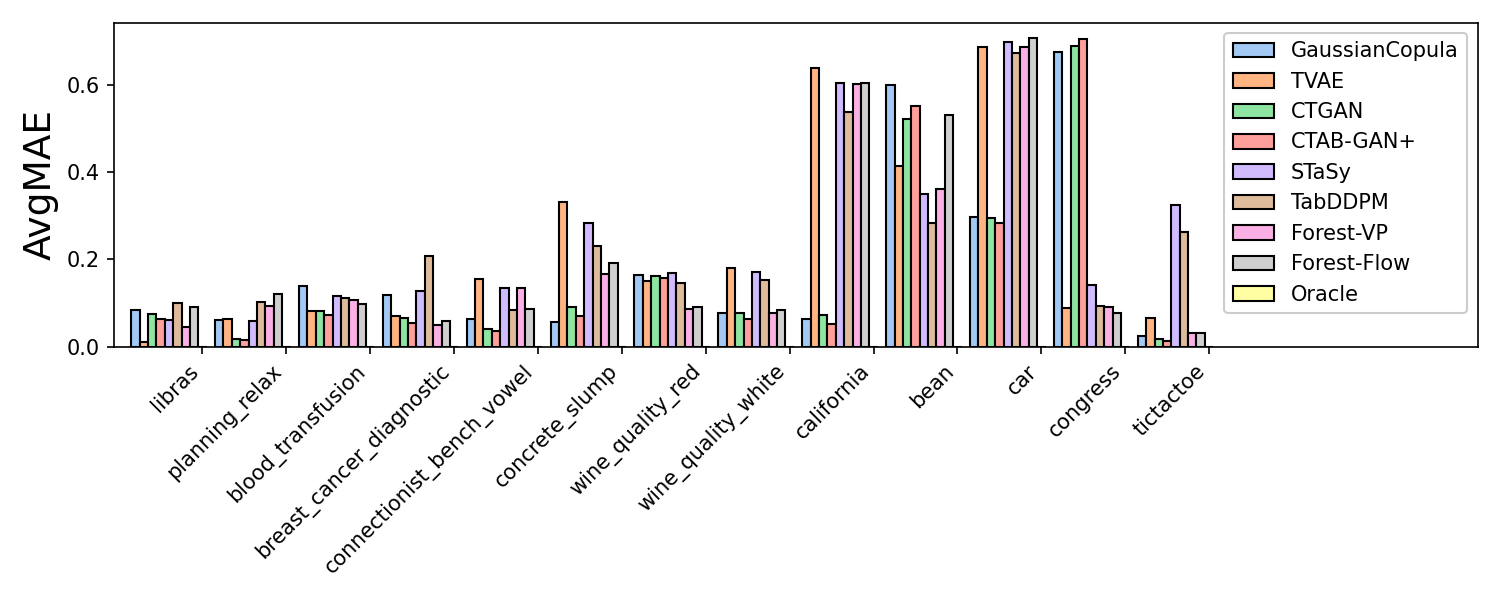}
\end{figure*}
\begin{figure*}[ht]
    \centering
    \includegraphics[width=1\textwidth]{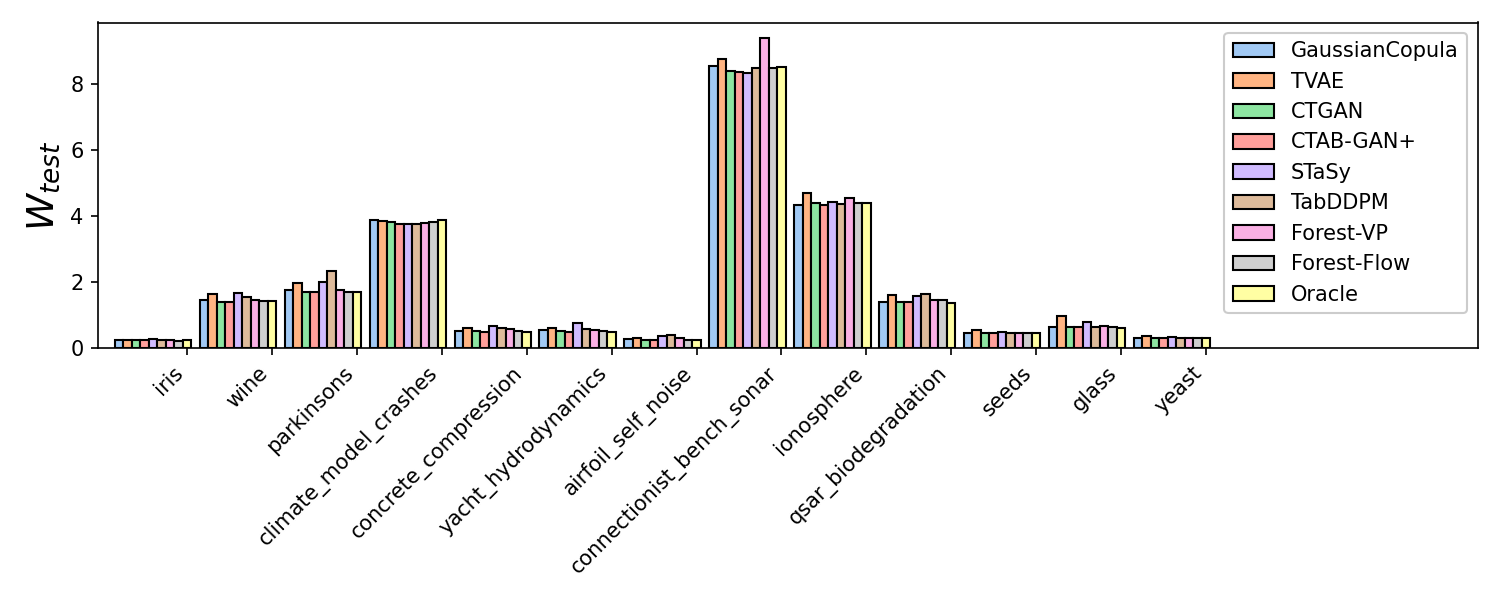}
\end{figure*}
\begin{figure*}[ht]
    \centering
    \includegraphics[width=1\textwidth]{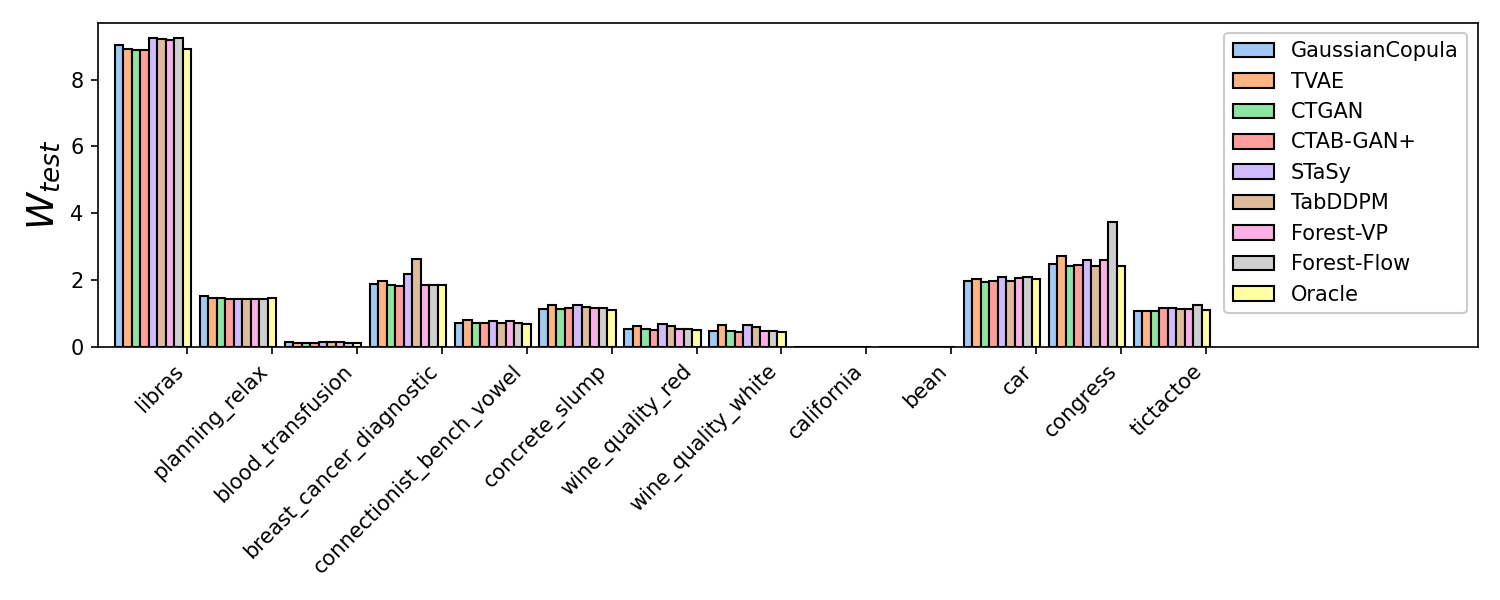}
\end{figure*}
\begin{figure*}[ht]
    \centering
    \includegraphics[width=1\textwidth]{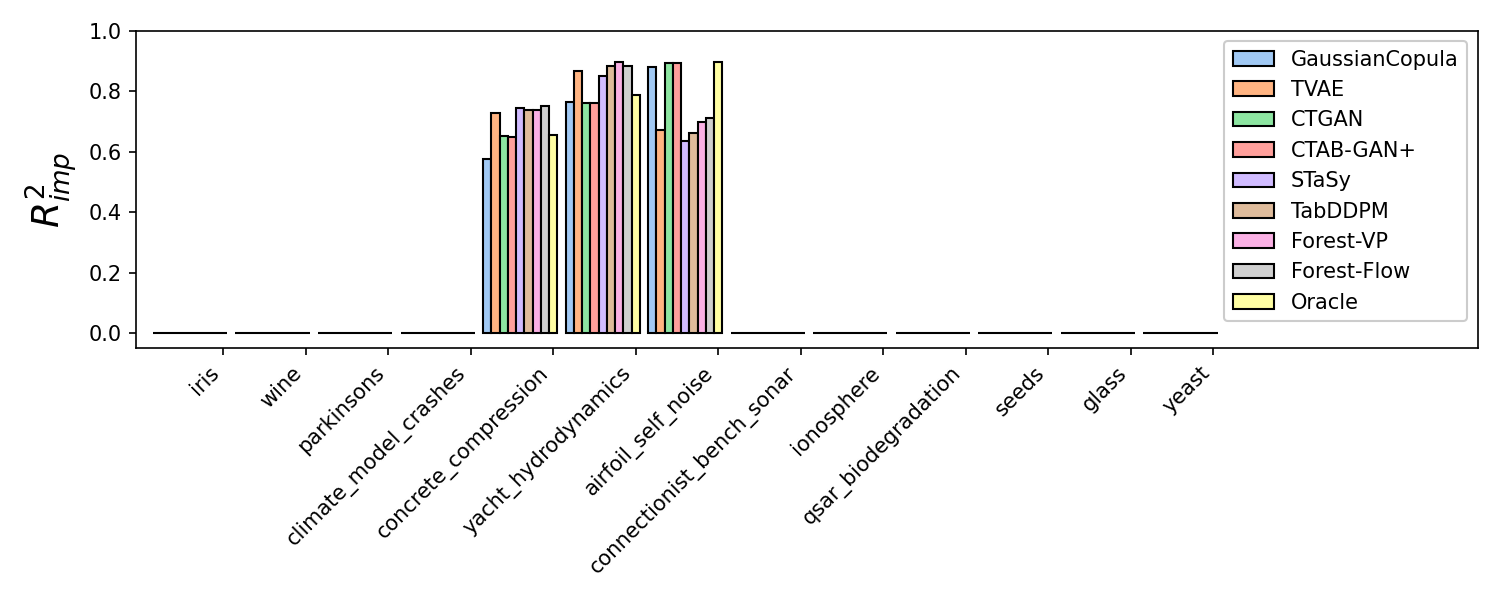}
\end{figure*}
\begin{figure*}[ht]
    \centering
    \includegraphics[width=1\textwidth]{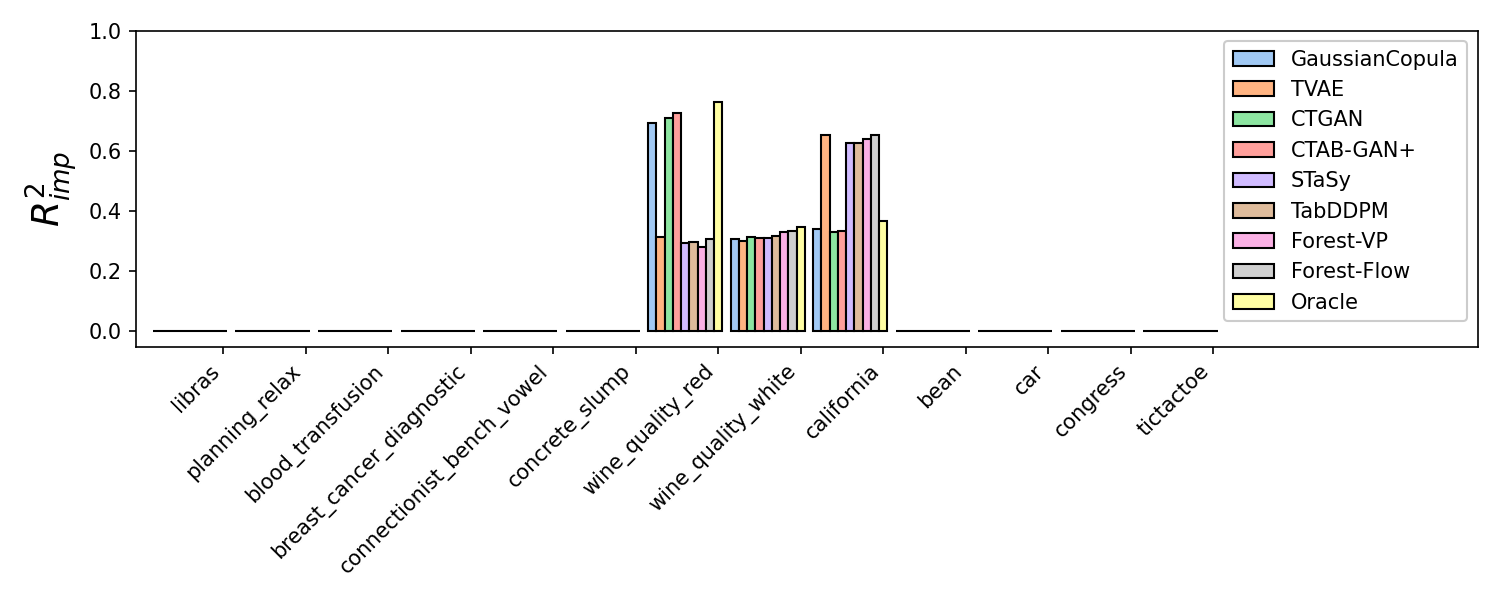}
\end{figure*}
\begin{figure*}[ht]
    \centering
    \includegraphics[width=1\textwidth]{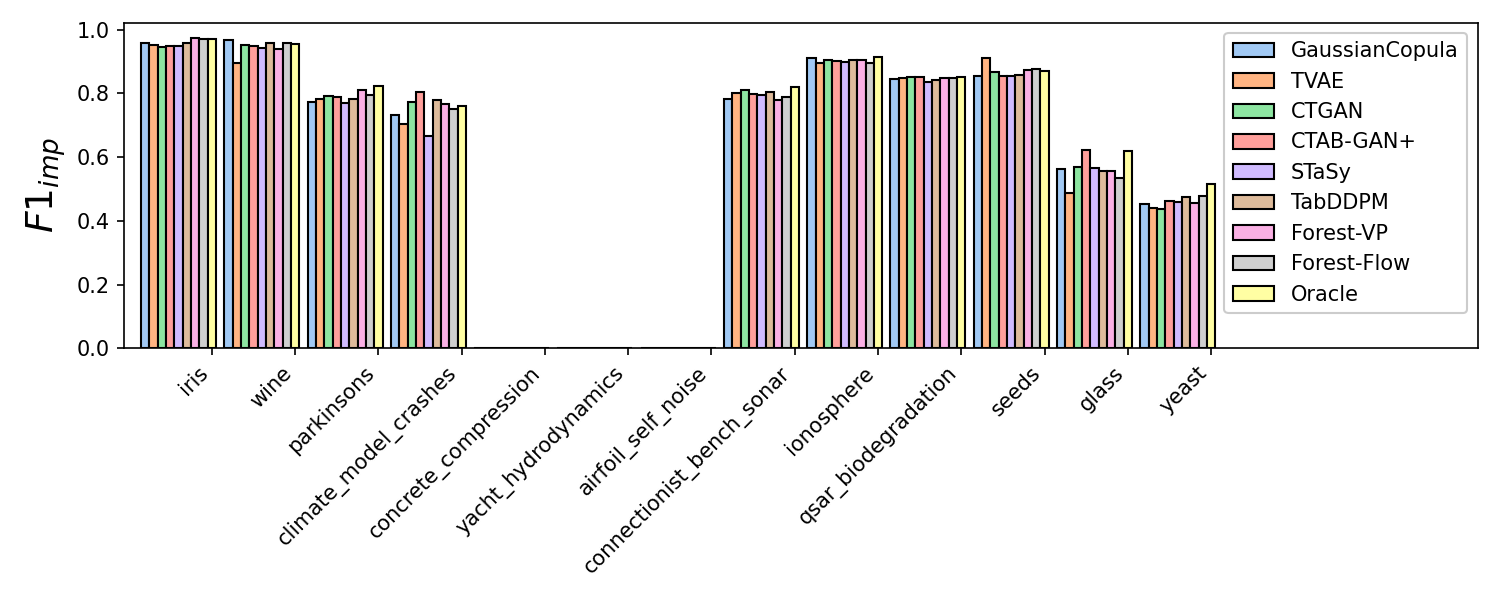}
\end{figure*}
\begin{figure*}[ht]
    \centering
    \includegraphics[width=1\textwidth]{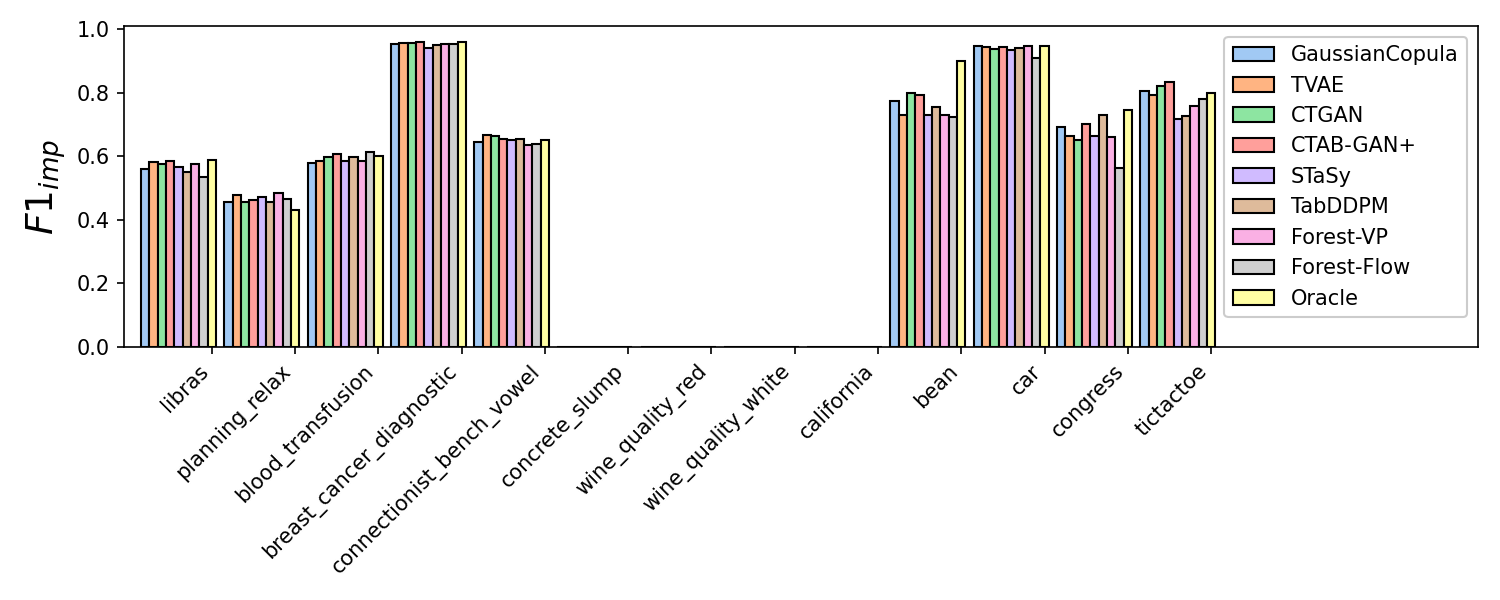}
\end{figure*}
\begin{figure*}[ht]
    \centering
    \includegraphics[width=1\textwidth]{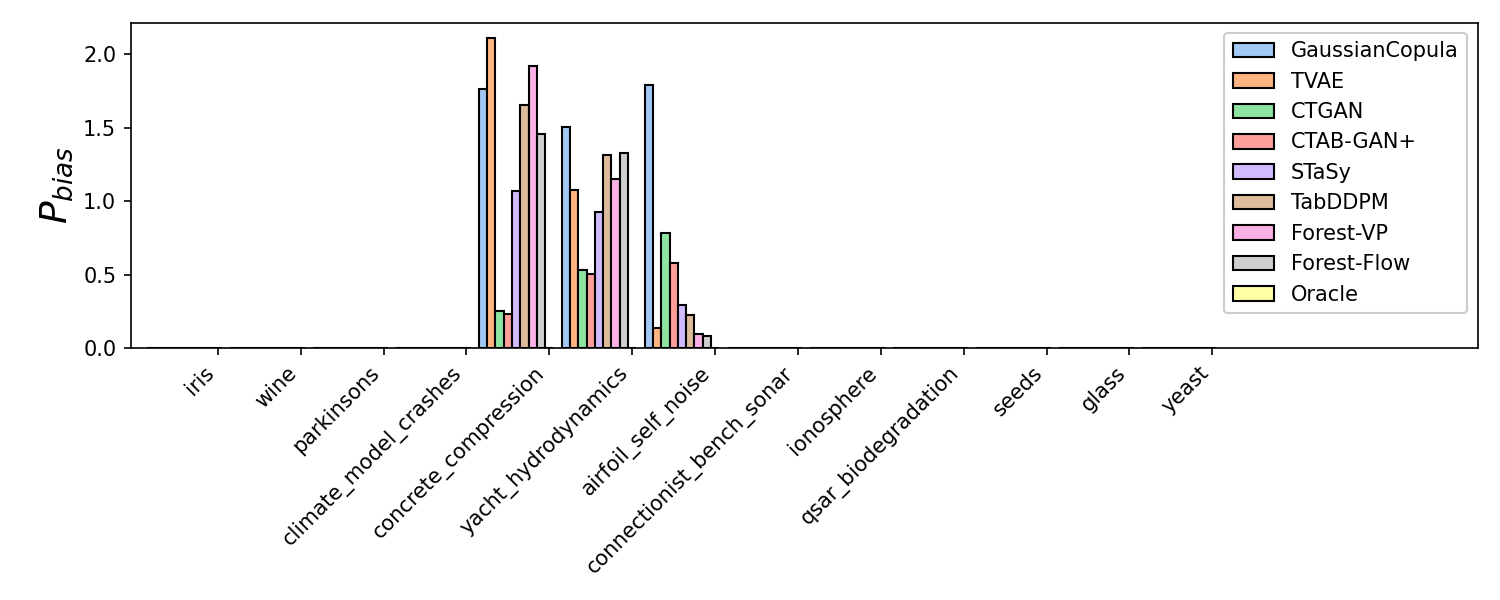}
\end{figure*}
\begin{figure*}[ht]
    \centering
    \includegraphics[width=1\textwidth]{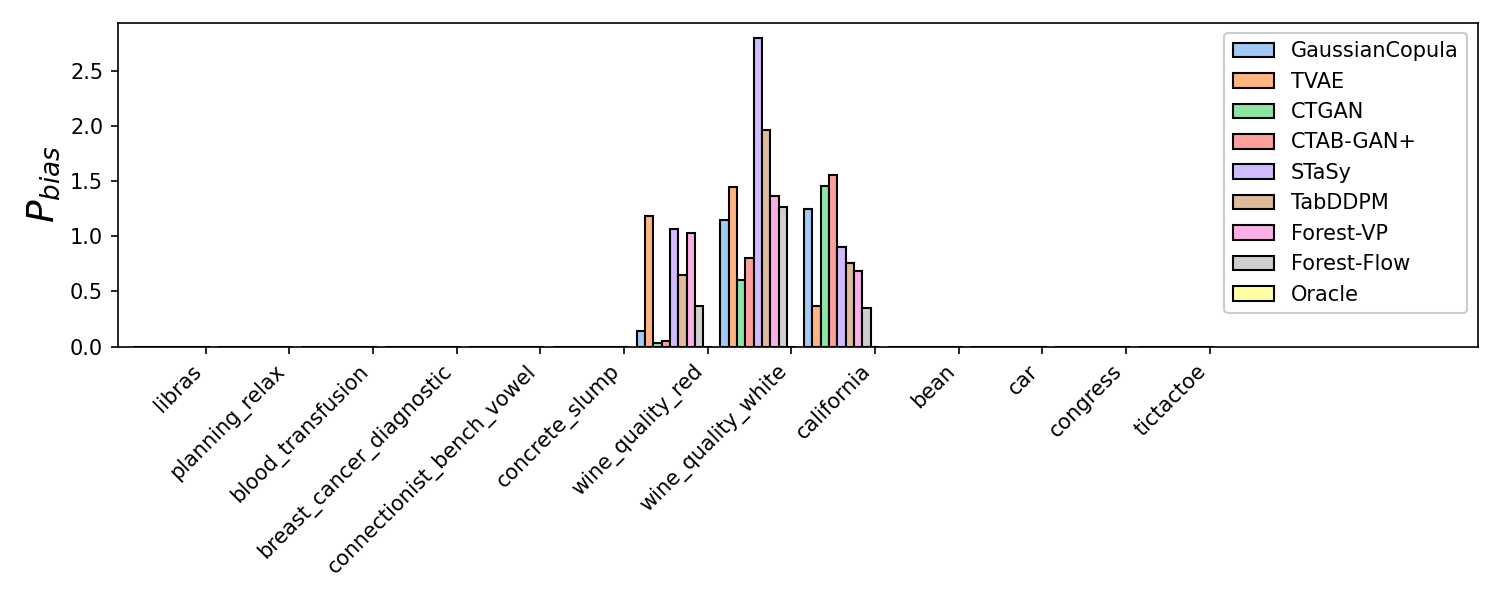}
\end{figure*}
\begin{figure*}[ht]
    \centering
    \includegraphics[width=1\textwidth]{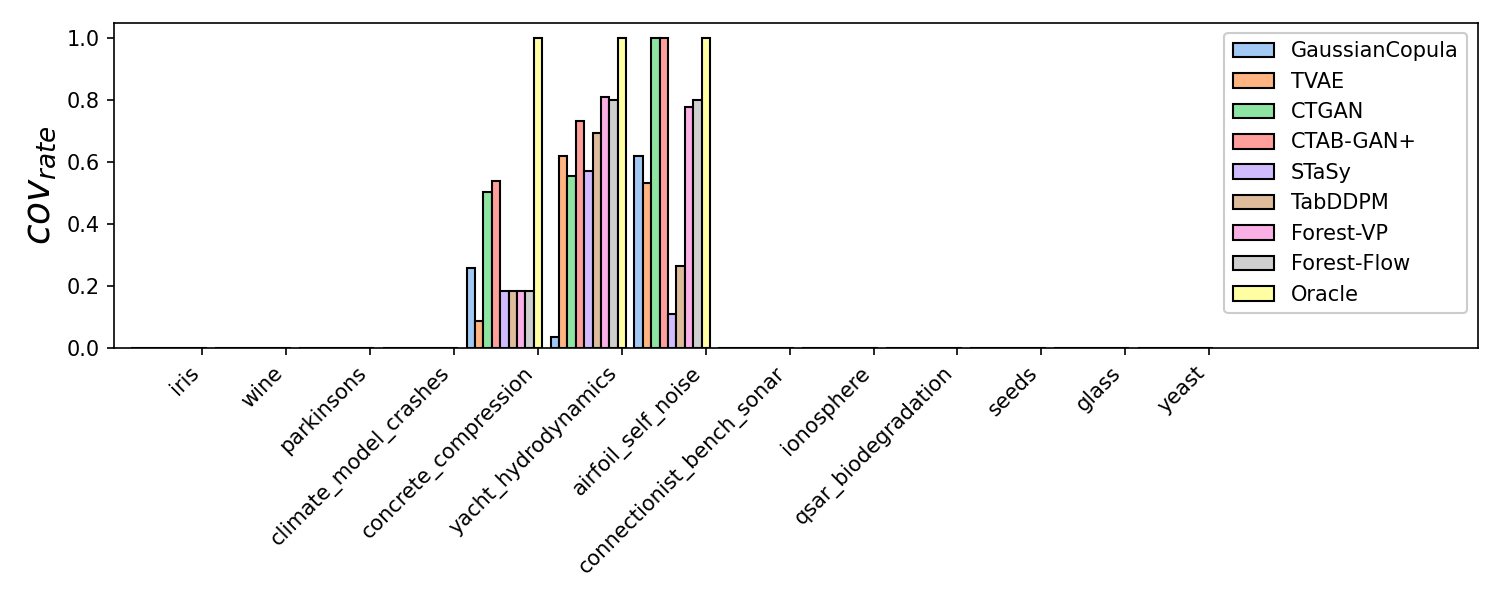}
\end{figure*}
\begin{figure*}[ht]
    \centering
    \includegraphics[width=1\textwidth]{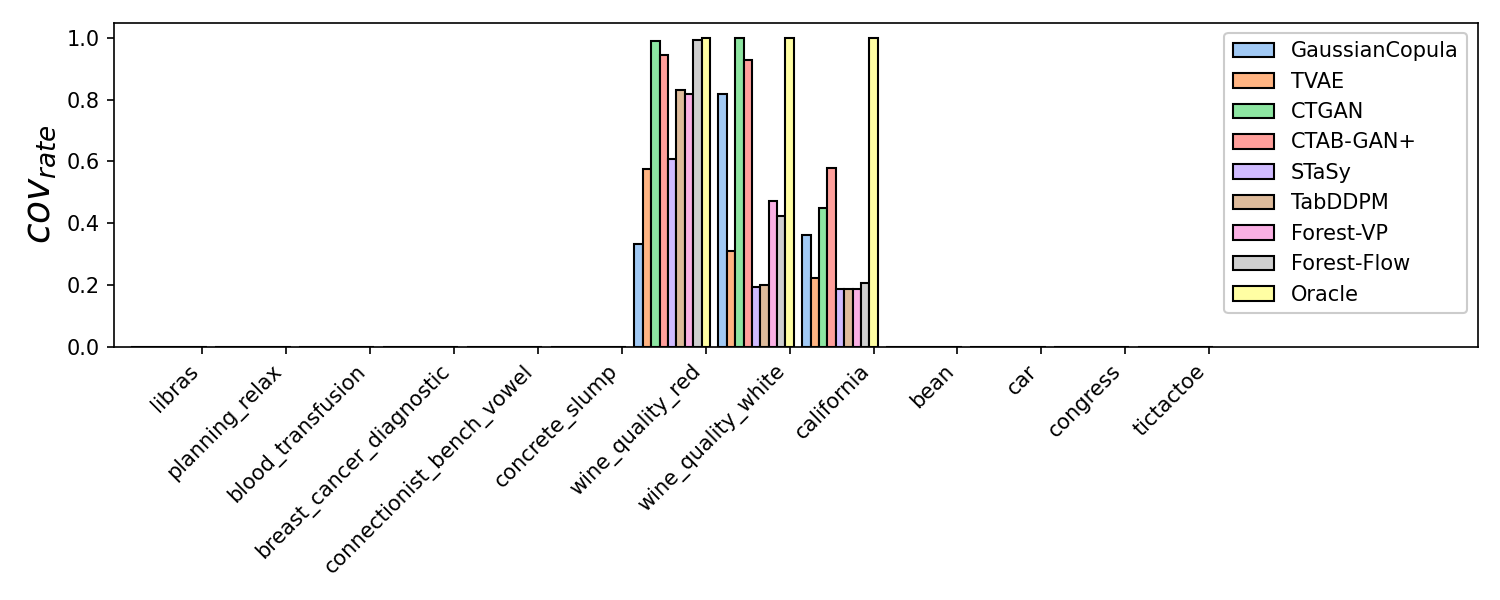}
\end{figure*}

\clearpage

\subsubsection{Bar plots for generation with incomplete data}\label{app:bar_plot_gen_miss}

\begin{figure*}[ht]
    \centering
    \includegraphics[width=1\textwidth]{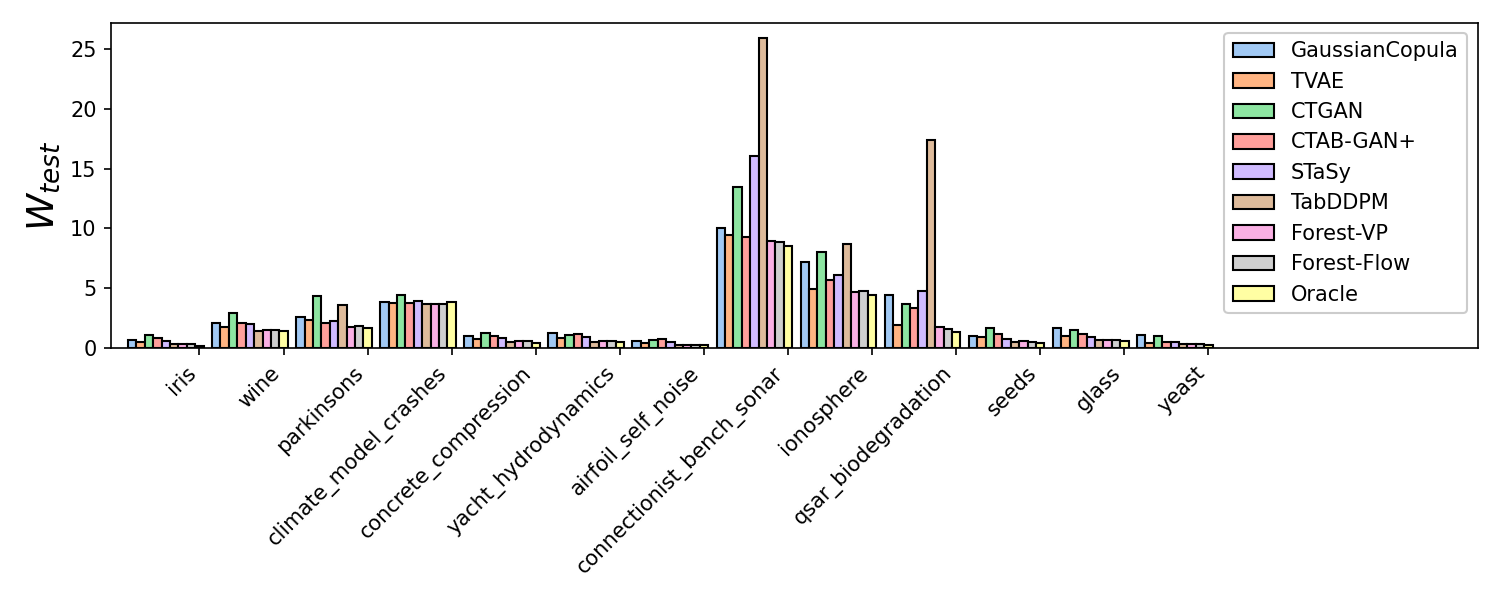}
\end{figure*}
\begin{figure*}[ht]
    \centering
    \includegraphics[width=1\textwidth]{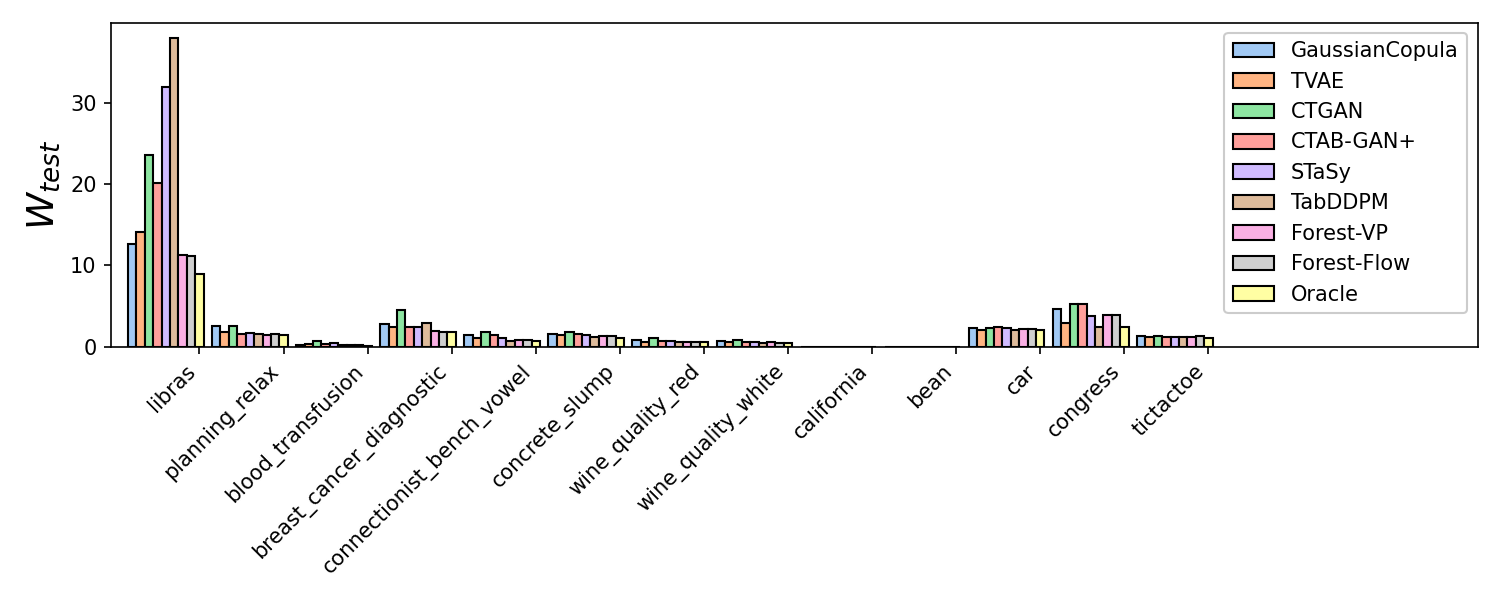}
\end{figure*}
\begin{figure*}[ht]
    \centering
    \includegraphics[width=1\textwidth]{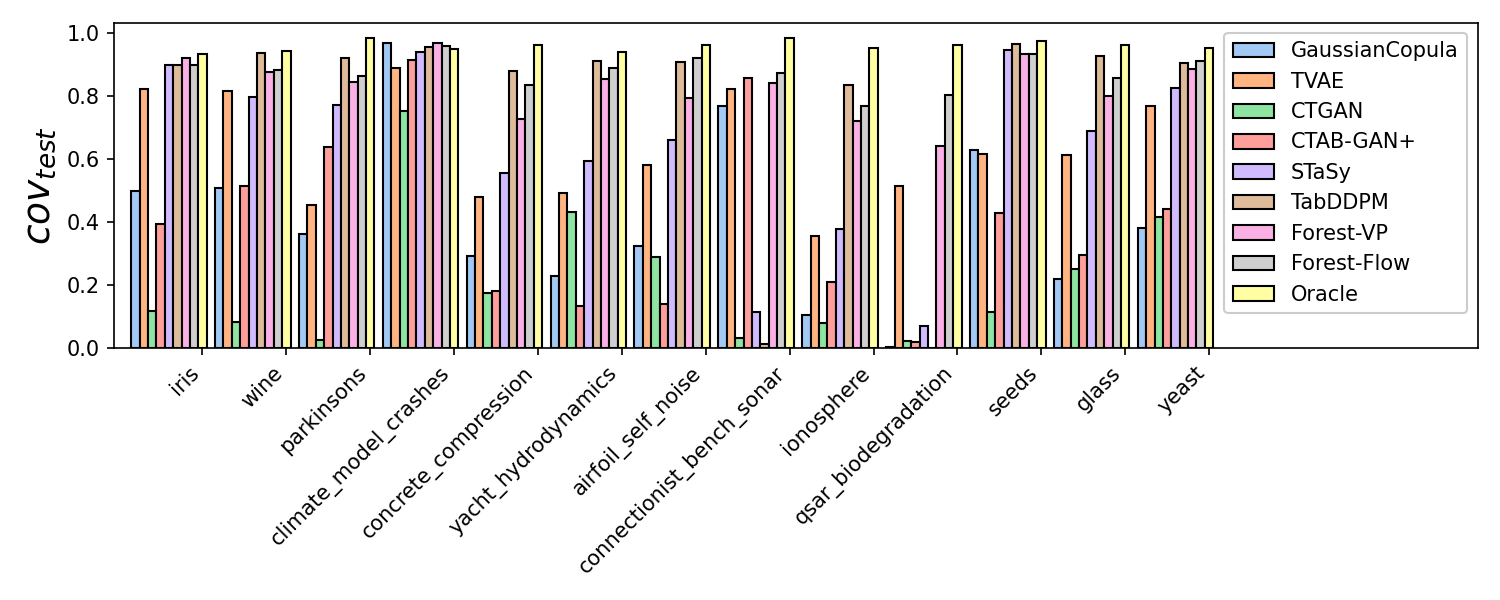}
\end{figure*}
\begin{figure*}[ht]
    \centering
    \includegraphics[width=1\textwidth]{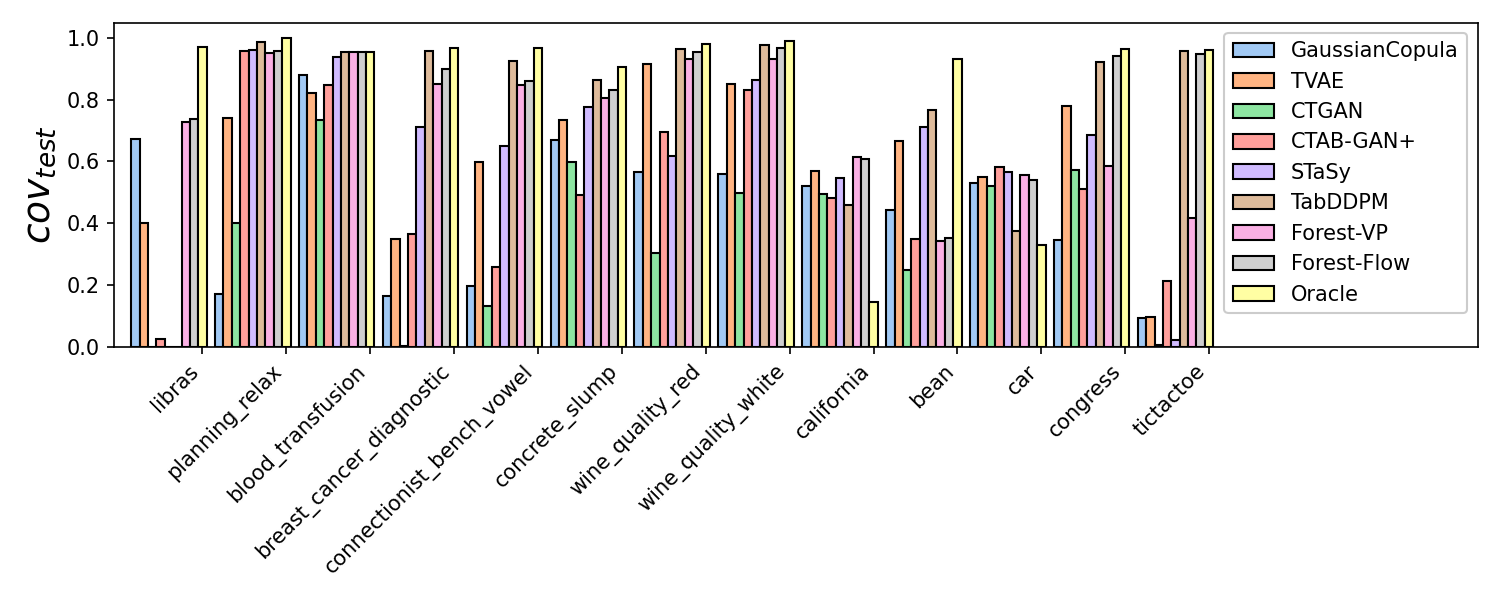}
\end{figure*}
\begin{figure*}[ht]
    \centering
    \includegraphics[width=1\textwidth]{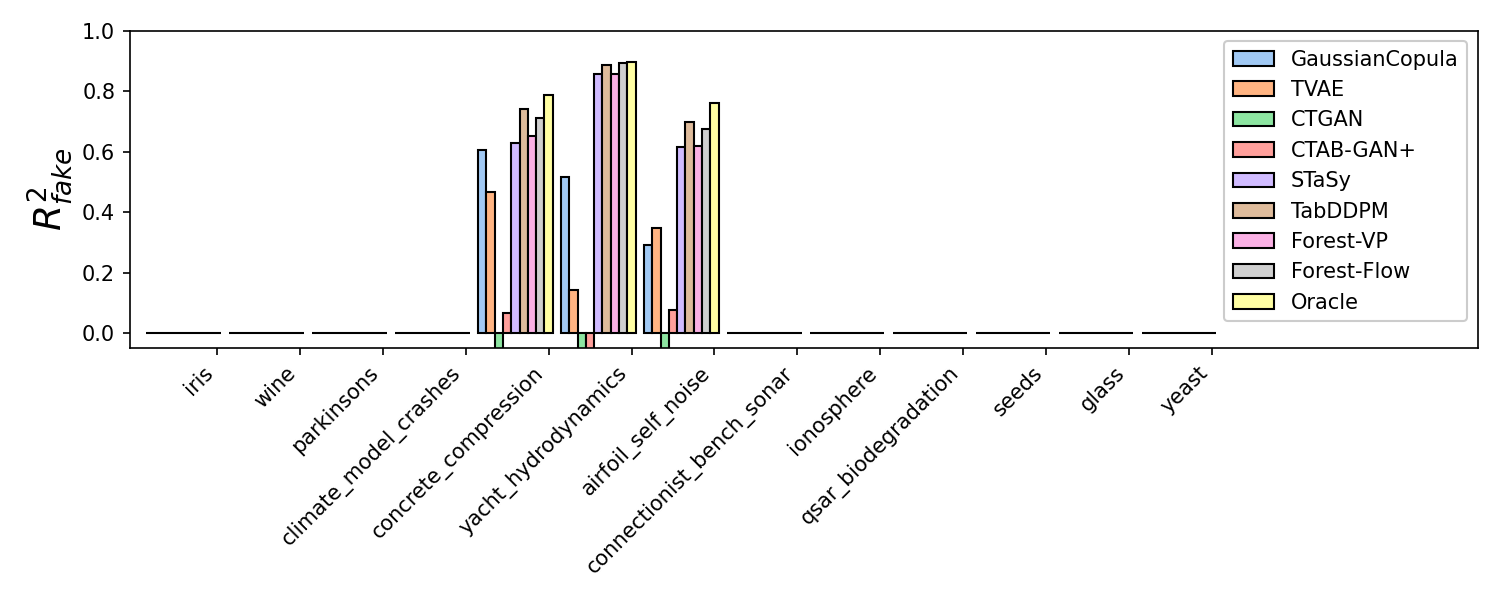}
\end{figure*}
\begin{figure*}[ht]
    \centering
    \includegraphics[width=1\textwidth]{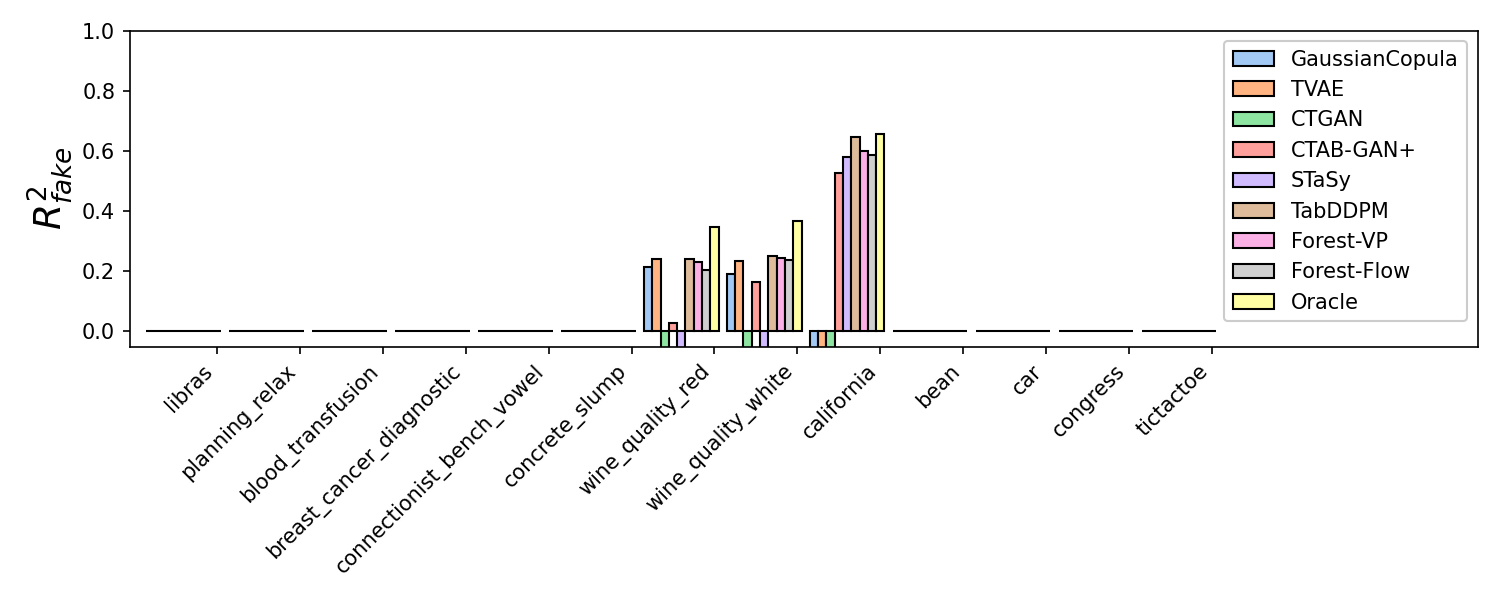}
\end{figure*}
\begin{figure*}[ht]
    \centering
    \includegraphics[width=1\textwidth]{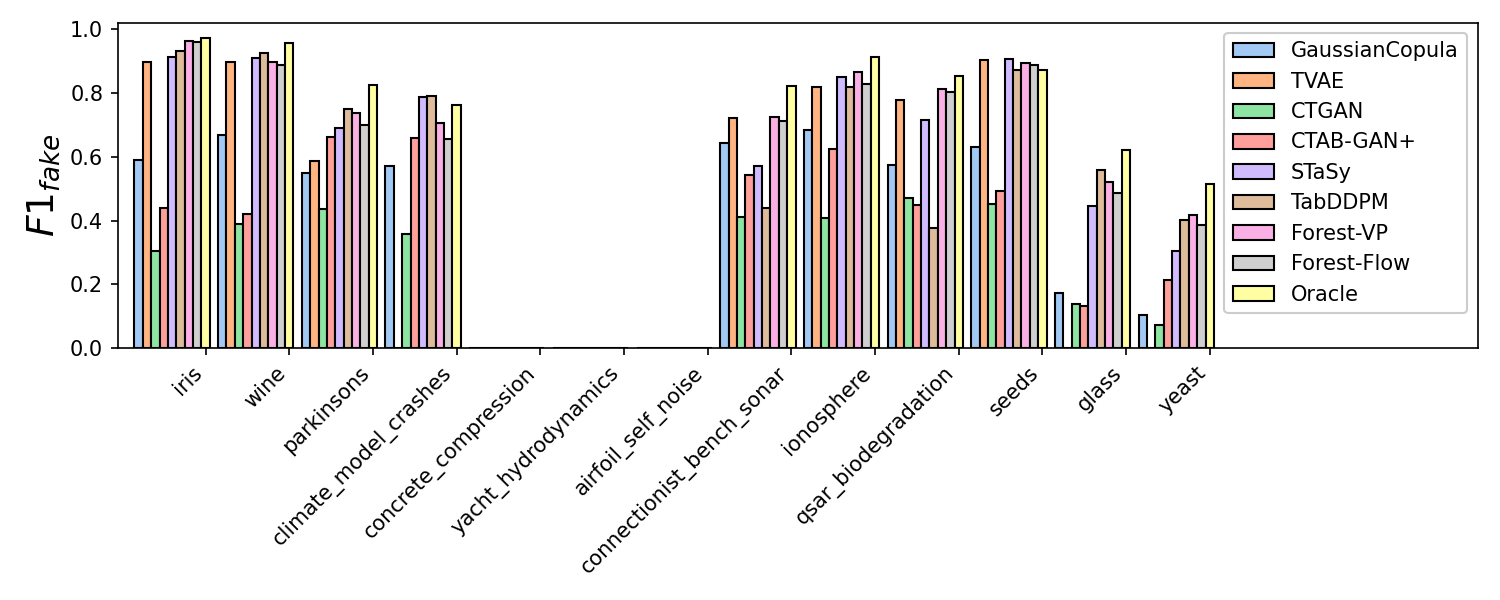}
\end{figure*}
\begin{figure*}[ht]
    \centering
    \includegraphics[width=1\textwidth]{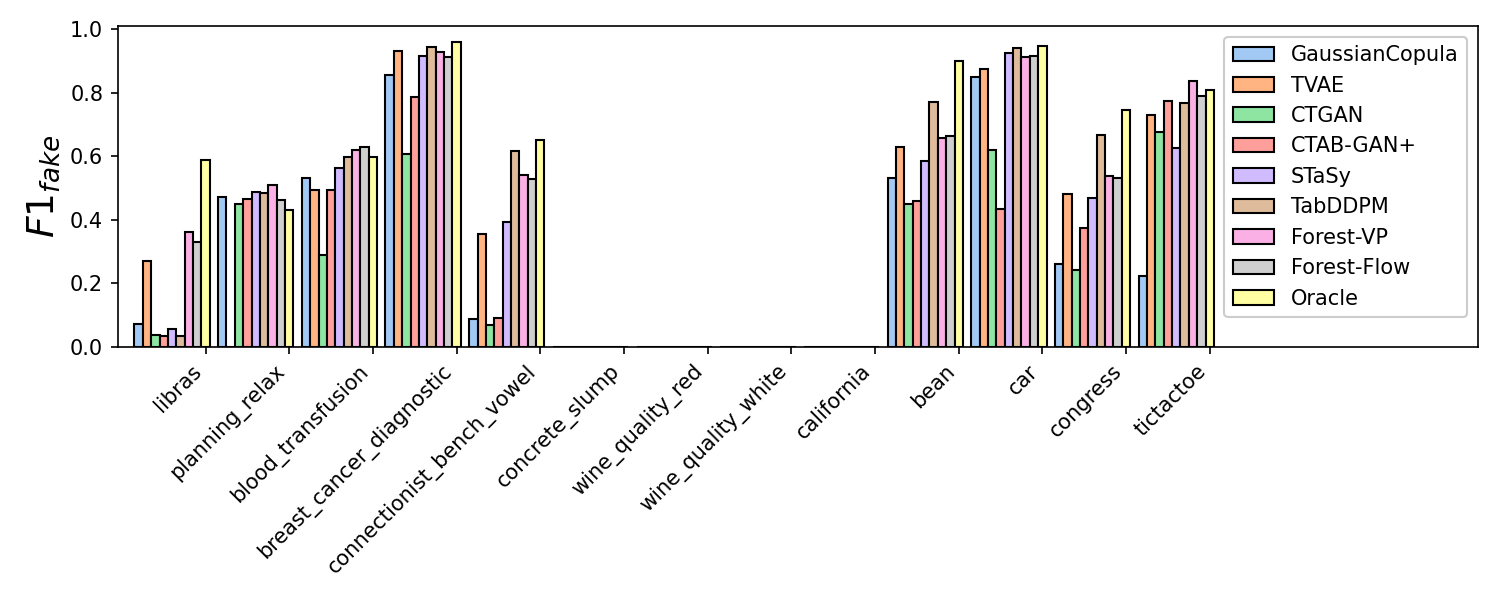}
\end{figure*}
\begin{figure*}[ht]
    \centering
    \includegraphics[width=1\textwidth]{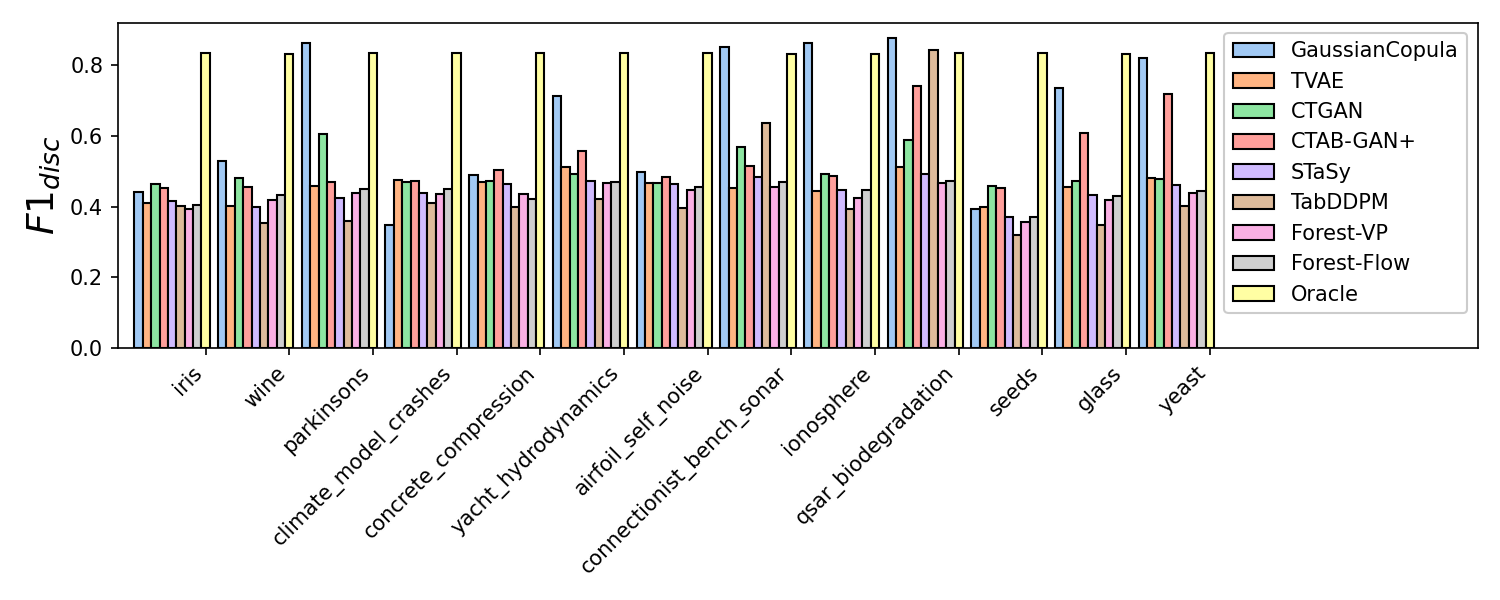}
\end{figure*}
\begin{figure*}[ht]
    \centering
    \includegraphics[width=1\textwidth]{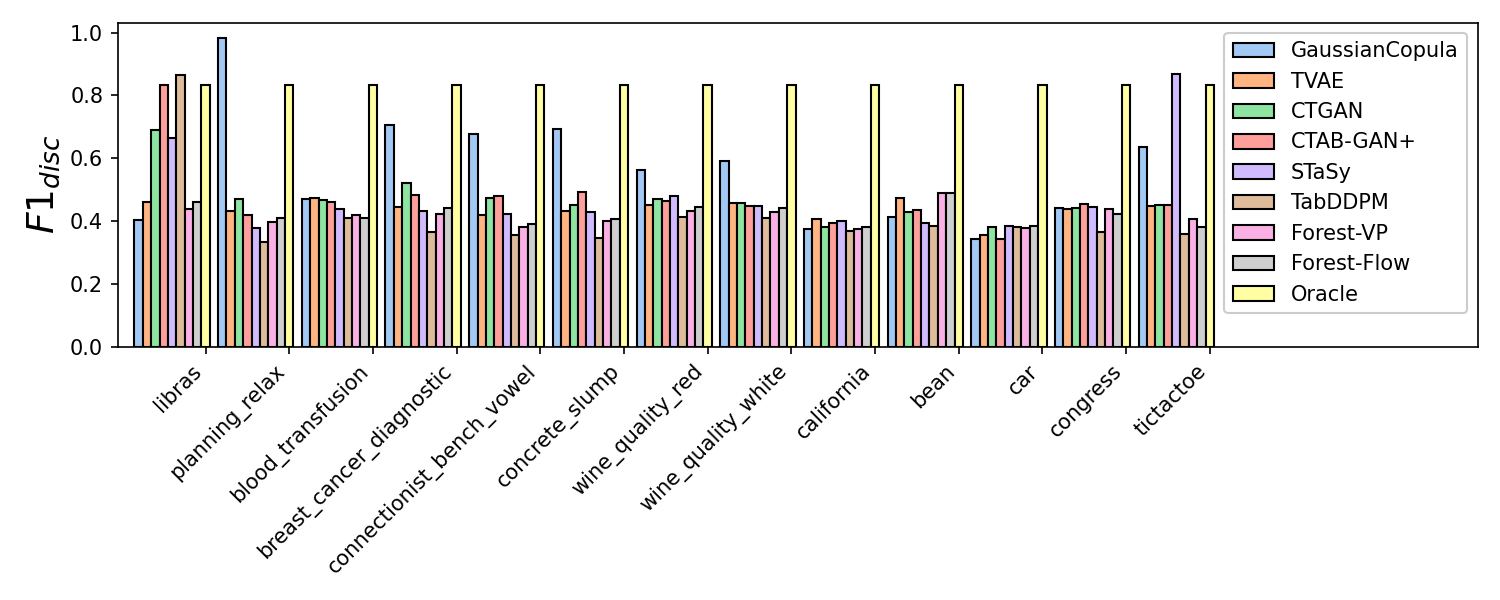}
\end{figure*}
\begin{figure*}[ht]
    \centering
    \includegraphics[width=1\textwidth]{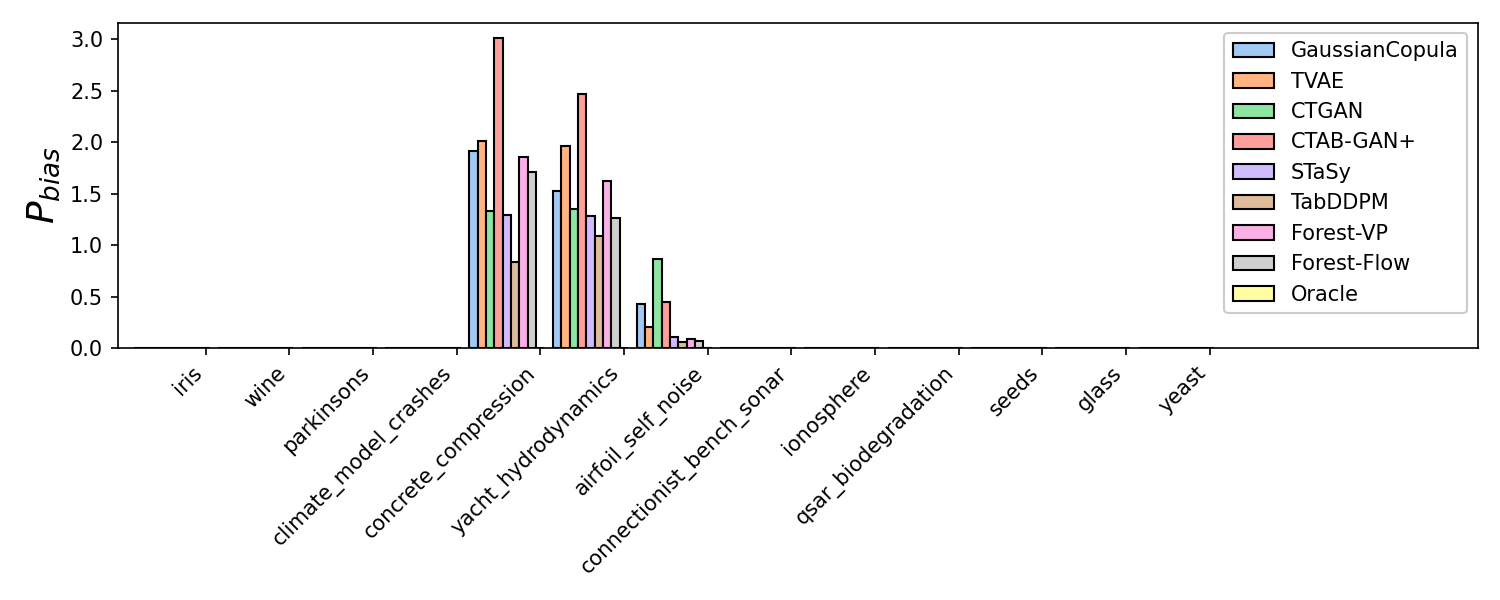}
\end{figure*}
\begin{figure*}[ht]
    \centering
    \includegraphics[width=1\textwidth]{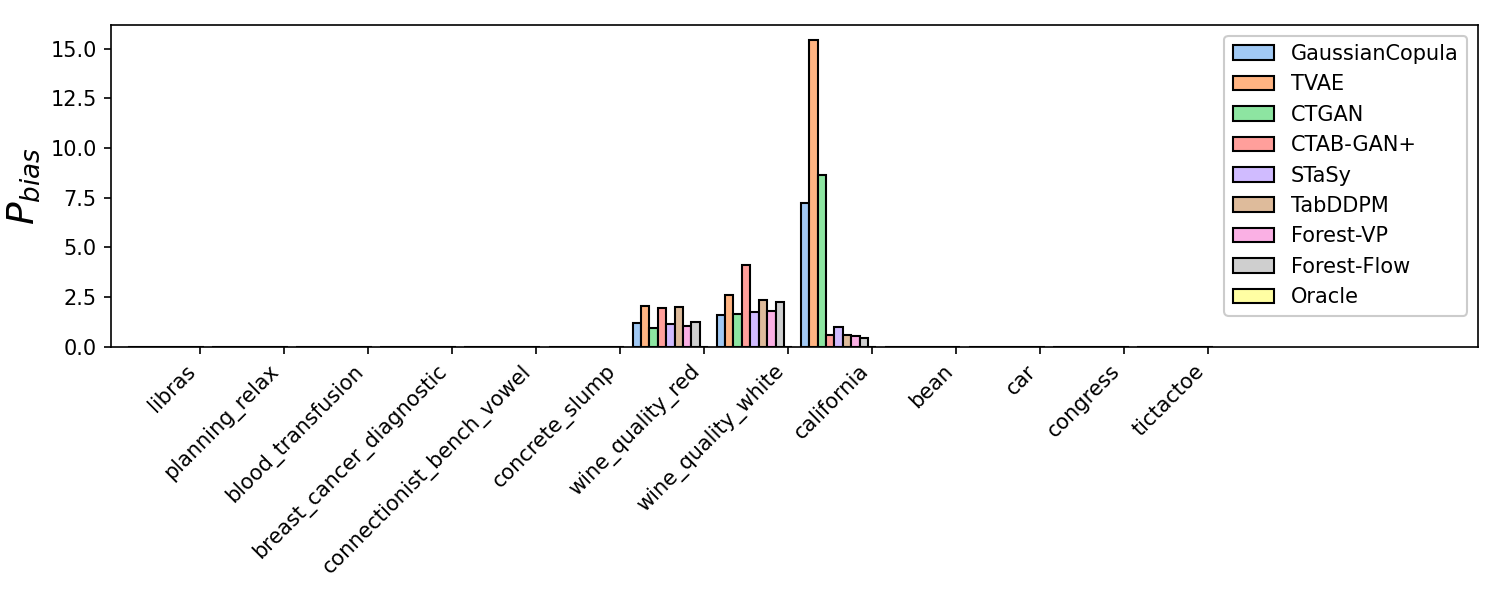}
\end{figure*}
\begin{figure*}[ht]
    \centering
    \includegraphics[width=1\textwidth]{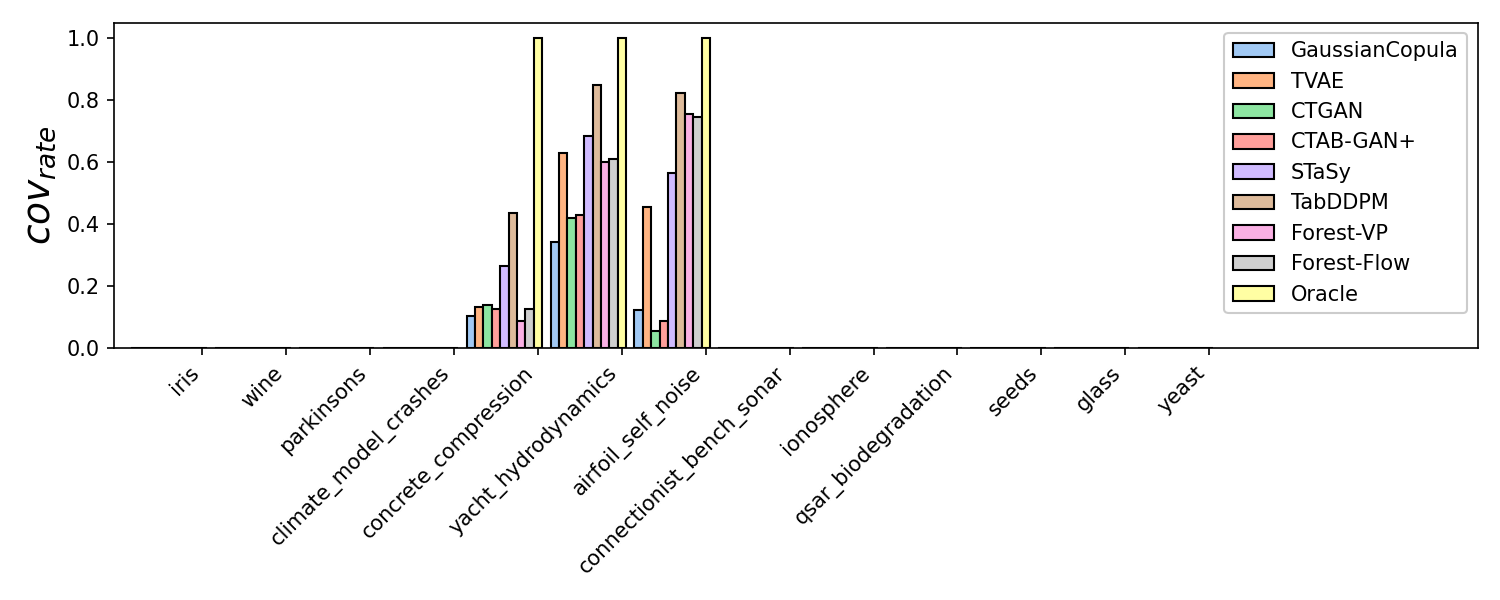}
\end{figure*}
\begin{figure*}[ht]
    \centering
    \includegraphics[width=1\textwidth]{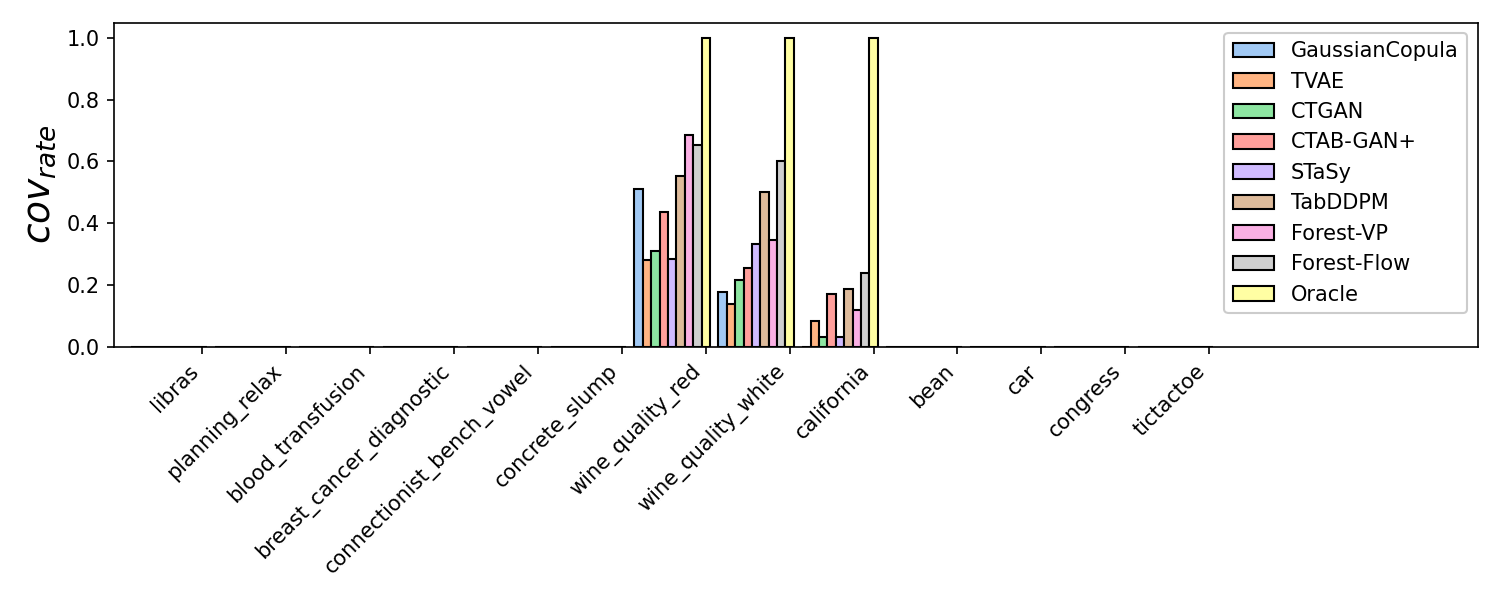}
\end{figure*}

\clearpage

\subsection{UMAP}\label{app:umap}
In this section, we show a Uniform Manifold Approximation (UMAP) \citep{mcinnes2018umap} visualization of the red wine quality dataset comparing oracle, ForestFlow (ours), and TabDDPM.

\begin{figure*}[ht]
    \centering
    \includegraphics[width=0.7\textwidth]{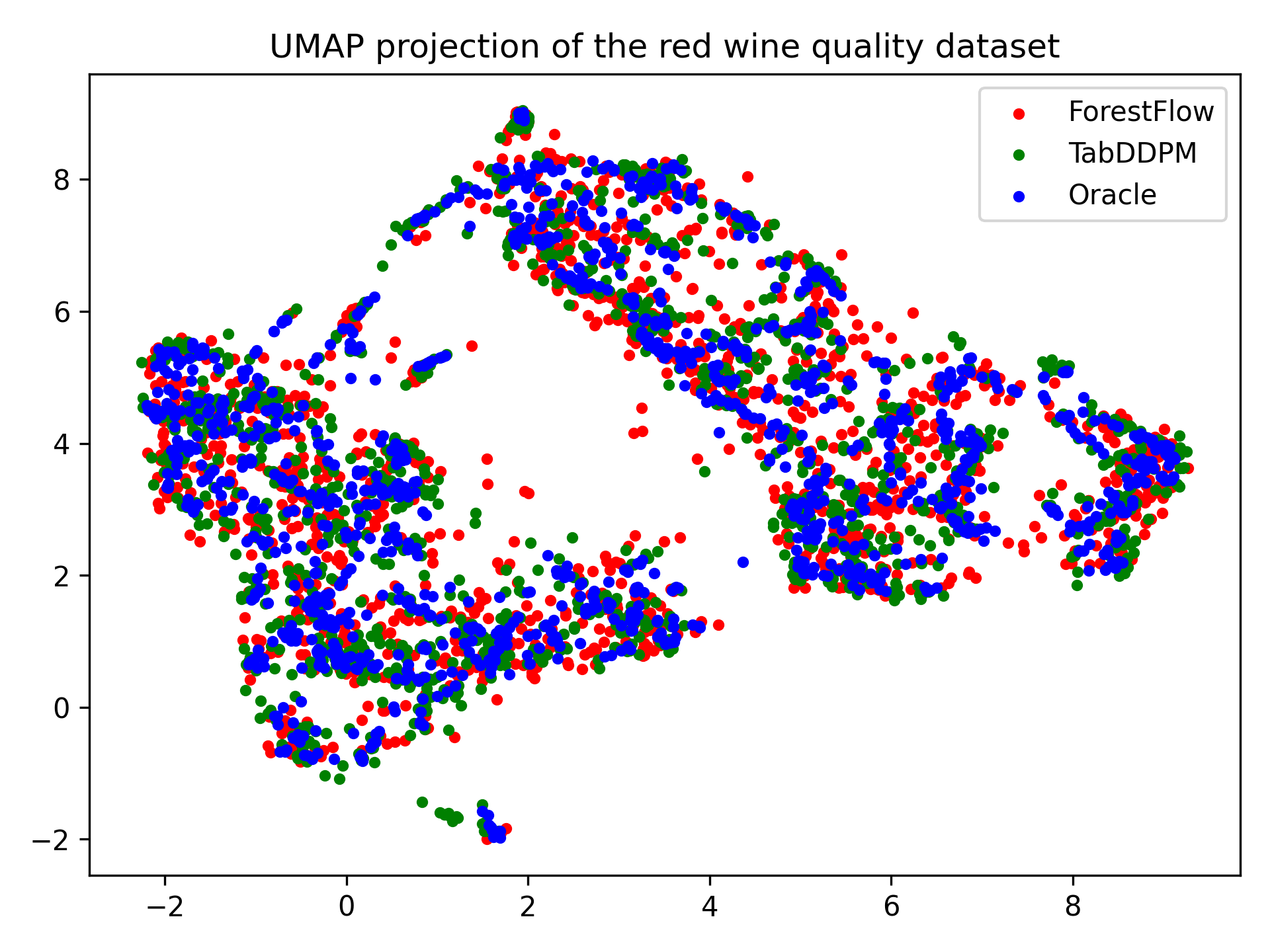}
\end{figure*}

\end{document}